\documentclass[journal,full]{IEEEtran}
\IEEEoverridecommandlockouts         
\usepackage{graphics} 
\usepackage[linesnumbered,ruled]{algorithm2e}
\graphicspath{{img/}}
\usepackage{epsfig} 
\usepackage{amsmath}
\usepackage{changepage}
\usepackage[graphicx]{realboxes}
\usepackage{adjustbox}
\usepackage{footnote}
\usepackage{amssymb}  
\usepackage{pifont}
\usepackage{cite}
\usepackage{tablefootnote} 
\usepackage[makeroom]{cancel}
\usepackage[table]{xcolor} %
\usepackage{varwidth}
\usepackage{makecell}

\usepackage[center]{subfigure}
\usepackage{multirow}
\usepackage{booktabs}
\usepackage{subfigure}
\usepackage{caption}
\usepackage{epstopdf}
\usepackage{bm}

\usepackage{algpseudocode,algorithm} 
\usepackage{color}
\usepackage{mathptmx}
\usepackage[11pt]{moresize} 
\usepackage{amsmath}
\usepackage{hhline}
\usepackage{pdflscape}
\usepackage{balance}
\usepackage{ltablex}
\usepackage{hyperref}
\usepackage{adjustbox}
\usepackage{threeparttablex}
\usepackage{enumitem}
\usepackage{units}
\usepackage{mathtools}
\usepackage{dsfont}
\usepackage{newtxtext, newtxmath} 
\usepackage{comment}
\usepackage[colorinlistoftodos]{todonotes}

\usepackage[normalem]{ulem}

\newcommand{\etal}{\textit{et al}.}

\newcommand\copyrighttext{%
	\footnotesize \copyright 2021 IEEE. Personal use of this material is permitted. Permission from IEEE must be obtained for all other uses, in any current or future media, including reprinting/republishing this material for advertising or promotional purposes, creating new collective works, for resale or redistribution to servers or lists, or reuse of any copyrighted component of this work in other works. DOI: 10.1109/TITS.2021.3096854}
\newcommand\copyrightnotice{%
	\begin{tikzpicture}[remember picture,overlay]
	\node[anchor=south,yshift=4pt] at (current page.south) {\fbox{\parbox{\dimexpr\textwidth-\fboxsep-\fboxrule\relax}{\copyrighttext}}};
	\end{tikzpicture}%
}
\usepackage{tikz}
\usepackage{textcomp}
\def\BibTeX{{\rm B\kern-.05em{\sc i\kern-.025em b}\kern-.08em
    T\kern-.1667em\lower.7ex\hbox{E}\kern-.125emX}}

\usepackage{eso-pic}
\newcommand\AtPageUpperMyright[1]{\AtPageUpperLeft{
 \put(\LenToUnit{0.5\paperwidth},\LenToUnit{-1cm}){
     \parbox{0.5\textwidth}{\raggedright\fontsize{9}{11}\selectfont #1}}
 }}
\newcommand{\journalnotice}[1]{
\AddToShipoutPictureBG*{
\AtPageUpperMyright{#1}
}
}

\newcommand{\subparagraph}{}
\usepackage{titlesec}
\titlespacing*{\section}
{0pt}{1mm plus0.7mm minus0.7mm}{1mm plus0.7mm minus0.7mm}
\titlespacing*{\subsection}
{0pt}{1mm plus0.7mm minus0.7mm}{1mm plus0.7mm minus0.7mm}
\titlespacing*{\subsubsection}
{0pt}{1mm plus0.7mm minus0.7mm}{1mm plus0.7mm minus0.7mm}
\begin{document}

\bstctlcite{IEEE_BSTcontrol}

\title{A Review and Comparative Study on Probabilistic Object Detection in Autonomous Driving}
\journalnotice{Preprint: IEEE Transactions on Intelligent Transportation Systems} 
\author{Di Feng$^{\dagger \text{1}}$, Ali Harakeh$^{\dagger \text{2}}$, Steven Waslander$^{\text{2}}$, Klaus Dietmayer$^{\text{1}}$
%
\thanks{$^{\dagger}$ Di Feng and Ali Harakeh contributed equally to this work. Listing order is random.}
\thanks{$^{\text{1}}$ Institute of Measurement, Control and Microtechnology, Ulm University, 89081 Ulm, Germany.}
\thanks{$^{\text{2}}$ Toronto Robotics and Artificial Intelligence Laboratory, University of Toronto Institute for Aerospace Studies (UTIAS), Toronto, Canada. }
\thanks{Corresponding author: \texttt{di.feng@uni-ulm.de}}}

\maketitle
\begin{abstract}
Capturing uncertainty in object detection is indispensable for safe autonomous driving. In recent years, deep learning has become the de-facto approach for object detection, and many probabilistic object detectors have been proposed. However, there is no summary on uncertainty estimation in deep object detection, and existing methods are either built with different network architectures and uncertainty estimation methods, or evaluated on different datasets with a wide range of evaluation metrics. As a result, a comparison among methods remains challenging, as does the selection of a model that best suits a particular application. This paper aims to alleviate this problem by providing a review and comparative study on existing probabilistic object detection methods for autonomous driving applications. First, we provide an overview of practical uncertainty estimation methods in deep learning, and then systematically survey existing methods and evaluation metrics for probabilistic object detection. Next, we present a strict comparative study for probabilistic object detection based on an image detector and three public autonomous driving datasets. Finally, we present a discussion of the remaining challenges and future works. Code has been made available at \url{https://github.com/asharakeh/pod_compare.git}.
\end{abstract}

\IEEEpeerreviewmaketitle

\copyrightnotice

\section{\textbf{Introduction}}\label{sec:introduction}
Capturing perceptual uncertainties is indispensable for safe autonomous driving. Consider a self-driving car operating \textit{in snowy days}, when on-board sensors can be compromised by snow; \textit{during the night-time}, when the image quality of RGB cameras is diminished; or \textit{on an unfamiliar street}, where we encounter a motorized-tricycle, which can be often seen in Asian cities but are exceedingly rare in Western Europe. In these complex, unstructured driving environments, the perception module may make predictions with varied errors and increased failure rates. Determining reliable perceptual uncertainties, which reflect perception inaccuracy or sensor noises, could provide valuable information to introspect the perception performance, and help an autonomous car react accordingly. Further, cognitive psychologists have found that humans are good intuitive statisticians, and have a frequentist sense of uncertainties~\cite{cosmides1996humans}. Therefore, reliable perceptual uncertainties could help humans better interpret the intention of autonomous cars, and enhance the development of trust in this rapidly evolving technology. As the machine learning methods (especially deep learning) have been widely applied to safety-critical computer vision problems~\cite{janai2017computer}, efforts to improve a network's self-assessment ability, reliability and interpretability with uncertainty estimation are steadily increasing \cite{mcallister2017concrete,willers2020safety}.

\begin{figure}[tpb]
	\centering
	\includegraphics[width=1\linewidth]{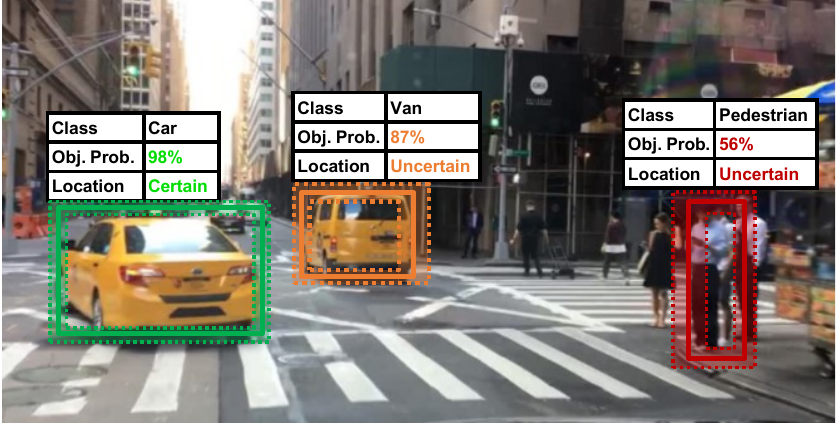}
	\caption{A \textit{conceptual} illustration of probabilistic object detection in an urban driving scenario. Each object is classified with a classification probability, and its bounding box is predicted with a confidence interval. The RGB camera image is from the BDD100k dataset~\cite{yu2020bdd100k}.} \label{fig:introduction}
\end{figure}

In this paper, we focus on object detection, one of the most important perception problems in autonomous driving. An object detector is targeted to jointly classify and localize relevant traffic participants from on-board sensor data (e.g. RGB camera images, LiDAR and Radar points) in a frame-by-frame manner. When modeling probability in object detection, we need to estimate the probability an object belongs to the classes of interest (also called semantic uncertainty as introduced in~\cite{hall2020probability}), as well as the probability distribution and the confidence interval of bounding box (i.e. spatial uncertainty~\cite{hall2020probability}), as shown conceptually in Fig~\ref{fig:introduction}. In recent years, deep learning has become the de-facto approach in object detection, and many methods of modeling uncertainties in deep neural networks have been proposed~\cite{miller2018dropout,miller2019evaluating,feng2018towards,feng2018leveraging,feng2019deep,feng2019can,feng2020leveraging,harakeh2019bayesod,he2019bounding,meyer2019lasernet,meyer2019learning,wirges2019capturing}.
However, to our knowledge, there is no work that provides a summary on uncertainty estimation in deep object detection, making it difficult for researchers to enter this field. Besides, existing probabilistic object detection models are often built with different network architectures, different uncertainty modeling approaches, and different sensing modalities. They are also tested on different datasets with a wide range of evaluation metrics. As a result, a comparison among methods remains challenging, as does the selection of the model that best suits a particular application.
\subsection*{Contributions} 
In this work, we systematically review the uncertainty estimation approaches that have been applied to deep object detectors, and conduct a comparative study for autonomous driving applications. 

First, we provide an overview of generic uncertainty estimation in deep learning (Sec.~\ref{sec:uncertainty_estimation_deep_learning}). We summarize four practical uncertainty estimation methods, discuss the types of uncertainty that can be modelled in perception problems, and introduce common metrics to quantify uncertainties. Next, we survey existing methods and the evaluation metrics designed specifically for probabilistic object detection (Sec.~\ref{sec:probabilistic_object_detection}). We then present a strict comparative study for probabilistic object detectors (Sec.~\ref{sec:comparative_study}). Based on the one-stage 2D image detector RetinaNet~\cite{lin2018focal}, we benchmark several practical uncertainty estimation methods on three public autonomous driving datasets, namely, the KITTI object detection benchmark~\cite{Geiger2012CVPR}, the BDD100k Diverse Driving Video Database~\cite{yu2020bdd100k}, and the Lyft Perception Dataset~\cite{kesten2019lyft}. In addition, we compare how uncertainties behave in RGB camera images and LiDAR point clouds, as camera and LiDAR sensors have very different sensing properties and observation noise. Finally, we present a discussion of the remaining challenges in probabilistic object detection (Sec.~\ref{sec:conclusion}).



Open source code for benchmarking probabilistic object detection is made available at: \url{https://github.com/asharakeh/pod_compare.git}.
\section{\textbf{Background for Uncertainty Estimation in Deep Learning}}\label{sec:uncertainty_estimation_deep_learning}
This section presents background knowledge of general uncertainty estimation in deep learning. First, Sec.~\ref{uncertainty_estimation:notations} defines the notation and problem formulation for predictive probability estimation in deep learning. Then, Sec.~\ref{uncertainty_estimation:bnn} introduces Bayesian Neural Networks (BNNs), which provide a natural interpretation of uncertainty estimation in deep learning. Based on the BNN framework, Sec.~\ref{uncertainty_estimation:what_uncertainty_can_we_model} decomposes predictive uncertainty into \textbf{epistemic} and \textbf{aleatoric} uncertainty, which will be widely used in this paper. Afterwards, Sec.~\ref{uncertainty_estimation:methods} introduces four practical methods for uncertainty estimation in the literature, namely, Monte-Carlo Dropout, Deep Ensembles, Direct Modeling, and Error Propagation. Finally, Sec.~\ref{uncertainty_estimation:metrics} summarizes common evaluation metrics used in uncertainty estimation. Interested readers can also find a short summary of existing uncertainty estimation benchmarks in Appx.~\ref{appendix:uncertainty_benchmarking}.

\subsection{Notation and Problem Formulation}\label{uncertainty_estimation:notations}
Denote a labeled training dataset of $N$ data pairs as $\mathcal{D} = \{\mathbf{x}_n, \mathbf{y}_n\}_{n=1}^{N}$, where $\mathbf{x}_n$ is an input data sample in the domain $\mathcal{X}$, and $\mathbf{y}_n$ its corresponding target value in the domain $\mathcal{Y}$. In classification, a target label can be one of the $C$ classes, $y \in \{1,\ldots, C\}$. In regression, a target value is usually a continuous vector with $D$ dimensions denoted by $\mathbf{y} \in \mathbb{R}^D$. In supervised learning, we aim to learn a model $f^{\mathbf{W}}:\mathcal{X}\rightarrow \mathcal{Y}$ parametrized with weights $\mathbf{W} \in \mathcal{W}$ from the training dataset $\mathcal{D}$, where $\mathcal{W}$ is the domain of weights. The model maps an input sample $\mathbf{x}$ to its corresponding output target value $\mathbf{y}$ via the model output prediction $\mathbf{\hat{y}}$. Usually, deep neural networks are interpreted as point estimators with deterministic network weights $\mathbf{W}$, and provide deterministic output predictions denoted by $\hat{\mathbf{y}} = f(\mathbf{x}, \mathbf{W})$. We expand the network output definition to encompass a predictive probability distribution denoted by $p(\mathbf{y}|\mathbf{x}, \mathcal{D})$.
\subsection{Bayesian Neural Networks}\label{uncertainty_estimation:bnn}
Bayesian Neural Networks (BNNs)~\cite{mackay1992practical,neal1995bayesian} provide a natural interpretation of uncertainty estimation in deep learning, by inferring distributions over a network's weights $\mathbf{W}$. Given a data sample $\mathbf{x}$, BNNs produce the predictive distribution $p(\mathbf{y}|\mathbf{x},\mathcal{D})$ by integrating over all values of network weights:
\begin{equation} \label{eq:bnn}
p(\mathbf{y}|\mathbf{x}, \mathcal{D}) = \int p(\mathbf{y}|\mathbf{x}, \mathbf{W}) p(\mathbf{W}|\mathcal{D}) \mathrm{d} \mathbf{W}.
\end{equation}
In this equation, $p(\mathbf{y}|\mathbf{x}, \mathbf{W})$ represents the observation likelihood, and $p(\mathbf{W}|\mathcal{D})$ represents the weight posterior distribution over the dataset. Modeling the observation likelihood is usually straight-forward by Direct Modeling (cf. Sec.~\ref{uncertainty_estimation:methods:direct_modelling}). However, analytically calculating the posterior distribution is intractable due to its high dimensionality and multi-modality (often in a space of millions of weight parameters)~\cite{sensoy2018evidential}. Multiple techniques for generating approximate solutions of the posterior distribution have been proposed~\cite{blundell2015weight,graves2011practical,depeweg2018decomposition,neal2012bayesian,welling2011bayesian,chen2014stochastic,chen2020statistical,mandt2017stochastic,maddox2019simple,ritter2018scalable}. Still, it is reported that approximation techniques in BNNs are highly sensitive to hyper-parameters, and hard to scale to large datasets and network architectures~\cite{maddox2019simple}. 

\subsection{Epistemic and Aleatoric uncertainty}\label{uncertainty_estimation:what_uncertainty_can_we_model}
Based on the BNN framework, predictive uncertainty in deep neural networks can be decomposed into \textit{epistemic} uncertainty and \textit{aleatoric} uncertainty~\cite{kendall2017uncertainties}. \textit{Epistemic}, or model uncertainty, indicates how certain a model is in using its parameters $\mathbf{W}$ to describe an observed dataset, and can be expressed by $p(\mathbf{W}|\mathcal{D})$ in Eq.~\ref{eq:bnn}. For instance, detecting an unknown object which is not present in the training dataset is expected to show high epistemic uncertainty. \textit{Aleatoric}, or data uncertainty, reflects observation noise inherent in sensor measurements of the environment, and can be modeled by $p(\mathbf{y}|\mathbf{x}, \mathbf{W})$ in Eq.~\ref{eq:bnn}. For example, detecting a distant object with only sparse LiDAR reflections, or using RGB cameras during a night drive should produce high aleatoric uncertainty. Capturing both types of uncertainty in a perception system is crucial for safe autonomous driving, as epistemic uncertainty displays the capability of a model in fitting training data, while aleatoric uncertainty reflects sensor limitations in changing environments.

In addition to object detection, which we will explicitly introduce in Sec.~\ref{sec:probabilistic_object_detection}, epistemic and aleatoric uncertainties have been estimated in different components of autonomous driving such as semantic segmentation~\cite{kendall2015bayesian,postels2019sampling,cortinhal2020salsanext}, optical flow~\cite{ilg2018uncertainty,gast2018lightweight}, depth estimation~\cite{walz2020uncertainty,gustafsson2020evaluating,eldesokey2020uncertainty,kendall2017uncertainties}, visual odometry~\cite{kendall2016modelling,yang2020d3vo,costante2020uncertainty}, image annotation~\cite{Beluch_2018_CVPR,mackowiak2018cereals,haussmann2020scalable,chitta2018large}, visual tracking~\cite{danelljan2020probabilistic}, trajectory prediction~\cite{hu2020probabilistic,meyer2020laserflow,hudnell2019robust,makansi2019overcoming} and end-to-end perception~\cite{segu2019general,michelmore2019uncertainty,tai2019visual}. Tab.~\ref{tab:summary_uncertainty_estimation_application} in the appendix lists several uncertainty estimation applications in autonomous driving. Beyond epistemic and aleatoric uncertainty, Czarnecki \etal~\cite{czarnecki2018towards} define seven sources of perceptual uncertainty to be considered for the development and operation of a perception system, which is explained in Appx.~\ref{appendix:epistemic_aleatoric_uncertainty} in detail.

\subsection{Practical Methods for Uncertainty Estimation}
\label{uncertainty_estimation:methods}
In the following, we summarize four practical methods for predictive uncertainty estimation in deep learning: Monte Carlo Dropout (MC-Dropout), Deep Ensembles, Direct Modeling, and Error Propagation. MC-Dropout and Deep Ensembles are used to model epistemic uncertainty, while Direct Modeling is used to capture aleatoric uncertainty. Error Propagation can be used to capture either epistemic or aleatoric uncertainty depending on its focus.

\subsubsection{Monte-Carlo Dropout}\label{uncertainty_estimation:methods:bnn}
The MC-Dropout method proposed by Gal~\etal~\cite{Gal2016Uncertainty} links dropout-based neural network training to Variational Inference (VI) in Bayesian Neural Networks (BNNs). They show that training a network with Stochastic Gradient Descent (SGD) with dropout is equivalent to optimizing a posterior distribution that approximates Eq.~\ref{eq:bnn}. The method is later extended in~\cite{gal2017concrete} to find the optimal dropout rate by treating the dropout probability as a hyperparameter. During test time, samples from the approximate posterior distribution are generated by performing inference multiple times with dropout enabled. Let $T$ be the total number of feed-forward passes with dropout, and $\mathbf{W}_t$ be a sample of network weights after dropout. The predictive probability can be approximated from $T$ generated samples as: 
\begin{equation} \label{eq:mc_dropout}
p(\mathbf{y}|\mathbf{x}, \mathcal{D}) \approx \frac{1}{T} \sum_{t=1}^{T} p(\mathbf{y}|\mathbf{x}, \mathbf{W}_t).
\end{equation} 
For classification, Eq.~\ref{eq:mc_dropout} corresponds to averaging the $T$ number of predicted
classification probabilities, such as softmax scores. For regression, this equation can be viewed as a mixture of distribution with $T$ equally-weighted components. In this case, the sample mean and variance can be used to describe the predictive probability distribution. In general, MC-Dropout provides a practical way to perform approximate inference in Bayesian Neural Netowrks (BNNs), and is scalable to large datasets and network architectures such as in image classification~\cite{Beluch_2018_CVPR} or semantic segmentation~\cite{blum2019fishyscapes,gustafsson2020evaluating}. However, MC-Dropout requires multiple stochastic runs during test time, usually $10$ to $50$ runs as shown in~\cite{Gal2016Uncertainty}, making it still infeasible for real-time critical systems due to high computational cost.

\subsubsection{Deep Ensembles}\label{uncertainty_estimation:methods:deep_ensembles} The method proposed by Lakshminarayanan \etal~\cite{lakshminarayanan2017simple} estimates predictive probability using an ensemble of networks, where the outputs from each network are treated as independent samples from a mixture model. Each network in the ensemble uses the same architecture, but is trained with randomly shuffled training data using a different initialization of its parameters. Let $M$ be the number of networks in an ensemble, $\{\mathbf{W}_m\}_{m=1}^M$ their weights, and $p(\mathbf{y}|\mathbf{x}, \mathbf{W}_m)$ a prediction from the $m$-th network. The predictive probability is approximated by a uniformly-weighted mixture model with $M$ components, given by:
\begin{equation}\label{eq:deep_ensembles}
p(\mathbf{y}|\mathbf{x}, \mathcal{D}) \approx \frac{1}{M} \sum_{m=1}^{M} p(\mathbf{y}|\mathbf{x}, \mathbf{W}_m).
\end{equation} 
The equation is  similar to Eq.~\ref{eq:mc_dropout}, as MC-Dropout can also be interpreted as an ensemble of networks~\cite{lakshminarayanan2017simple}. In practice, an ensemble of $5$ networks has been shown to be sufficient to approximate predictive probabilities~\cite{lakshminarayanan2017simple,snoek2019can}. Although easy to implement, the computation and memory costs of deep ensembles scale linearly with the number of networks both during training and inference, limiting its applicability for large network architectures, as discussed in \cite{gustafsson2020evaluating}.

\subsubsection{Direct Modeling}\label{uncertainty_estimation:methods:direct_modelling}
Direct Modeling assumes a certain probability distribution over the network outputs, and uses the network output layers to directly predict parameters for such a distribution. Unlike Bayesian Neural Networks which perform marginalization over weights $\mathbf{W}$ (Eq.~\ref{eq:bnn}), Direct Modeling uses point estimates of these weights to generate the predictive probability distribution $p(\mathbf{y}|\mathbf{x}, \mathcal{D})$ through $p(\mathbf{y}|\mathbf{x},\mathbf{W})$. 

The most common way to estimate the classification probability of the class $c$ is by the softmax score $p(y=c|\mathbf{x},\mathbf{W})=\hat{s}_c$, which is equivalent to a multinomial mass function. As for the regression probability, Gaussian distributions~\cite{kendall2017uncertainties,gast2018lightweight} or Gaussian Mixture Models (GMM)~\cite{choi2018uncertainty,meyer2019lasernet} are often used. For instance, it is often assumed that the target value $y$ in an one-dimensional regression problem is Gaussian distributed, given by: $p(y|\mathbf{x},\mathbf{W}) = \mathcal{N}\big(y|\hat{\mu}(\mathbf{x},\mathbf{W}), \hat{\sigma}^2(\mathbf{x},\mathbf{W})\big)$. Here, the mean, $\hat{\mu}(\mathbf{x},\mathbf{W})$, is the network standard prediction i.e. $\hat{\mu}(\mathbf{x},\mathbf{W})=f(\mathbf{x},\mathbf{W})$. The variance, $\hat{\sigma}^2(\mathbf{x},\mathbf{W})$, is also predicted by the network with an additional output layer.

Training the network is achieved by maximum likelihood estimates, where $\mathbf{W}$ are optimized to minimize the negative log likelihood $L(\mathbf{x}, \mathbf{W}) = -\log \big(p(y|\mathbf{x}, \mathbf{W}) \big)$. When using the softmax function for classification, $L(\mathbf{x}, \mathbf{W})$ is widely known as the cross-entropy loss. When using the Gaussian distribution for regression, the negative log likelihood can be written as:
\begin{equation} \label{eq:attenuated_loss}
L(\mathbf{x}, \mathbf{W}) = \frac{\big(y-\hat{\mu}(\mathbf{x},\mathbf{W})\big)^2}{2\hat{\sigma}^2(\mathbf{x},\mathbf{W})} + \frac{\log \hat{\sigma}^2(\mathbf{x},\mathbf{W})}{2}.
\end{equation}
This function can be viewed as the standard $L_2$ loss being weighted by the inverse of the predicted variance $\hat{\sigma}^2(\mathbf{x},\mathbf{W})$, and regularized with the $\log \hat{\sigma}^2(\mathbf{x},\mathbf{W})$ term.

Instead of using the standard softmax with the cross-entropy loss, Kendall \etal~\cite{kendall2017uncertainties} combine the softmax function with a Gaussian distribution to estimate classification uncertainty, by assuming that each element in the softmax logit vector is independently Gaussian distributed, with its mean and variance directly predicted by the network output layers (More detailed explanation can be found in Appx.~\ref{appendix:softmax_function_with_gaussian}). Furthermore, several works propose to estimate higher-order conjugate priors in addition to directly predicting the output probability distributions \cite{malinin2018predictive,sensoy2018evidential,amini2019deep}. Finally, Gustafsson \etal~\cite{gustafsson2020energy} use a deep neural network to directly predict the conditional target density of an energy-based model.


Since Direct Modeling estimates uncertainty via single forward pass, it is much more efficient than MC-Dropout or Deep Ensembles. However, the network's output layers and the loss function need to be modified for uncertainty estimation. It has also been shown to produce miscalibrated probabilities in classification~\cite{Gal2016Uncertainty,guo2017calibration} and regression~\cite{kuleshov2018accurate}.

\subsubsection{Error Propagation}\label{uncertainty_estimation:methods:error_propagation}
Error Propagation approximates variances (or uncertainty) in each activation layer, and then propagates variances through the whole network from input layers to output layers. For example, Postels \etal \cite{postels2019sampling} view dropout and batch normalization as a noise-injection procedure to learn uncertainty during training and propose to approximate dropout error as a covariance matrix in the noisy layers. They then propagate the error through the downstream activation layers in closed-form. In this way, the method replaces the time-consuming MC-Dropout~\cite{Gal2016Uncertainty} using a single inference, significantly reducing the computation time. Similarly, Gast \etal~\cite{gast2018lightweight} convert a standard activation layer such as ReLu into an uncertainty propagation layer by matching its first and second-order central moments. Due to the computational efficiency at inference and limited modifications required for training, Error Propagation appeals to practitioners with real-world applications.

\subsection{Uncertainty Evaluation}\label{uncertainty_estimation:metrics}
When evaluating the quality of uncertainty, one needs to study the probability distributions that capture this uncertainty. Evaluating predicted probability distributions has been extensively studied in the context of pure inference problems~\cite{brier_1950, bernardo_1979}, and more recently in the context of machine learning applications~\cite{quinonero2005evaluating, kohonen2005lessons}. Geinting~\etal\cite{gneiting2007strictly} contend that the goal of a predicted probability distribution is to maximize the \textbf{sharpness} around the correct ground truth target, subject to \textbf{calibration}. Calibration~\cite{kuleshov2018accurate, kumar2019verified} is a joint property of the estimated predicted distribution as well as the correct ground truth events or values that materialized, and reflects their statistical consistency~\cite{gneiting2007strictly}. In simple terms, if a well-calibrated distribution assigns a $0.8$ probability to an event (e.g. recognizing cars in object detection), the event should occur around $80\%$ of the time. Sharpness on the other hand quantifies the concentration of the predictive distribution around the true materialized ground truth target and is \textbf{a property of the predictive distribution only}. As an example, if the correct ground truth class is ``car'', the predicted probability distribution should ideally assign a probability of $1$ to the car category. In the context of machine learning, good predicted probability distributions need to be sharp, but ideal predicted probability distributions should be both sharp and well-calibrated~\cite{gneiting2007strictly}. 

\subsubsection{Measuring Calibration}\label{uncertainty_estimation:metrics:calibration_plot}
\begin{figure}[tpb]
	\centering
	\begin{minipage}{0.4\linewidth}
		\centering
		\includegraphics[width=1\linewidth]{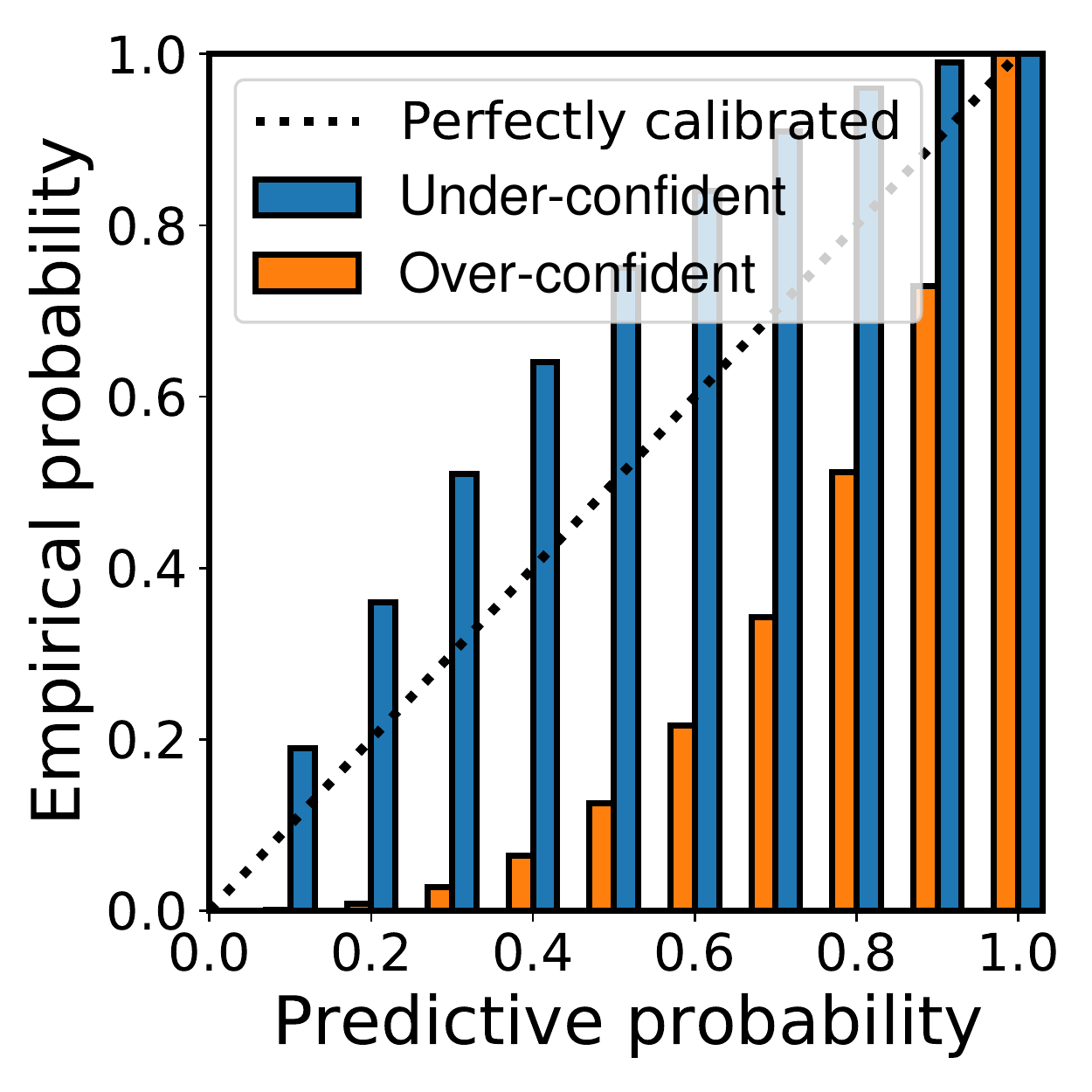}
	\end{minipage}
	\hskip 2pt
	\begin{minipage}{0.55\linewidth}
		\centering
		\caption{A calibration plot. The horizontal axis represents the predictive probability, and the vertical axis the empirical probability. A miscalibrated network can make over-confident or under-confident predictions, as indicated in blue and orange curves, respectively.}\label{fig:calibration_plot}
	\end{minipage}
	\vspace{-3mm}
\end{figure}

A calibration plot is a common tool to evaluate the calibration of distributions generated from a probabilistic model. 
It draws the predicted probability from a model on the horizontal axis and its corresponding empirical probability on an evaluation dataset on the vertical axis, as shown in Fig.~\ref{fig:calibration_plot}. A well-calibrated deep network produces a diagonal line. However, Guo~\etal~\cite{guo2017calibration} and Kuleshov~\etal~\cite{kuleshov2018accurate} have empirically identified the over- or under-confident uncertainty in many classification and regression problems, and propose propose uncertainty recalibration techniques improve uncertainty estimation.

To draw a calibration plot for an evaluation dataset with the size $N_{eval}$, the network needs to run probabilistic predictions for all data samples. The predicted probabilities from the network, denoted by $p$, are partitioned into $T$ intervals (or quantiles), i.e. $0 < p_1 < ... < p_t < ... < 1$. They correspond to the values in the horizontal axis of Fig.~\ref{fig:calibration_plot}. In each interval, the normalized empirical frequency denoted by $\hat{p}_t$ is calculated by comparing predictions with ground truths. An empirical frequency corresponds to a value in the vertical axis of Fig.~\ref{fig:calibration_plot}. Drawing the calibration plots for regression probability distributions is more complicated, requiring the evaluation of cumulative distribution functions. We refer the reader to the work of Kuleshov~\cite{kuleshov2018accurate} for a detailed description of how to draw calibration plots for regression predictive distributions.

From the calibration plot we can derive several evaluation metrics. For example, Expected Calibration Error (ECE)~\cite{guo2017calibration} measures the weighted absolute error between the actual calibration curve and the optimal calibration curve (i.e. the diagonal line). A ECE score is given by:
$\text{ECE}=\sum_{t=1}^T \frac{N_t}{N_{eval}}|p_t-\hat{p}_t|$,
where $p_t$ is the predicted probability from a network, $\hat{p}_t$ the empirical frequency, and $N_t$ the number of samples in the $t$-th interval, $\sum_{t=1}^T N_t = N_{eval}$. An ECE score ranges between $[0,1]$ with smaller values indicating better uncertainty calibration performance. A regression alternative to ECE that can be extracted from regression calibration plots has been presented by Kuleshov in~\cite{kuleshov2018accurate}. Besides ECE, Average Calibration Error (ACE) and Maximum Calibration Error are also common metrics. ACE calculates the absolute error averaged over all intervals equally, while the Maximum Calibration Error finds the maximum absolute error in all intervals. Finally, Kumar~\etal~\cite{kumar2019verified} proposed the Marginal Calibration Error (MCE) metric, as an extension of ACE. While ACE considers the uncertainty calibration quality only on an object ground truth category, MCE measures the uncertainty of a classifier's predicted distribution over all categories of interest. 

\subsubsection{Proper Scoring Rules}
Unfortunately, measuring calibration is not enough to determine the quality of output probability distributions; we are required to also evaluate its sharpness, which we have described as the concentration of probability mass assigned by output probability distributions around the correct ground truth targets. We can evaluate both calibration and sharpness of probability distributions by using \textbf{proper scoring rules}. A scoring rule $S(p(\mathbf{y}|\mathbf{x}, \mathcal{D}), \mathbf{y}_n)$ is a function that maps a predicted probability distribution and the correct ground truth target to single scalar value. A lower value of a scoring rule usually signifies predicted probability distributions that are closer to the data generating distribution from which the ground truth target is sampled. A \textbf{proper scoring rule} is a scoring rule that is only minimized if the predicted probability distribution is exactly equal to the distribution that generated the correct ground truth target. Note that we do not need access to the ``theoretical'' data generating distribution to compute a scoring rule, we only require a single sample from this distribution which is readily available in dataset $\mathcal{D}$. In addition, proper scoring rules have been shown in~\cite{brocker2009reliability} to be sum of terms relating to calibration as well as sharpness, and as such proper scoring rules are capable of considering both when evaluating predictive distributions~\cite{gneiting2007probabilistic}. For more thorough arguments on the importance of using proper scoring rules for evaluating predicted probability distributions, we refer the reader to~\cite{kohonen2005lessons, gneiting2007strictly, gneiting2007probabilistic, brocker2007properscoring, brocker2009reliability}. We now present two common proper scoring rules used in machine learning literature for evaluating predicted probability distributions~\cite{snoek2019can, lakshminarayanan2017simple}.

\textbf{Negative Log Likelihood (NLL).}
NLL is a proper scoring rule to measure the quality of predicted probability distributions of a test dataset with $N_{test}$ data points. NLL is calculated as: $\text{NLL}=-\frac{1}{N_{test}}\sum_{n=1}^{N_{test}} \log p(\mathbf{y}_n|\mathbf{x}_n, \mathcal{D})$, where $\mathbf{x}_n$ is a test data point, and $\mathbf{y}_n$ its corresponding ground truth label. NLL ranges in $(-\infty, +\infty)$, with a lower NLL score indicates a better fitting predictive distribution for that specific ground truth label. NLL has been extensively used in machine learning literature for evaluating output probability distributions for regression~\cite{lakshminarayanan2017simple} and classification~\cite{snoek2019can} tasks.

\textbf{Brier Score (BS).}
A Brier Score~\cite{brier_1950} is another common proper scoring rule used to measure the quality of output probability distributions specifically in classification. The Brier score is calculated by the squared error between a predictive probability from the network and its one-hot encoded ground truth label. Given $N_{test}$ number of test data samples, for a classification problem with $C$ number of classes, a Brier Score is given by: $\text{BS} = \frac{1}{N_{test}}\sum_{n=1}^{N_{test}} \sum_{c=1}^{C} (\hat{s}_{n,c} - y_{n,c})^2$, where $\hat{s}_{n,c}$ is a predicted classification score for the $c$-th class, such as the softmax score, and $y_{n,c}$ is ground truth label (``1'' if the ground truth class is $c$ and ``0'' otherwise). Brier scores range between 0 and 1, with lower values indicating better uncertainty estimation.

We will use NLL, Brier Score, alongside calibration errors for evaluating both regression and classification probability distributions \textbf{of in-distribution detection results} predicted by our probabilistic detectors in Sec.~\ref{sec:comparative_study:experimental_results}. It is noteworthy to mention that there exist several non-proper scoring metrics to measure the magnitude of uncertainty, such as Mutual Information, Shannon Entropy, and Total Variance. Interested readers can find additional non-proper scoring metrics in Appx.~\ref{appendix:non-proper_scoring_rules}.
\begin{figure*}[t]
	\centering
	\includegraphics[width=\textwidth,trim=1.3 1.3 1.3 1.3,clip]{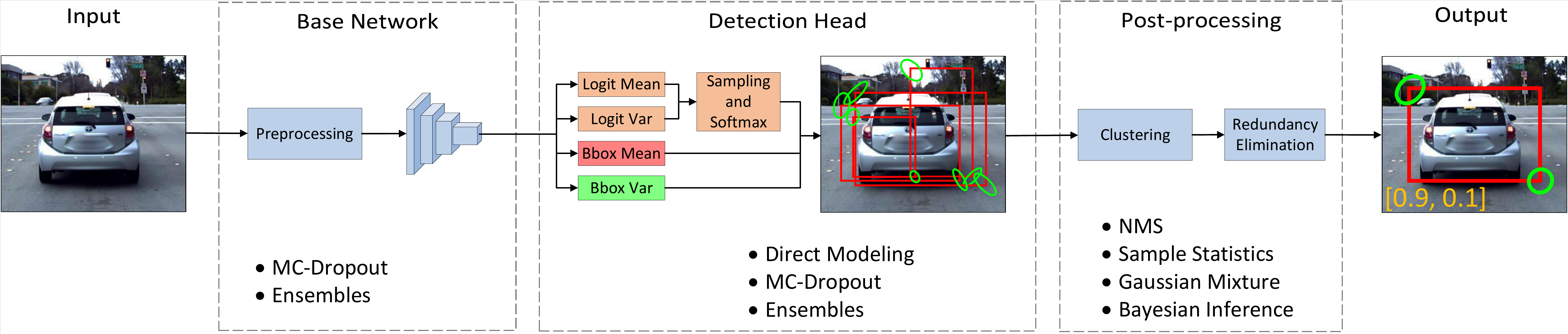}
	\caption{Illustration of the key building blocks usually present in state-of-the-art probabilistic object detectors, including the base network, detection head and post-processing stages. A list of possible variants of each building block is also presented bellow the architectural diagram. The output detections on 2D images are visualized as category probabilities (\textbf{orange}), a bounding box mean (\textbf{red}), and the $95\%$ confidence iso-contours of the bounding box corner covariance matrices (\textbf{green}).} \label{fig:probabilistic_detectors}
\end{figure*}

\section{\textbf{Probabilistic Object Detection}}\label{sec:probabilistic_object_detection}
As defined in~\cite{hall2020probability}, probabilistic object detection aims to accurately detect objects, while estimating the \textbf{semantic (category classification)} and \textbf{spatial (bounding box regression)} uncertainties in each detection. In this section, we provide a systematic summary of probabilistic object detection using deep learning approaches. We start with a background introduction of generic object detection (Sec.~\ref{subsec:deterministic_object_detection}), and then summarize each probabilistic object detection method in detail (Sec.~\ref{subsec:methodology_pods}). Afterwards, we list several common evaluation metrics and discuss their properties (Sec.~\ref{subsec:evaluation_metrics_pods}). Finally, we briefly summarize sensing modalities and use cases for existing probabilistic object detection methods in Sec.~\ref{subsec:sensor_modality} and Sec.~\ref{subsec:evaluation_metrics_pods}, respectively.
\subsection{Introduction}
\label{subsec:deterministic_object_detection}

In general, state-of-the-art object detectors are not designed to capture reliable predictive uncertainty. They predict bounding box regression variables without any uncertainty estimation, and usually classify objects with softmax scores, which may not necessarily represent reliable classification uncertainties (as discussed in~\cite{Gal2016Uncertainty}). As a result, most object detectors are \textit{deterministic}. They only predict \textit{what} they have seen, but not \textit{how uncertain} they are about it. In this regard, probabilistic object detectors are targeted to predict reliable uncertainties both in object classification and bounding box regression tasks. 

\subsection{Methodology}
\label{subsec:methodology_pods}
Probabilistic object detectors usually extend the network architectures described in the previous section (Sec.~\ref{subsec:deterministic_object_detection}) to predict probability distributions describing the category as well as the bounding box of objects in the scene~\cite{hall2020probability}. Fig.~\ref{fig:probabilistic_detectors} provides an overview of the general building blocks used to perform probabilistic object detection, which mainly consists of a base network, a detection head, and a post-processing stage. Note that both one stage and two stage approaches can be extended to predict uncertainty in a similar manner, so as 2D and 3D object detection. Therefore, we do not distinguish between the two for the remainder of this paper.

\textit{A base network} (Fig.~\ref{fig:probabilistic_detectors}, Left) maps sensor inputs to high-level feature maps, which will be further processed by a detection head. Probabilistic object detectors usually adapt base networks from well-studied deterministic object detectors to capture epistemic uncertainty using Deep Ensembles or MC-Dropout~\cite{feng2018towards, feng2019modal,kraus2019uncertainty, miller2019benchmarking, bertoni2019monoloco}. Typical base networks include VGG~\cite{simonyan2014very}, ResNet~\cite{he2016deep}, Inception \cite{szegedy2015going}, to name a few. \textit{A detection head} (Fig.~\ref{fig:probabilistic_detectors}, Middle) takes the feature maps from the base network as inputs and predicts probabilistic detection outputs. Deep Ensembles~\cite{feng2019modal, miller2019benchmarking} and MC-Dropout~\cite{harakeh2019bayesod,miller2018dropout,miller2019benchmarking,miller2019evaluating,wirges2019capturing, kraus2019uncertainty, bertoni2019monoloco} are usually used to model epistemic uncertainty in the detection head. On the other hand, the Direct Modeling approach, introduced in Sec.~\ref{sec:uncertainty_estimation_deep_learning}, models aleatoric uncertainty by introducing additional output layers to the detection head for estimating the variance of output bounding boxes~\cite{feng2018leveraging,feng2018towards, feng2019can,feng2019modal,feng2020leveraging, kraus2019uncertainty,le2018uncertainty,he2019bounding,harakeh2019bayesod,choi2019gaussian,meyer2019lasernet,meyer2019learning,pan2020towards,bertoni2019monoloco,chen2020statistical,wirges2019capturing,wang2020inferring,dong2020probabilistic} or the variance of output logits~\cite{feng2020leveraging,le2018uncertainty}. It has to be noted that the variance of the logits does not need to be estimated for generating a categorical distribution describing object categories; any detector that employs softmax as an output layer while learning with cross-entropy loss captures aleatoric uncertainty in classification.  Finally, \textit{a post-processing stage} is used to suppress or merge redundant detection outputs as is common in deterministic object detectors. Tab.~\ref{tab:summary_object_detection} summarizes the post-processing step in state-of-the-art probabilistic object detectors. The majority of methods employ the standard Non-Maximum Suppression (NMS), because it is a common component from deterministic object detectors, upon which probabilistic object detectors are built. Other methods~\cite{meyer2019lasernet,meyer2019learning,choi2019gaussian} modify NMS to take into account spatial uncertainty when deciding which redundant detection to remove, or replace NMS with Bayesian Inference~\cite{harakeh2019bayesod} to include information from all redundant detection outputs. Finally, \cite{le2018uncertainty,miller2018dropout,miller2019evaluating} take the benefits of redundant detection outputs to compute sample statistics of probability distributions, instead of directly-modeling the output probability and discarding redundant detections. 

The following section summarizes each state-of-the-art probabilistic object detector in detail. They are grouped based on what kinds of uncertainty are modelled, namely, epistemic uncertainty, aleatoric uncertainty, or both. Tab.~\ref{tab:summary_object_detection} provides an overview of all methods we have reviewed.

\subsubsection{Probabilistic Object Detectors with Epistemic Uncertainty}
\label{subsubsec:epistemic_uncertainty_pods}
Several works model epistemic uncertainty with the MC-Dropout approach. Miller \etal~\cite{miller2018dropout} estimate the classification epistemic uncertainty by modifying SSD~\cite{liu2016ssd} deterministic backend. SSD's detection head is modified with dropout layers to generate samples during test time with dropout inferences. At the post-processing step, output samples from multiple stochastic MC-Dropout runs are gathered, after performing the NMS operation on each stochastic run independently. Those redundant detections from multiple inferences are then clustered using spatial affinity. For every cluster, the categorical probability distribution of a single object is modeled using Eq.~\eqref{eq:mc_dropout}. The same architecture is used in~\cite{miller2019evaluating} for studying the effect of various clustering techniques on the quality of epistemic uncertainty in the object classification task. It was concluded that simple clustering techniques such as Basic Sequential Algorithmic Scheme (BSAS) provides higher quality uncertainty values, when compared to more complicated clustering techniques such as the Hungarian algorithm.

The successive work by Miller~\etal~\cite{miller2019benchmarking} avoid the clustering process when computing classification epistemic uncertainty. They build two detectors based on the Faster-RCNN~\cite{ren2015faster} and SSD \cite{liu2016ssd} architectures, each generates multiple samples at a certain anchor grid location by the MC-Dropout inference. For each detector separately, samples are used to compute the mean of the softmax classification output at every anchor location using Eq.~\ref{eq:mc_dropout} or Eq.~\ref{eq:deep_ensembles}. Uncertainty is then modeled using the entropy of the probability estimates from the softmax classification output. Finally, redundant detections are merged via the standard NMS operation. Experimental results show that deep ensembles outperform the MC-Dropout from a single model on the two studied architectures, in terms of the predictive uncertainty estimation.

Finally, Feng~\etal~\cite{feng2019deep} use MC-Dropout~\cite{Gal2016Uncertainty} and Deep Ensemble~\cite{lakshminarayanan2017simple} methods to estimate epistemic uncertainty for the category classification only. The classification uncertainty is employed to train a 3D LIDAR object detector in an active learning framework. Dropout layers are used only in the detection head of the proposed 3D detection architecture, and the standard NMS operation is used at the post-processing step.

\subsubsection{Probabilistic Object Detectors with Aleatoric Uncertainty}
\label{subsubsec:aleatoric_uncetainty_pods}
Most probabilistic object detectors use the Direct Modeling approach introduced in Sec.~\ref{sec:uncertainty_estimation_deep_learning} to estimate aleatoric uncertainty. These detectors are often built following four steps 1) selecting a deterministic object detector as the basic network, 2) assuming a certain probability distribution in its outputs, 3). using additional layers at the detection head to regress probability parameters, and 4) training the modified detector by incorporating uncertainty in the loss function. The Direct Modeling approach has been used in various object detection architectures, including SSD~\cite{le2018uncertainty}, Faster-RCNN~\cite{he2019bounding,feng2018leveraging, feng2020leveraging}, FCOS~\cite{lee2020localization}, Point-RCNN~\cite{pan2020towards}, and PIXOR~\cite{feng2019can}. 

The softmax function is widely used to estimate classification probability, which corresponds to a multinomial mass function. Le~\etal~\cite{le2018uncertainty} and Feng~\etal~\cite{feng2020leveraging} assume a Gaussian distribution on the logit vectors, which is then transformed to a categorical distribution through the monte-carlo sampling process described in~\cite{kendall2017uncertainties}. As for the regression probability, a majority of works assume that each bounding box regression variable is independent and follows a simple probability distribution, such as a univariate Gaussian distribution, Laplace distribution, or a combination of both (Huber mass function)~\cite{feng2018leveraging, choi2019gaussian, harakeh2019bayesod, meyer2019lasernet}. The regression variables usually include the bounding box centroid positions, extents (length, width, height), and orientations. Though this probability assumption is simple and straight-forward, it ignores correlations among regression variables, and may not fully reflect the complex uncertainties of bounding boxes, especially when objects are occluded. Instead, Pan~\etal~\cite{pan2020towards} transform the regression variables back to the eight-corner representation for 3D bounding boxes, and directly learn the corner uncertainty; Meyer~\etal~\cite{meyer2019lasernet} place a mixture of Gaussians on each regression variable; Harakeh~\etal~\cite{harakeh2019bayesod} learn a multivariate Gaussian distribution with the full covariance matrix for regression variables. The recent work by He~\etal~\cite{he2020deep} propose a more generic probability distribution which considers both correlations and multi-modal behaviours. They use a multi-variate mixture of Gaussians to describe 2D bounding boxes, and show its superiority to univariate Gaussian, multi-variate Gaussian, and a univariate mixture of Gaussian. 

Training probabilistic object detectors to predict aleatoric uncertainty is usually achieved by minimizing the Negative Log Likelihood (NLL), resulting in the well-known cross-entropy loss for classification, and the attenuated regression loss (Eq.~\ref{eq:attenuated_loss}) for bounding box regression with Gaussian distributions. In contrast, \cite{he2019bounding,meyer2019learning} introduce a probability distributions related to the ground truth bounding box parameters, and minimize the Kullback-Leibler divergence between the predictive probability distribution and the prior distribution. In this way, the predictive uncertainty is regularized with the groundtruth probability distribution, which results in more stable training and improved detection when compared to standard negative log likelihood loss~\cite{meyer2019learning,feng2020labels}. 

In addition to the standard Non-Maximum Suppression (NMS) operation, some works leverage aleatoric uncertainty at the post-processing step to merge duplicate detections. For instance, Meyer~\etal~\cite{meyer2019lasernet} propose to group detections using mean-shift clustering over corners, and combine the bounding box uncertainty and classification scores in an uncertainty-aware NMS frawework. Similarly, Choi~\etal~\cite{choi2019gaussian} design a detection criterion that considers both regression and classification uncertainty. This criterion is used to rank detections while performing the standard NMS.

Instead of the Direct Modeling method, a unique approach based on output redundancy is proposed in~\cite{le2018uncertainty} to model aleatoric uncertainty for both category classification and bounding box regression. This output redundancy method replaces the standard NMS with spatial clustering, and uses the Intersection-Over-Union (IOU) metric as a measure for detection affinity. Two probability distribution describing each output detection can then be generated by computing the sample mean of the softmax output for the category, and the sample mean and variance of cluster members for the output bounding box. Though the output redundancy method has been shown in~\cite{harakeh2019bayesod} to produce lower quality uncertainty estimates than the Direct Modeling method (measured with PDQ), it is time-efficient, and requires only small modifications to the original deterministic detection network.

\subsubsection{Probabilistic Object Detectors with both Aleatoric and Epistemic Uncertainties}
\label{subsubsec:aleatoric_epistemic_uncertainty_pods}
Aleatoric and epistemic uncertainties are jointly estimated mainly by the Direct Modeling approach together with the MC-Dropout or Deep Ensembles approaches. Given the detection samples from $T$ stochastic runs of dropout inference (or $M$ models in an network ensemble), and assuming that bounding boxes are Gaussian distributed, the mean and variance of a bounding box regression variable can be computed by an uniformly-weighted Gaussian mixture model. Using the same notation from Sec.~\ref{uncertainty_estimation:notations}, we have:
\begin{align}
\label{eq:sample_mean}
& \hat{\mu}(\textbf{x}) = \frac{1}{T} \sum\limits_{t=1}^T \hat{\mu}(\textbf{x}, \mathbf{W}_t)\\
\label{eq:sample_covariance}
& \hat{\sigma}^2(\textbf{x}) = \hat{\sigma}^2_{e}(\textbf{x}) + \hat{\sigma}^2_{a}(\textbf{x})\\
& \hat{\sigma}^2_{e}(\textbf{x})=\frac{1}{T} \sum\limits_{t=1}^T \big(\hat{\mu}(\textbf{x}, \mathbf{W}_t)\big)^2 - \big(\hat{\mu}(\textbf{x})\big)^2\label{eq:epistemic_covar}\\
& \hat{\sigma}^2_{a}(\textbf{x}) = \frac{1}{T} \sum\limits_{t=1}^T \hat{\sigma}^2(\textbf{x}, \mathbf{W_t}) \label{eq:aleatoric_covar}
\end{align}
where $\hat{\mu}(\mathbf{x},\mathbf{W}_t) = f(\mathbf{x},\mathbf{W}_t)$ and $\hat{\sigma}^2(\mathbf{x},\mathbf{W}_t)$ are the predicted mean and variance of the bounding box or classification logit vector outputs of sample $t \in [1\dots T]$. Furthermore, $\hat{\sigma}^2_{e}(\textbf{x})$ is the epistemic component of uncertainty, and is estimated as the sample variance using $T$ output samples, and $\hat{\sigma}^2_{a}(\textbf{x})$ is the aleatoric component of uncertainty, and can be computed as the average of $T$ predicted variance estimates $\hat{\sigma}^2(\mathbf{x},\mathbf{W}_t)$. Note that the final mean $\hat{\mu}(\bm x)$ and variance $\hat{\sigma}^2(\bm x)$ do not depend on the weights $\mathbf{W}$, because this variable is marginalized using averaging over $T$ samples. For more details on the uniformly-weighted Gaussian mixture model in Eq.~\eqref{eq:sample_covariance}-\eqref{eq:aleatoric_covar}, we refer the reader to~\cite{lakshminarayanan2017simple}.

While all existing works use the output layers to directly model aleatoric uncertainty, they are different in epistemic uncertainty estimation. For example, Feng~\etal~\cite{feng2018towards} add dropout layers and perform multiple inferences only in the detection head of a Faster-RCNN object detector to estimate uncertainty for both category and bounding box of objects in the scene. Wirges~\etal~\cite{wirges2019capturing} follow the same network meta-architecture, and study the effect of adding dropout layers in either the base network or the detection head. Harakeh~\etal~\cite{harakeh2019bayesod} modify a 2D image detector based on RetinaNet~\cite{lin2018focal}, and incorporate MC-Dropout in the detection head. Kraus~\etal~\cite{kraus2019uncertainty} extend a Yolov3 network, and add dropout inference after each convolutional layer in both the base network and the detection head.

As for the post-processing step, \cite{feng2018towards,kraus2019uncertainty,wirges2018object} use standard NMS. Harakeh~\etal~\cite{harakeh2019bayesod} replace it with Bayesian inference, which combines all information from redundant detections to generate the final output. Prior to the Bayesian inference step, a Gaussian mixture model is used to fuse epistemic and aleatoric uncertainty estimates per anchor. This method is shown to provide a large increase in uncertainty quality compared to the methods using standard NMS.

\begin{table*}[!htpb] 
	\centering
	\caption{A summary of probabilistic object detectors.}\label{tab:summary_object_detection}
	\resizebox{1\textwidth}{!}{\begin{tabular}{
	>{\raggedright\arraybackslash}p{0.12\textwidth}
	>{\raggedright\arraybackslash}p{0.03\textwidth}
	>{\raggedright\arraybackslash}p{0.04\textwidth}
	>{\raggedright\arraybackslash}p{0.15\textwidth}
	>{\raggedright\arraybackslash}p{0.08\textwidth}
	>{\raggedright\arraybackslash}p{0.11\textwidth}
	>{\raggedright\arraybackslash}p{0.18\textwidth}
	>{\raggedright\arraybackslash}p{0.19\textwidth}}
	\rowcolor{lightgray!35} \textbf{Reference}&\textbf{Year} & \textbf{Sensor}$^\dagger$ & \textbf{Dataset} & \textbf{Unc. Type}$^\ddagger$ & \textbf{Modeling Method}& \textbf{Post-processing} & \textbf{Evaluation Metrics}\\
	Feng \etal~\cite{feng2018towards} & 2018 & L & KITTI~\cite{Geiger2012CVPR} & E+A & MC-Dropout, Direct Modeling & NMS & F-1 Score \\
	Feng \etal~\cite{feng2018leveraging} & 2018 & L & KITTI~\cite{Geiger2012CVPR} & A & Direct Modeling & NMS & mAP \\
	Le \etal~\cite{le2018uncertainty} & 2018 & C & KITTI~\cite{Geiger2012CVPR} & A & Direct Modeling, Output Redundancy & NMS, Output Statistics & True Positive/False Positive count \\ 
	Miller \etal~\cite{miller2018dropout} & 2018 & C & COCO~\cite{lin2014microsoft}, SceneNet RGB-D~\cite{mccormac2017scenenet}, QUT Campus Dataset~\cite{sunderhauf2016place} & E & MC-Dropout & NMS, Clustering& Precision and Recall, Absolute Open-set Error, F1-Score\\
    Choi \etal~\cite{choi2019gaussian} & 2019 & C & BDD~\cite{yu2020bdd100k}, KITTI~\cite{Geiger2012CVPR} & A & Direct Modeling& Uncertainty-aware NMS & AP  \\
	Feng \etal~\cite{feng2019can} & 2019 & L & KITTI~\cite{Geiger2012CVPR}, NuScenes~\cite{nuscenes2019} & A & Direct Modeling & NMS & mAP, Expected Calibration Error \\
    He \etal~\cite{he2019bounding} & 2019 & C & COCO~\cite{lin2014microsoft}, Pascal-VOC~\cite{everingham2010pascal} & A & Direct Modeling & NMS & mAP \\
	Kraus \etal~\cite{kraus2019uncertainty} & 2019 & C  & EuroCity Persons~\cite{braun2019eurocity} & E+A & MC-Dropout, Direct Modeling & NMS, Clustering & Log Average Miss Rate\\
    Meyer \etal~\cite{meyer2019lasernet} & 2019 & L & KITTI~\cite{Geiger2012CVPR}, ATG4D~\cite{yang2018pixor} & A & Direct Modeling & Mean-shift clustering, Uncertainty-aware NMS & mAP \\
    Meyer \etal~\cite{meyer2019learning} & 2019 & L & ATG4D~\cite{yang2018pixor} & A & Direct Modeling & Mean-shift clustering, Uncertainty-aware NMS & mAP, Calibration Plots, Label Uncertainty.\\
    Miller \etal~\cite{miller2019evaluating} & 2019 & C & COCO~\cite{lin2014microsoft}, VOC~\cite{everingham2010pascal}, Underwater Scenes & E & MC-Dropout & NMS, Clustering& mAP, Uncertainty Error, AUPR, AUROC\\
    Miller \etal~\cite{miller2019benchmarking} & 2019 & C & COCO~\cite{lin2014microsoft} & E & MC-Dropout, Deep Ensembles & NMS, Clustering & PDQ\\
    
    Wirges \etal~\cite{wirges2019capturing} & 2019 & L  & KITTI~\cite{Geiger2012CVPR} & E+A & MC-Dropout, Direct Modeling& NMS, Clustering & mAP \\
    
    Lee \etal~\cite{lee2020localization} & 2020 & C & COCO~\cite{lin2014microsoft} & A & Direct Modeling & Uncertainty-aware NMS & mAP\\
     Chen \etal~\cite{chen2020monopair} & 2020 & C & KITTI~\cite{Geiger2012CVPR} & A & Direct Modeling & Pairwise spatial constrain optimization & mAP\\
    Dong \etal~\cite{dong2020probabilistic} & 2020 & R & Self-recorded data & A & Direct Modeling & Soft NMS~\cite{bodla2017soft} & mAP \\
	Feng \etal~\cite{feng2020leveraging} & 2020 & L+C & KITTI~\cite{Geiger2012CVPR}, NuScenes~\cite{nuscenes2019}, Bosch & A & Direct Modeling  & NMS & mAP \\ 
    Harakeh \etal~\cite{harakeh2019bayesod} & 2020 & C & COCO~\cite{lin2014microsoft}, VOC~\cite{everingham2010pascal}, BDD~\cite{yu2020bdd100k}, KITTI~\cite{Geiger2012CVPR} & E+A & MC-Dropout, Direct Modeling & Bayesian Fusion & mAP, Uncertainty Error, PDQ\\
    He \etal~\cite{he2020deep} & 2020& C & COCO~\cite{lin2014microsoft}, VOC~\cite{everingham2010pascal}, CrowdHuman~\cite{shao2018crowdhuman}, VehicleOcclusion~\cite{wang2017detecting} & A & Direct Modeling & Mixture of Gaussians, NMS & mAP, PDQ\\
    
    Pan \etal~\cite{pan2020towards} & 2020 & L & KITTI~\cite{Geiger2012CVPR} & A & Direct Modeling&NMS & mAP \\
    Wang \etal~\cite{wang2020inferring} & 2020 & L & KITTI~\cite{Geiger2012CVPR}, Waymo~\cite{sun2019scalability} & A & Direct Modeling & NMS & mAP, ROC curves, Jaccard-IOU\\
	\bottomrule
	\end{tabular}}
    \begin{tablenotes}
    \item \footnotesize $\dagger$ L: LiDAR, C: RGB Camera, R: Radar \ \ \ \ \ $\ddagger$ A: Aleatoric uncertainty, E: Epistemic uncertainty
    \end{tablenotes}
\end{table*}

\subsection{Evaluation Metrics}\label{subsec:evaluation_metrics_pods}
Based on the papers reviewed, there seems to be little agreement on how to evaluate probabilistic object detectors. Tab. ~\ref{tab:summary_object_detection} lists the evaluation metrics for each probabilistic object detection method in the literature, where it can be seen that each method uses different evaluation metrics, ranging from simple true positive and false positive counts~\cite{le2018uncertainty} to complex metrics such as Probability-based Detection Quality (PDQ)~\cite{hall2020probability}. The following section summarizes the common evaluation metrics, and discusses their properties when evaluating probabilistic object detection.

\subsubsection{Precision, Recall and the F1-Score}
In the context of object detection, only predictions above a pre-defined classification score threshold, $\delta_{cls}$, are considered for evaluation. They are further divided into True Positives (TP) and False Positives (FP), based on their Intersection over Union (IOU) scores with ground truth objects. Detections above a certain pre-defined IOU threshold, $\delta_{iou}$, are considered as True Positives if they share the same class labels as the ground truth objects. Detections that are misclassified, or have an IOU below the $\delta_{iou}$, are considered False Positives. False Negatives (FN) are further defined as the ground truth bounding boxes, which are either missed or not associated with predictions due to low IOU scores. The counts of TP and FP were used in \cite{le2018uncertainty} to evaluate their proposed probabilistic object detector. TP and FP scores were also used to construct the Receiver Operating Characteristic (ROC) curves in~\cite{miller2019evaluating, wang2020inferring}.

Based on TP, FP, and FN, one can derive Precision, Recall, and the F1-Score as: 
$\text{Precision} = \text{TP}/(\text{TP} + \text{FP})$, $\text{Recall} = \text{TP}/(\text{TP} + \text{FN})$, $\text{F1-Score} = 2\cdot \text{Precision}\cdot \text{Recall}/(\text{Precision} + \text{Recall})$.
Precision, recall and F1 metrics are used in \cite{feng2018leveraging, miller2018dropout} to estimate probabilistic object detector performance. One major issue of the above-mentioned evaluation metrics is that they do not take into account the values of the predicted uncertainty when determining if a detection is correct. As an example, a bounding box is considered correct if it achieves an $\text{IOU}\geq \delta_{iou}$ with a groundtruth box, a rule that does not depend on the bounding box's predicted uncertainty in any way, rendering the uncertainty prediction irrelevant to the evaluation.

\subsubsection{Mean Average Precision}
Average Precision (AP) was proposed by Everingham \etal~\cite{everingham2010pascal} to evaluate the performance of object detectors. It is defined as the area under the continuous precision-recall (PR) curve, approximated through numeric integration over a finite number of sample points~\cite{everingham2010pascal}. Mean AP (mAP) is the average of all AP values computed for every object category found in the test dataset. It can also be averaged across multiple IOU thresholds $\delta_{iou}\in[0.5,\dots,0.95]$ as proposed in \cite{lin2014microsoft}, a measure referred to as COCO mAP. Though mAP is the standard evaluation metric in object detection, it does not take the predictive uncertainties into account. In fact, if two probabilistic object detectors predict bounding boxes with the same mean values but wildly different covariance matrices, they will have the same mAP performance. Nevertheless, the majority of recent work on probabilistic object detection~\cite{feng2018leveraging,choi2019gaussian,he2019bounding, meyer2019lasernet, meyer2019learning, wirges2019capturing, lee2020localization,dong2020probabilistic,feng2020leveraging,wang2020inferring} still use mAP as the only metric to provide a quantitative assessment of their proposed methods, emphasizing a secondary effect of accuracy improvement when integrating probabilistic detection methods, instead of focusing on the correctness of the output distribution. This leads to the need of better and more consistent metrics, which we highlight in Sec.~\ref{sec:challenges}.
\subsubsection{Probability-based Detection Quality}
Hall \etal~\cite{hall2020probability} proposed the Probability-based Detection Quality (PDQ) as a metric to measure the quality of 2D probabilistic object detection on images. PDQ was later used by \cite{harakeh2019bayesod, miller2019benchmarking} for uncertainty evaluation. PDQ is designed to jointly evaluate semantic uncertainties and spatial uncertainties in image-based object detection. The semantic uncertainties are evaluated by matching the predicted classification scores with the ground truth labels for each object instance in images. The spatial uncertainties are encoded by covariance matrices, by assuming a Gaussian distribution on the top-right or the bottom-left corner of a bounding box. The optimal PDQ is achieved, when a predicted probability correlates the prediction error, for example, when a large spatial uncertainty correlates with an inaccurate bounding box prediction.  

PDQ assigns every ground truth an optimal corresponding detection using the Hungarian algorithm, removing the dependency on IOU thresholding that is required for mAP. Furthermore, PDQ measures the probability mass assigned by the detector to \textit{true positive detection} results, and is evaluated at a single classification score threshold requiring practitioners to filter low scoring output detection results prior to evaluation. PDQ assumes 2D Gaussian corner distributions and cannot evaluate methods assuming a Laplace distribution such as~\cite{meyer2019lasernet, meyer2019learning}. Finally, because of how the spatial quality is defined, PDQ can only be computed for 2D probabilistic detection results defined in image space, with no straightforward extensions available for 3D probabilistic object detectors.

\subsubsection{Minimum Uncertainty Error}
The Uncertainty Error (UE) was first proposed  by Miller \etal~\cite{miller2019evaluating} to evaluate probabilistic object detectors. UE can be thought of as the probability that a simple threshold-based classifier makes a mistake when classifying output detections into true positives and false positives using their predicted spatial uncertainty estimates. UE ranges between $0$ and $0.5$, and as the uncertainty error approaches $0.5$, using the predicted uncertainty estimates to separate true positives from false positives is no better than a random classifier. The best uncertainty error achievable by a detector over all possible thresholds is called the Minimum Uncertainty Error (MUE), and is used to compare probabilistic object detectors in \cite{miller2019evaluating, harakeh2019bayesod}.

Similar to mAP, MUE requires an IOU threshold $\delta_{iou}$ to determine which detections are counted as true positives, and is therefore threshold-dependant. 
Furthermore, MUE is not affected by the scale of uncertainty estimates. In other words, rescaling or shifting the uncertainty values for all detections with the same value results in a constant MUE score. Therefore, MUE is only capable of providing information on how well the estimated uncertainty can be used to separate true positives from false positives, but not on the quality of estimated uncertainty.

\subsubsection{Jaccard IoU}
Recently, Wang \etal~\cite{wang2020inferring} propose Jaccard IoU (JIoU) as a probabilistic generalization of IoU. Unlike IoU which simply compares the (deterministic) geometric overlap between two bounding boxes, JIoU measures the similarity of their spatial distributions. Such distributions can be predicted by probabilistic object detection networks, or by inferring the uncertainties inherent in ground truth labels (which serve as the ``references of probability distribution'')~\cite{wang2020inferring}. In fact, JIoU simplifies to IoU when bounding boxes are assumed to follow simple uniform distributions. Similar to IoU, JIoU ranges within $[0,1]$. It is maximized only when two bounding boxes have the same locations, same extents, as well as same spatial distributions. In general, JIoU provides a natural extension of IoU to evaluate probabilistic object detection. It also considers ambiguity and uncertainty inherent in ground truth labels, which are ignored by other evaluation metrics such as mAP and PDQ. However, a separate model is needed to approximate ground truth spatial uncertainty in the labeling process, which has only been proposed for LiDAR point cloud~\cite{wang2020inferring,feng2020labels2}. Therefore, the use of JIoU is limited to evaluating LiDAR-based probabilistic object detection to date.

\subsubsection{Proper Scoring Rules}
To our surprise, none of the previous work on probabilistic object detectors use proper scoring rules for evaluating the quality of their predicted probability distributions. This observation was also noted in the recent work of Harakeh and Waslander~\cite{harakeh2021estimating}, where it was argued that proper scoring rules are essential for a theoretically sound evaluation of predicted distributions from probabilistic object detectors. It was also shown in~\cite{harakeh2021estimating} that mAP, MUE, and PDQ \textbf{are not proper scoring rules}. For our experiments in Sec.~\ref{sec:comparative_study:experimental_results}, we will use NLL as a proper scoring rule for evaluating the quality of predicted category and bounding box probability distributions.

\subsection{Sensor Modalities}\label{subsec:sensor_modality}
Different sensors have different sensing properties and observation noises, and thus can affect the behaviours of aleatoric uncertainty in probabilistic object detection. Most studies to date employ LiDARs~\cite{feng2018towards,feng2018leveraging,feng2019can,feng2019deep,feng2019modal,meyer2019lasernet,meyer2019learning,wirges2019capturing,pan2020towards} and RGB cameras~\cite{kraus2019uncertainty,le2018uncertainty,miller2018dropout,miller2019benchmarking,miller2019evaluating,harakeh2019bayesod,he2019bounding,choi2018uncertainty,bertoni2019monoloco,chen2020monopair} in probabilistic object detection. Only Dong \etal~\cite{dong2020probabilistic} propose to model uncertainty in radar sensors. The LiDAR-based networks often deal with 3D detection for single object class or a few object classes, such as ``Car'', ``Cyclist'' and ``Pedestrian''. In comparison, the camera-based approaches focus on the multi-class 2D detection with more diverse classes. For example, Harakeh \etal~\cite{harakeh2019bayesod} evaluate 2D probabilistic object detection with ten common road scene object categories in the BDD dataset~\cite{yu2020bdd100k}, as well as 80 object categories in the COCO dataset~\cite{lin2014microsoft}.

\subsection{Applications and Use Cases of Predictive Uncertainty}\label{subsec:application_pods}
The behaviours of epistemic and aleatoric uncertainties in a network are studied in~\cite{feng2018towards,wirges2019capturing,kraus2019uncertainty,bertoni2019monoloco}. These works verify that epistemic uncertainty is related to the detections different from the training samples (open-set detections), whereas aleatoric uncertainty reflects complex noises inherent in sensor observations such as distance and occlusion. In addition, Bertoni \etal~\cite{bertoni2019monoloco} study the task ambiguity in depth estimation with monocular images, and Wang \etal~\cite{wang2020inferring} study the uncertainty inherent in ground truth labels for LiDAR-based object detection.

Epistemic uncertainty is applied to detect out-of-distribution (OOD) objects in~\cite{miller2018dropout}. They find that thresholding the classification uncertainty captured by MC-Dropout reduces  detection errors for OOD objects when compared to the standard softmax confidence. Epistemic uncertainty is also used in~\cite{feng2019deep} to select unseen samples in an active learning framework, in order to train LIDAR 3D object detectors with minimal human labeling effort.

Aleatoric uncertainty is mainly used to improve detection accuracy. This can be achieved by directly modeling uncertainty in the training loss (e.g. the attenuated regression loss shown in Eq.~\ref{eq:attenuated_loss})~\cite{feng2018leveraging,feng2020leveraging,he2019bounding,pan2020towards}, such that networks learn to handle noisy data and enhance the robustness. It can also be achieved by incorporating uncertainty in the post-processing step when merging duplicate detections. For example, Choi \etal~\cite{choi2019gaussian} rank detections during NMS based on location uncertainty in addition to classification scores. Meyer \etal~\cite{meyer2019lasernet} adapt the IoU threshold when merging detections to their location uncertainty instead of a pre-defined value for all detections. Finally, thresholding aleatoric uncertainty has been shown to effectively remove False Positive samples in~\cite{le2018uncertainty}.

\section{\textbf{Comparative Study}}\label{sec:comparative_study}
A major challenge for new researchers and practitioners entering the probabilistic object detection domain is the lack of a consistent benchmark among methods presented in literature. This problem is evident when looking at Tab.~\ref{tab:summary_object_detection}, where one can see that it is uncommon for state-of-the-art methods to use the same dataset and evaluation metric combinations, leading to difficulty in determining which method works best for autonomous driving. In this section, we attempt to alleviate this problem by performing a comparative study of the performance of common modifications described in Sections~\ref{sec:uncertainty_estimation_deep_learning} and~\ref{sec:probabilistic_object_detection} that are used to extend deterministic object detectors to predict probability distributions. By using the same base network, datasets, and multiple evaluation metrics, we aim to have a controlled setup to perform a fair comparison. Specifically, we focus on 2D image-based detection for seven object categories. All networks are trained on the large-scale BDD100K dataset~\cite{yu2020bdd100k}, and evaluated on the BDD100K, KITTI~\cite{Geiger2012CVPR}, and Lyft~\cite{kesten2019lyft} datasets. Note that due to the lack of large-scale open datasets with diverse multiclass labels, we do not conduct experiments for 3D object detection and for LiDAR point cloud. 

Sec.~\ref{sec:comparative_study:pod} summarizes the configurations of probabilistic object detectors used for the comparative study. Sec.~\ref{sec:comparative_study:experimental_setup} and Sec.~\ref{sec:comparative_study:training_inference} provide the experimental setup and implementation details, respectively. Sec.~\ref{sec:comparative_study:experimental_results} presents the main experimental results and our conclusions. Finally, Sec.~\ref{sec:comparative_study:lidar_camera} compares the behaviours of aleatoric uncertainties in an image and a LiDAR-based object detectors. 
\begin{table*}[!htpb] 
	\centering
	\caption{A summary of uncertainty modeling methods used for the Base Network, Detection Head, and Post-processing stages of each of our implemented probabilistic object detectors.}\label{tab:our_object_dets}
	\resizebox{1\textwidth}{!}{\begin{tabular}{
	>{\raggedright\arraybackslash}p{0.2\textwidth}
	>{\raggedright\arraybackslash}p{0.08\textwidth}
	>{\raggedright\arraybackslash}p{0.1\textwidth}
	>{\raggedright\arraybackslash}p{0.23\textwidth}
	>{\raggedright\arraybackslash}p{0.31\textwidth}}
	\rowcolor{lightgray!35} \textbf{Method}& \textbf{Unc. Type$\ddagger$}& \textbf{Base Network} &\textbf{Detection Head}& \textbf{Post-processing}\\
	1. Output Redundancy&A&Unmodified &Unmodified & Data Association $\rightarrow$ Sample Statistics\\
    2. Loss Attenuation&A & Unmodified& Reg and Cls variance& NMS\\
    3. BayesOD&A&Unmodified& Reg and Cls variance&Data Association $\rightarrow$ Bayesian Fusion\\
    4. Loss Attenuation + Dropout &A + E&Unmodified& Reg and Cls variance, MC-Dropout & NMS\\
    5. BayesOD + Dropout&A + E&Unmodified& Reg and Cls variance, MC-Dropout&Data Association $\rightarrow$ Bayesian Fusion\\
    6. Pre-NMS Ensembles &A + E&Ensembles&Reg and Cls variance, Ensembles& Data Association $\rightarrow$ Gaussian Mixture $\rightarrow$ NMS\\
    7. Post-NMS Ensembles&A + E&Ensembles &Reg and Cls variance, Ensembles& NMS $\rightarrow$ Data Association $\rightarrow$ Gaussian Mixture\\
    8. Black Box&E&Unmodified& MC-Dropout&NMS $\rightarrow$ Data Association $\rightarrow$ Gaussian Mixture\\
	\bottomrule
	\end{tabular}}
    \begin{tablenotes}
    \item \footnotesize $\ddagger$ A: Aleatoric uncertainty, E: Epistemic uncertainty
    \end{tablenotes}
\end{table*}
\subsection{Our Probabilistic Object Detectors}\label{sec:comparative_study:pod}
For a fair evaluation, we re-implement the various uncertainty estimation mechanisms described in Sec.~\ref{subsec:methodology_pods} using a 2D image-based object detector, RetinaNet~\cite{lin2018focal}, as the basic network. The new network models the bounding box parameters as multivariate Gaussians with a diagonal covariance matrices, and uses the softmax function to predict the parameters of categorical distributions for object classification. The probabilistic extensions to RetinaNet are summarized in Tab.~\ref{tab:our_object_dets}.

To extend RetinaNet for modeling aleatoric uncertainty, we implement three different approaches: Loss Attenuation, BayesOD, and Output Redundancy, which are summarized in rows $1-3$ of Tab.~\ref{tab:our_object_dets}. \textbf{Loss Attenuation} uses \textbf{Direct Modeling} for estimating the uncertainty in regression and classification outputs, by predicting the means and variances for bounding box regression variables and softmax logits. The network is trained using the attenuated regression losses as well as the modified classification loss to learn softmax logit variances (details cf. Appx.~\ref{appendix:softmax_function_with_gaussian}). Existing probabilistic object detectors which use this Direct Modeling approach are summarized in Sec.~\ref{subsubsec:aleatoric_uncetainty_pods}. \textbf{BayesOD} modifies the post-processing step of \textbf{Loss Attenuation} to replace non-maximum suppression (NMS) with data association followed by Bayesian Inference as proposed in~\cite{harakeh2019bayesod}. Since only a single modification is required to formulate BayesOD from Loss Attenuation, comparing the performance of the two provides insights on how useful Bayesian inference is when compared to standard NMS. The final approach for estimating aleatoric uncertainty is \textbf{Output Redundancy}, which leaves the base network and detection head of the deterministic RetinaNet model intact, and only modifies the post-processing step to cluster redundant output detections (Sec.~\ref{subsubsec:aleatoric_uncetainty_pods}). To estimate category and bounding box uncertainty, Output Redundancy extracts sample statistics from clusters of output detections, avoiding the auxiliary variance predictions used in most probabilistic object detection approaches.

To model epistemic uncertainty, methods proposed in \cite{miller2018dropout, miller2019evaluating} are implemented by incorporating dropout into the detection head of a deterministic RetinaNet model. We follow the implementation described in~\cite{miller2019evaluating} which clusters output after the NMS step, treating the object detector as a \textbf{Black Box} (row 8 of Tab.~\ref{tab:our_object_dets}). Since the \textbf{Black Box} and \textbf{Output Redundancy} implementations do not explicitly model the variance parameters of the bounding box output of RetinaNet, comparing the quality of their bounding box output uncertainty to methods that use Direct Modeling of the variance parameters should provide an indicator on the value of Direct Modeling in probabilistic object detection.

Finally, the probabilistic object detectors which jointly model epistemic and aleatoric uncertainties are summarized in rows $4-7$ of Tab.~\ref{tab:our_object_dets}. To model both types of uncertainty, we extend \textbf{Loss Attenuation} and \textbf{BayesOD} to perform stochastic dropout runs by modifying the detection head with dropout layers with two extensions: \textbf{Loss Attenuation + Dropout} and \textbf{BayesOD + Dropout}. \textbf{Loss Attenuation + Dropout} extends \textbf{Loss Attenuation} by performing multiple stochastic MC-Dropout runs during inference, followed by the standard NMS. The same setting has been implemented in~\cite{feng2018towards}. \textbf{BayesOD + Dropout} performs MC-Dropout during inference, and then associates detections by a Gaussian Mixture Model to fuse aleatoric and epistemic uncertainty estimates according to Eq.~\ref{eq:sample_mean} and Eq.~\ref{eq:sample_covariance}. This method has been implemented previously in~\cite{harakeh2019bayesod}.

Additionally, we train independent \textbf{Loss Attenuation} models to construct deep ensembles~\cite{lakshminarayanan2017simple}. We provide two implementations to fuse detections from ensemble results: \textbf{Pre-NMS Ensembles} fuse detections before the NMS step, and \textbf{Post-NMS Ensembles} performs data association and fusion after the NMS step. These ensemble + loss-attenuation approaches have not been attempted in literature, and can be considered a direct application of the uncertainty estimation mechanisms in~\cite{lakshminarayanan2017simple} to the object detection problem.

\subsection{Experimental Setup}\label{sec:comparative_study:experimental_setup}
Recent work on evaluating uncertainty estimates of deep learning models for classification tasks~\cite{snoek2019can} suggests the necessity of uncertainty evaluation under dataset shift. In this regard, we chose three common object detection datasets with RGB camera images for autonomous driving, in order to evaluate and compare our probabilistic object detectors: the \textbf{Berkeley Deep Drive 100K (BDD) Dataset~\cite{yu2020bdd100k}}, the \textbf{KITTI autonomous driving dataset~\cite{Geiger2012CVPR}}, and the \textbf{Lyft autonomous driving dataset~\cite{kesten2019lyft}}. All presented probabilistic object detectors are trained using the official training frames of BDD to detect seven common dynamic object categories relevant to autonomous driving: Car, Bus, Truck, Person, Rider, Bicycle, and Motorcycle. The detectors are then evaluated across all three datasets to quantify the uncertainty estimation quality with or without dataset shift.

\begin{table*}[!tpb]
\centering
\resizebox{0.8\textwidth}{!}{
\begin{tabular}{c c c c c c}
\rowcolor{lightgray!35} \textbf{Method} & \textbf{mAP}~(\%) $\uparrow$ & \textbf{PDQ}~(\%) $\uparrow$ & \textbf{Cls NLL} $\downarrow$  & \textbf{Reg NLL} $\downarrow$& \textbf{Frame Rate}~(FPS) $\uparrow$\\ 
Deterministic RetinaNet (Baseline) & 28.62& - &0.70 &-&22.01\\

Output Redundancy& 26.90& 18.40& 0.91&2421.67 & \textbf{15.21}\\
Loss Attenuation& 28.54&33.29 &0.70&15.27 & 14.97\\
BayesOD& 28.65 &\textbf{36.55} & 0.70& 13.06& 9.75\\
Loss Attenuation + Dropout&27.51 &32.34&0.74 & 15.36& 2.61\\
BayesOD + Dropout&27.60 &36.02 & 0.75& \textbf{13.03}& 2.21\\
Pre-NMS Ensembles&\textbf{29.23} &33.21 & 0.71&15.24  &6.06\\
Post-NMS Ensembles& 29.18 &32.47 & \textbf{0.69}& 15.23& 3.35\\
Black Box&28.10 &18.22 & \textbf{0.69}& 2806.52& 2.45\\
\bottomrule
\end{tabular}
}
\caption{An evaluation with no dataset shift by testing detectors on the BDD validation dataset. Results are averaged over all seven dynamic object categories.}
\label{table:bdd_all_classes}
\end{table*}

\begin{table}[!tpb]
\centering
\resizebox{0.9\linewidth}{!}{
\begin{tabular}{c c c c}
\rowcolor{lightgray!35} \textbf{Method} & \textbf{BDD} & \textbf{KITTI} & \textbf{Lyft}\\ 
Deterministic RetinaNet (Baseline) &40.63& 37.11 & 5.33\\ 
Output Redundancy& 38.75& 34.99&5.21\\
Loss Attenuation&41.09& 39.15 &5.96 \\
BayesOD&41.33& 39.32&\textbf{6.01}\\
Loss Attenuation + Dropout& 40.39& 38.90&5.97\\
BayesOD + Dropout& 40.52& 38.99&5.98\\
Pre-NMS Ensembles&41.85& 39.50&5.953\\
Post-NMS Ensembles& \textbf{41.95}& \textbf{39.52}&5.949\\
Black Box&40.53 & 37.01&5.231\\
\bottomrule
\end{tabular}
}
\caption{Mean average precision (mAP) results of evaluation under dataset shift. The mAP scores are computed using the car and pedestrian categories on the BDD, Lyft, and KITTI datasets.}
\label{table:comparison_studies_ood}
\end{table}

\textbf{Evaluation without Dataset Shift:}
First, we conduct comparative experiments without dataset shift, by testing the probabilistic object detectors on the official BDD validation dataset. This experimental setup corresponds to common object detection experiments in the literature, which quantify the performance of probabilistic object detectors on a test dataset with similar data distribution to the training set.

\textbf{Evaluation with Dataset Shift:}
To quantify performance under dataset shift, probabilistic detection models trained \textbf{only on the BDD dataset} are tested on $7,481$ frames from the \textbf{KITTI}~\cite{Geiger2012CVPR} dataset and $5,000$ randomly chosen frames from the \textbf{Lyft}~\cite{kesten2019lyft} autonomous driving dataset. The results on Lyft and KITTI are then compared against the results on the BDD validation dataset. Here, we only evaluate on the Car and Person object classes, as they are the only two classes that have the same definition in all three datasets. More detailed information on three datasets as well as the dataset shift from BDD to KITT and BDD to Lyft can be found in Appx.~\ref{appendix:datasets}.

\subsection{Training and Inference}\label{sec:comparative_study:training_inference}
For all probabilistic object detectors mentioned above, we follow the training setup proposed in the original RetinaNet paper~\cite{lin2018focal}, using the authors' original open source implementation provided by the Detectron2~\cite{Detectron2018} 2D object detection library. More specifically, we use RetinaNet with a Resnet-50 backbone and a feature pyramid network~(FPN). We train all models on the BDD training dataset split using a batch size of $4$ for $90k$ iterations using stochastic gradient descent with $0.9$ momentum. We use a base learning rate of $0.0025$ which is dropped by a factor of $10$ at $60,000$ and then at $80,000$ iterations. For methods requiring dropout, the dropout rate of $0.1$ is selected. We find out that using higher dropout rates resulted in up to $10\%$ drop in mAP when compared to the original RetinaNet model.

During inference, we use an NMS threshold of $0.5$ for methods requiring NMS. For methods requiring data association, we use the Basic Sequential Algorithmic Scheme (BSAS) with a spatial affinity threshold of an IOU of $0.9$ as recommended in~\cite{miller2019evaluating}. For methods requiring dropout, we perform $10$ stochastic dropout runs, as suggested by~\cite{kendall2017uncertainties}. Since the deep ensemble method is computationally very expensive, we use an ensemble of $5$ independently trained but identical models, the minimum number recommended in~\cite{lakshminarayanan2017simple}.
\subsection{Experimental Results and Discussions}\label{sec:comparative_study:experimental_results}
This section shows the key experimental results and findings from our proposed probabilistic object detectors. We first summarize the evaluation results in scenarios with and without dataset shift, and then list our findings with in-depth discussions.

\subsection*{General Results and Evaluation Metrics}
Tab.~\ref{table:bdd_all_classes} presents the results of our comparative experiments on BDD validation data split without dataset shift, where metrics are averaged over the seven dynamic object categories of interest. We quantify the performance of probabilistic object detection using the mean Average Precision (mAP) and Probability Detection Quality (PDQ) metrics introduced in Sec.~\ref{subsec:evaluation_metrics_pods}, as well as the Negative Log Likelihood (NLL) estimates in Sec.~\ref{uncertainty_estimation:metrics}. NLL is evaluated for both the estimated category and bounding box distributions of true positive detections, which have $\text{IOU}>0.7$ scores with ground truths in the scene. This IOU threshold of $0.7$ was chosen following the standard threshold set by the KITTI and BDD datasets when evaluating 2D object detection. Tab.~\ref{table:bdd_all_classes} also shows the runtime of each probabilistic object detection method on a machine with an Nvidia Titan V GPU and Intel i3770k CPU, in order to show the potential for deploying those methods on autonomous vehicles.

The first row of Tab.~\ref{table:bdd_all_classes} shows that our deterministic RetinaNet baseline achieves $28.62\%$ mAP, an acceptable value that conforms to the $28\%-30\%$ mAP range reported by baselines provided in the original BDD dataset paper~\cite{yu2020bdd100k}. We notice that all implemented probabilistic extensions to RetinaNet are less than $2\%$ mAP difference compared to the deterministic baseline. On the other hand, Tab.~\ref{table:comparison_studies_ood} presents the mAP of the implemented probabilistic object detectors under dataset shift, evaluated only on the Car and Person categories on the BDD validation, KITTI, and Lyft datasets. Tab.~\ref{table:comparison_studies_ood} shows that when we only consider the Car and Person categories for our evaluation, all detection methods show an increase of mAP values from $\sim 28\%$ to $\sim 40\%$ on the BDD dataset, because detecting two object categories is easier than seven categories. Compared to the mAP performance on the BDD dataset, a slight drop of $2-3\%$ mAP is observed when evaluating on the KITTI dataset, as well as a substantial drop of over $30\%$ mAP on the Lyft dataset. This is because the domain shift from BDD to Lyft data is much stronger than from BDD to KITTI data, as discussed in Sec.~\ref{sec:comparative_study:experimental_setup}.

\textbf{Tab.~\ref{table:bdd_all_classes} shows mAP to be correlated to the classification Negative Log Likelihood (NLL)}. As an example, the method that achieves the lowest mAP in Tab.~\ref{table:bdd_all_classes}, Output Redundancy, is seen to have the highest classification NLL and as such the worst classification performance.
\textbf{On the other hand, Tab.~\ref{table:bdd_all_classes} shows mAP to have no correlation to the regression NLL}. As an example, Tab.~\ref{table:bdd_all_classes} shows that Black Box has a regression NLL that is two orders of magnitude larger than other probabilistic detectors, while still performing on par with other probabilistic object detectors on mAP. Unlike mAP, we notice a direct correlation between PDQ and regression NLL, where the lowest PDQ is achieved by methods with the highest regression NLL (Black Box and Output Redundancy), while the highest PDQ is achieved by the methods with the lowest regression NLL. \textbf{However, the magnitude of increase in PDQ between probabilistic object detectors is not proportional to the magnitude of decrease in regression NLL.} As an example, Tab.~\ref{table:bdd_all_classes} shows that a reduction of $384.85$ in regression NLL when comparing Output Redundancy to Black Box translates into an increase of $0.18$ in PDQ.

\begin{figure*}[!tpb]
	\centering
	\includegraphics[width=\textwidth,trim=2 2 2 2,clip]{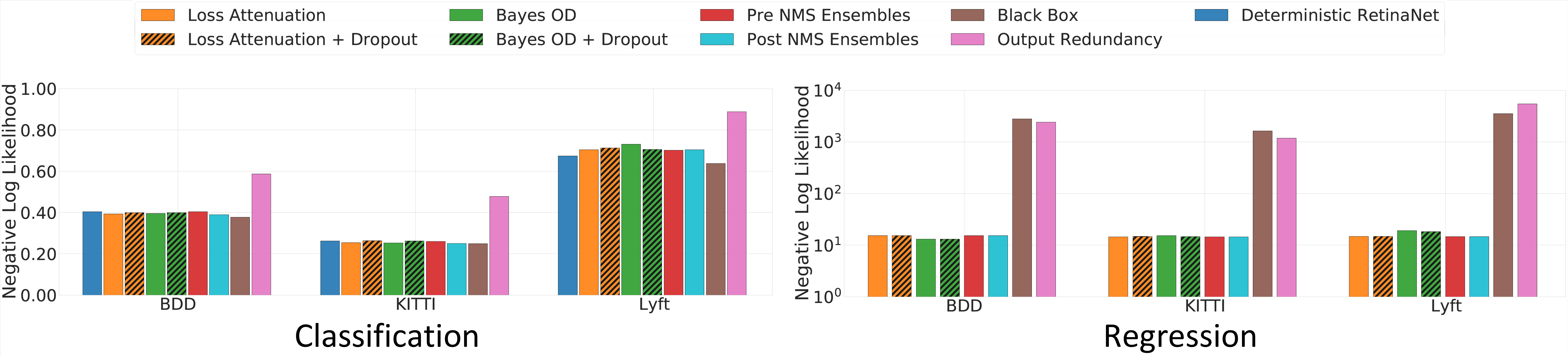}
	\caption{Bar plots of the \textbf{Negative Log Likelihood} of the category classification and bounding box regression probabilistic output for various probabilistic detection methods implemented with the RetinaNet architecture as a backend.}\label{fig:nll_results}
\end{figure*}

\begin{figure*}[!htpb]
	\centering
	\includegraphics[width=\textwidth,trim=2 2 2 2,clip]{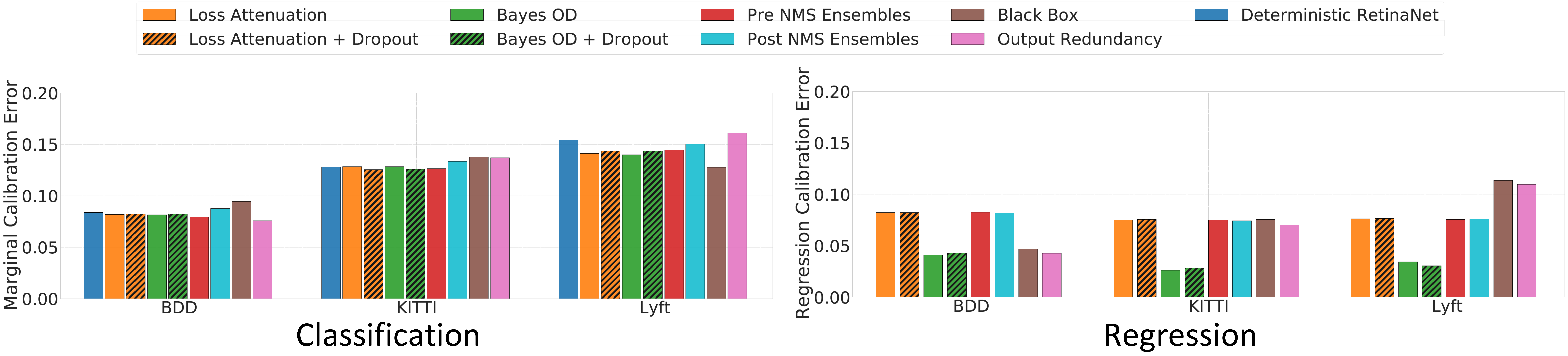}
	\caption{Bar plots of the \textbf{calibration errors} of the category classification and bounding box regression probabilistic output for various probabilistic detection methods implemented with the RetinaNet architecture as a backend.}\label{fig:calibration_results}
\end{figure*}

\begin{figure*}[!htpb]
	\centering
	\includegraphics[width=\textwidth,trim=2 2 2 2,clip]{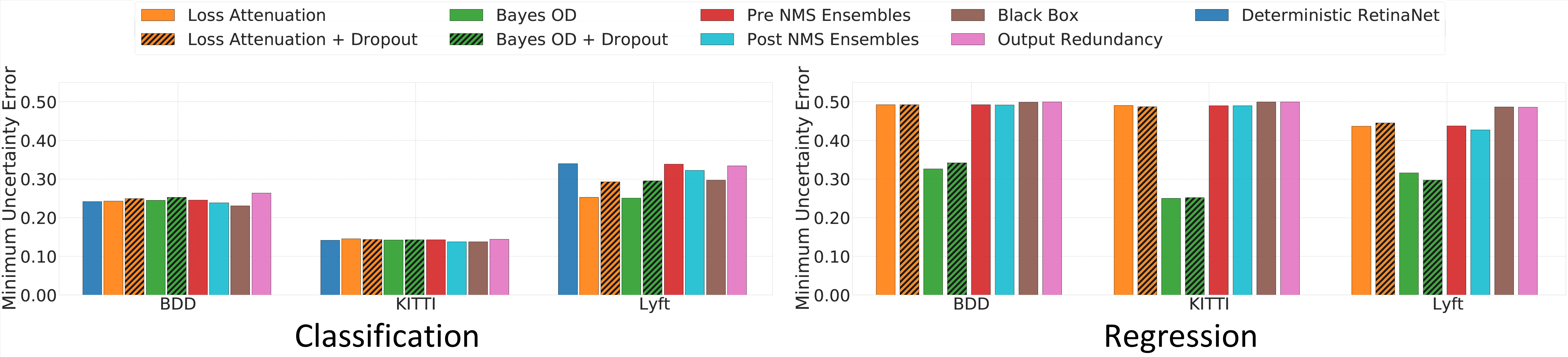}
	\caption{Bar plots of the \textbf{minimum uncertainty errors} for the category classification and bounding box regression probabilistic output for various probabilistic detection methods implemented with the RetinaNet architecture as a backend.}\label{fig:mue_results}
\end{figure*}

Fig.~\ref{fig:nll_results}, Fig.~\ref{fig:calibration_results}, and Fig.~\ref{fig:mue_results} show bar plots of the regression and classification NLL, calibration errors, and minimum uncertainty errors~\cite{harakeh2019bayesod} respectively. Those uncertainty estimation values are averaged over the Car and Person categories from the BDD validation, KITTI, and Lyft datasets. For evaluating calibration performance, we use the Marginal Calibration Error (MCE) metric for category classification and Expected Calibration Error (ECE) for bounding box regression (both metrics have been introduced in Sec.~\ref{uncertainty_estimation:metrics:calibration_plot}). Regarding on the uncertainty estimation in classification, Fig.~\ref{fig:nll_results} (left) shows that \textbf{all probabilistic object detectors exhibit a lower classification NLL scores on KITTI, but higher classification NLL scores on Lyft when compared to those on BDD}. Even though we train our probabilistic object detectors using BDD data, probabilistic object detectors are seen to have lower NLL scores on KITTI than on the BDD validation dataset, implying higher confidence in correct detections. However, Fig.~\ref{fig:calibration_results} (left) shows that the classification marginal calibration error increases to similar values on both KITTI and Lyft when compared to the marginal calibration error on the BDD validation set. \textbf{We conclude that although our probabilistic object detectors provide sharper categorical distributions on KITTI compared to BDD, the distributions of KITTI detections are worse calibrated than those of BDD detections.} Fig.~\ref{fig:mue_results} (left) also shows that the classification minimum uncertainty error follows the same trend as the classification Negative Log Likelihood in Fig.~\ref{fig:nll_results} (left) for all evaluated probabilistic object detectors. Regarding on the uncertainty estimation in regression, the right plots from Fig.~\ref{fig:nll_results}, Fig.~\ref{fig:calibration_results}, and Fig.~\ref{fig:mue_results} show that \textbf{regression metrics have little difference under dataset shift}. The reason for this phenomenon might be that we only evaluate well-localized regression outputs, where regressed bounding boxes are required to achieve an $\text{IOU} \geq 0.7$ with ground truth boxes.

Finally, looking at the \textbf{runtime} of each probabilistic object detector in the last column of Tab.~\ref{table:bdd_all_classes}, {we notice that all probabilistic extensions necessarily introduce a drop in frame rate when compared to deterministic RetinaNet}. However, the drop introduced by methods that estimate aleatoric uncertainty such as Loss Attenuation is much lower than the drop introduced by parameter-sampling methods that estimate epistemic uncertainty (MC-Dropout and Deep Ensembles). 


\subsection*{Comparison of each method and discussions}
\noindent In the following, we compare each probabilistic object detection methods in detail, and highlight the most prominent takeaways from our comparative studies. \textbf{We would like to note that unless stated otherwise, conclusions made in this section are specific to our experimental setup, where we used a single-stage probabilistic object detector that is trained on autonomous driving datasets.} We leave a more general comparative study on 3D object detection and LiDAR point cloud as an interesting future work.

\textbf{1. Using mAP as the only evaluation metric is not sufficient for quantifying the quality of predicted uncertainty in probabilistic object detectors.} It is important to emphasize this point due to the widespread practice of using only mAP as an evaluation metric when comparing state-of-the-art probabilistic object detection (See Tab.~\ref{tab:summary_object_detection}). mAP is clearly not related to the quality of uncertainty estimates provided by probabilistic object detectors, a quality that can be quantitatively observed in Tab.~\ref{table:comparison_studies_ood}. In this table, the mAP scores among different probabilistic object detection methods are comparable, with only $2-3\%$ mAP differences. However, the uncertainty scores (such as regression NLL) vary significantly, indicating very different uncertainty estimation performance. Taking a closer look at the performance of BayesOD and Pre-NMS ensembles, the table shows that BayesOD outperforms Pre-NMS ensembles on regression NLL, classification NLL and PDQ, but under-performs the same method in mAP, resulting in different rankings of BayesOD and Pre-NMS ensembles. Since mAP is important for evaluating the performance of deterministic object detectors, we argue that a good probabilistic object detector should maintain competitive mAP when compared to deterministic counterparts. Therefore, we suggest that researchers use proper scoring rules such as NLL as well as calibration errors for comparing and ranking probabilistic object detectors.

\textbf{2. Estimating spatial uncertainty through Loss Attenuation is essential for high quality predictive distributions in our experiments.}
Tab.~\ref{table:bdd_all_classes} shows that Black Box and Output Redundancy are a have a $\sim50\%$ drop in PDQ scores when compared to the rest of the implemented methods. Tab.~\ref{table:bdd_all_classes} also shows that the regression NLL values of Black Box and Output Redundancy are two orders of magnitude larger than all other implemented probabilistic detectors. This observation is consistent when looking at results under dataset shift in Fig.~\ref{fig:nll_results}, where regression NLL values of Black Box and Output Redundancy are seen to be around $\sim 10^3$ on BDD, KITTI, and Lyft data while the remaining probabilistic detectors have a regression NLL values that is closer to $10^1$. In addition, Fig.~\ref{fig:calibration_results} shows that the regression calibration error of both Black Box and Output Redundancy is around $0.04$, on par with the best calibrated method, BayesOD, on the BDD dataset. However, the regression calibration error is seen to increase as the magnitude of the dataset shift increases, where Black Box and Output Redundancy become the methods with the highest regression calibration error on the Lyft dataset.

Tab.~\ref{tab:our_object_dets} shows that the common design decision in both Black Box and Output Redundancy is the lack of Direct Modeling of uncertainty in their detection head. Instead, both Black Box and Output Redundancy methods use redundant output detections to compute estimates of the sample covariance for regression output. Theoretically, a sample covariance matrix in would require $2$ \textit{independent} output boxes to be estimate, and much more if this estimation is to be reliable. In addition, sample covariance estimates are highly sensitive to outliers~\cite{huber2004robust}. Spatial clustering approaches used by the Black Box and Output Redundancy by definition need to sacrifice localization accuracy to obtain larger cluster sizes, increasing the probability of having outliers as cluster members. The Black Box method uses a low number of redundant boxes that are clustered after the non-maximum suppression (NMS) stage for spatial uncertainty estimation, which is seen in Table~\ref{table:bdd_all_classes} as a higher regression NLL than that of Output Redundancy. On the other hand, Output Redundancy uses a large number of redundant output boxes collected before NMS, resulting in a better regression NLL at the expense of a much worse classification NLL, because low scoring detections are included in averaging their classification scores. For these reasons, we recommend the Direct Modeling approach for explicitly estimating the uncertainty of regressed bounding boxes when designing probabilistic object detectors.

\textbf{3. Estimating classification variance using Loss Attenuation offers no improvement in the quality of predictive uncertainty over baseline classification from deterministic RetinaNet.}
Unlike the benefits offered by directly modeling the spatial uncertainty, we found little benefits from modeling the classification variance using the formulation proposed by Kendall~\etal~\cite{kendall2017uncertainties} (and introduced in Sec.~\ref{uncertainty_estimation:methods:direct_modelling}). Tab.~\ref{table:bdd_all_classes} shows that methods that explicitly estimate the variance of class logit vectors output by the network prior to the softmax output layer, such as Loss Attenuation, have the same classification NLL scores as the vanilla network softmax classification output distribution when tested with no dataset shift.

Under dataset shift, modeling the variance parameters for the logit vector results in mixed results when looking at NLL in Fig.~\ref{fig:nll_results}, where Loss Attenuation showing a lower NLL value on the KITTI dataset but higher NLL on the Lyft dataset when compared to deterministic RetinaNet. On the other hand, Fig.~\ref{fig:calibration_results} and \ref{fig:mue_results} show that Loss Attenuation does have a lower classification calibration error and a lower classification minimum uncertainty error (MUE) on all three datasets when compared to deterministic RetinaNet. It should be noted that the improvement in performance observed using Loss Attenuation under domain shift is not substantial enough to validate the modifications to deterministic RetinaNet required to enable Direct Modeling of classification uncertainty. As a conclusion, we find that Direct Modeling of the variance of the logit vectors does not offer improvement over learning with cross-entropy in the deterministic RetinaNet model.

\textbf{4. Bayesian inference provides better spatial uncertainty qualities when compared to standard NMS.}
This phenomenon has been previously shown in~\cite{harakeh2019bayesod}, which employs Bayesian inference to fuse information from cluster members instead of selecting the detection with the highest score in the standard NMS. Tab.~\ref{tab:our_object_dets} shows that BayesOD follows the same probabilistic detection architecture as Loss Attenuation, with the only difference that BayesOD replaces NMS with the Bayesian Fusion. Tab.~\ref{table:bdd_all_classes} shows that with this simple change, BayesOD gains $3\%$ in PDQ and reduces the regression NLL by $2.24$ points compared to Loss Attenuation. Similarly, Fig.~\ref{fig:calibration_results} shows that BayesOD has the lowest regression calibration error among all methods. Fig.~\ref{fig:mue_results} shows that BayesOD is the only method with a regression minimum uncertainty error substantially lower than 0.48, implying that the regression entropy output from BayesOD can be used to distinguish false positive detections from true positive detections. Unfortunately, the improvement in spatial uncertainty estimation performance comes at a cost to runtime. Table~\ref{table:bdd_all_classes} shows that replacing NMS with Bayesian Inference in BayesOD causes a FPS from $15.21$ to $9.75$ when compared to Loss Attenution, which corresponds to approx. $\unit[37]{ms}$ more inference time. We therefore recommend that Bayesian Inference as a replacement for standard NMS, if better calibrated bounding box distributions are desired and the cost of additional computational can be compromised.

\textbf{5. The performance gains in the quality of predictive uncertainty when using sampling-based epistemic uncertainty estimation methods does not justify their computational cost when compared to aleatoric uncertainty estimation methods.}
Tab.~\ref{table:bdd_all_classes} and Tab.~\ref{table:comparison_studies_ood} show Loss Attenuation and BayesOD to exhibit a mAP drop when MC-Dropout is added to model epistemic uncertainty. On the other hand, both Pre-NMS and Post-NMS variants of ensembles are seen to provide a slight improvement to mAP both for evaluation without dataset shift in Tab.~\ref{table:bdd_all_classes} and for evaluation with dataset shift in Tab.~\ref{table:comparison_studies_ood}. However, the difference of mAP is only within $2-3\%$ range. 

Fig.~\ref{fig:nll_results} shows that ensemble methods do not provide a substantial improvement in the quality of predicted uncertainty estimates compared to a single model, when evaluated using classification and the regression NLL. The same conclusion can be reached by observing calibration and MUE results in Fig.~\ref{fig:calibration_results} and Fig.~\ref{fig:mue_results}, respectively, where methods using ensembles exhibit similar performance to a single model trained with Loss Attenuation. On the other hand, Fig.~\ref{fig:nll_results}, Fig.~\ref{fig:calibration_results}, and Fig.~\ref{fig:mue_results} show Loss Attenuation + Dropout and BayesOD + Dropout models to have a worse performance when compared to models trained with no dropout on NLL, calibration errors, and MUE respectively.

The reason behind the lack of significant improvement from Ensembles can be explained by looking at the size of the training dataset. As Kendall \textit{et al.} suggested in~\cite{kendall2017uncertainties}, epistemic uncertainty can be explained away under large data situations such as our training on $\unit[70]{K}$ frames from the BDD dataset. MC-Dropout on the other hand has been shown to significantly under-perform ensembles and other methods on simple predictive uncertainty estimation tasks such as classification in \cite{snoek2019can, ashukha2020pitfalls}. These results from literature also translate to our results on probabilistic object detection, where Dropout enabled Loss-Attenuation and BayesOD models are shown to provide worse predictive uncertainty when compared to their Dropout-free counterparts. One final disadvantage of these sampling-based methods is how computationally expensive both MC-Dropout and ensemble-based methods are, with both methods having the slowest runtimes in Tab.~\ref{table:bdd_all_classes}. We therefore conclude that the current methods used for epistemic uncertainty estimation do not provide enough improvement in the quality of estimated uncertainty to justify their high computational cost.

\subsection{Aleatoric Uncertainty in Camera and LiDAR Perception}\label{sec:comparative_study:lidar_camera}

\begin{table}[t]
	\centering
	\resizebox{1\linewidth}{!}{
	\centering
	\begin{tabular}{c c c c c c }
		\multicolumn{1}{c}{} & \multicolumn{2}{c}{\textbf{Distance}} & \multicolumn{1}{c}{} & \multicolumn{2}{c}{\textbf{Occlusion}} \\ 
		\rowcolor{lightgray!35} \textbf{Sensors} & Reg. Unc. & Cls. Unc. & &  Reg. Unc. & Cls. Unc. \\ 
		\textbf{Image} & 0.090 & 0.223 & & \textbf{0.485} & 0.147  \\
		\textbf{LiDAR} & \textbf{0.551} & \textbf{0.510} & & 0.126 & \textbf{0.158} \\
	\end{tabular}}
	\caption{The Pearson Correlation Coefficients (PCC) between aleatoric uncertainties and detection distances and occlusions, produced by a 2D image object detector and a 3D LiDAR object detector.}
	\label{tab:lidar_camera_uncertainty_vs_distance}
\end{table}
\begin{figure}[t]
	\centering
	\subfigure[]{\label{fig:distance_vs_uncertainty}\includegraphics[width=0.55\linewidth]{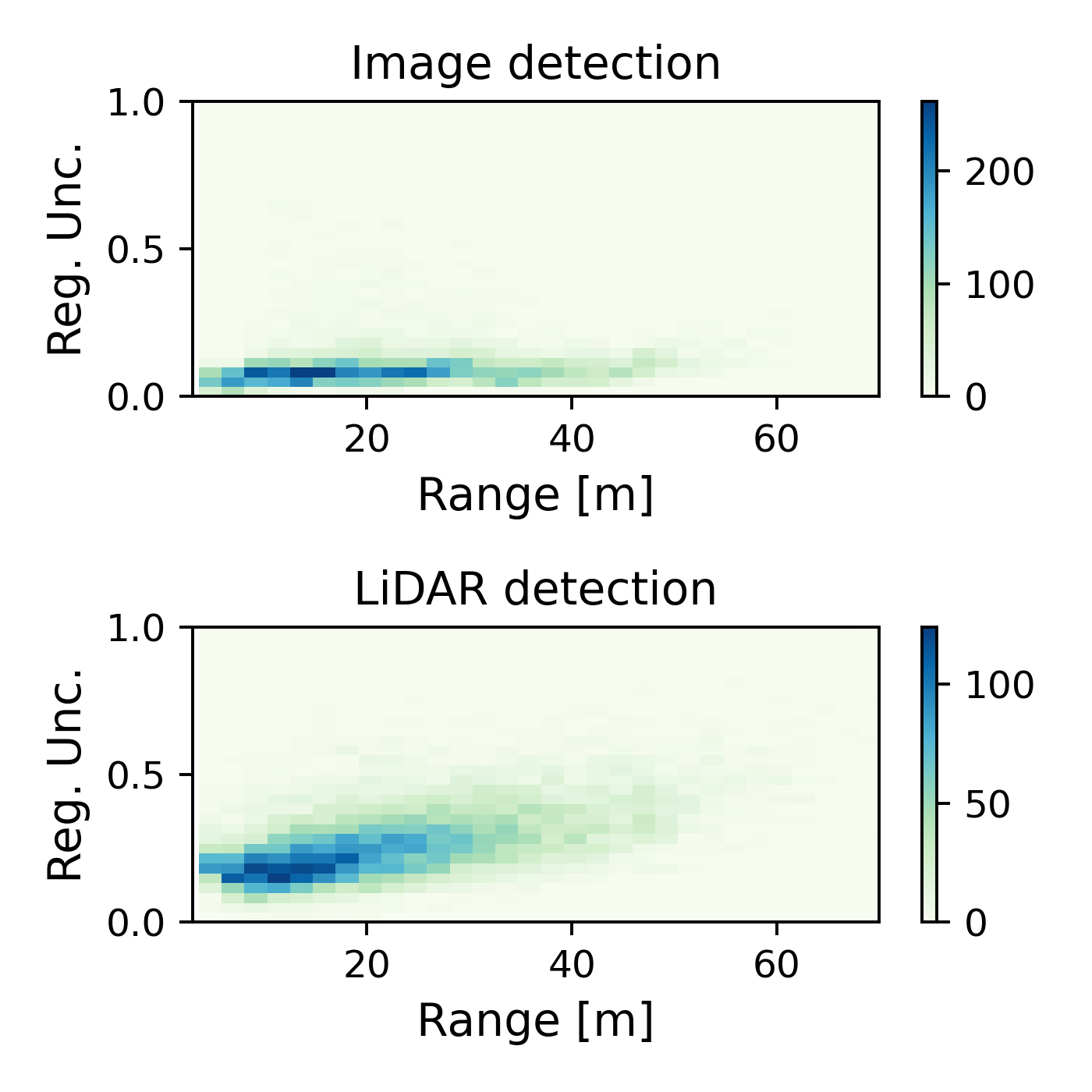}}
	\hfill
	\subfigure[]{\label{fig:lidar_camera_visualization}\includegraphics[width=0.43\linewidth]{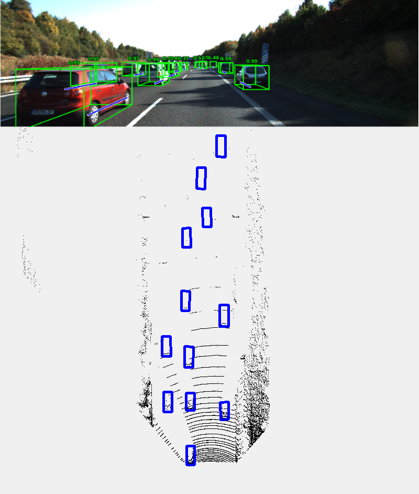}}
	\caption{(a). The heatmaps of regression aleatoric uncertainties in image detections and LiDAR detections. (b). An illustration of object detections in RGB camera images and LiDAR point clouds in the bird's eye view.}\label{fig:lidar_camera_uncertainty}
\end{figure}
RGB camera images and LiDAR point clouds are two dominant sensing modalities in autonomous driving. They have unique sensing properties and observation noise characteristics, which fall in the category of aleatoric uncertainty. In this section, we compare aleatoric uncertainty in image perception by common front-view RGB images, and LiDAR perception by Bird's Eye View representation. To do this, we conduct experiments on the ``Car'' objects in the KITTI dataset~\cite{Geiger2012CVPR}. We collect 2D image detections from the RetinaNet (Loss Attenuation) introduced in Sec.~\ref{sec:comparative_study:pod}, as well as their associated 3D LiDAR detections from ProbPIXOR~\cite{feng2019can} (cf. Tab.~\ref{tab:summary_object_detection}). Tab.~\ref{tab:lidar_camera_uncertainty_vs_distance} illustrates the correlations between aleatoric uncertainties in image or LiDAR perception and detection distance and occlusion, in terms of Pearson Correlation Coefficients (PCC). Defined in KITTI~\cite{Geiger2012CVPR}, occlusion is an ordered categorical value, with 0 indicating fully visible, 1 partly occluded, and 2 largely occluded. The aleatoric uncertaintes in regression and classification are measured by the Total Variance and Shannon Entropy metrics, respectively (cf. Sec.~\ref{uncertainty_estimation:metrics}). Note that the size of bounding boxes from image detection vary significantly (big at near range and small at far range). To have a fair comparison, we normalize the Total Variance values by the diagonal length of bounding boxes (both in LiDAR and image perception). Aleatoric uncertainties (both classification and regression) in LiDAR detections are more correlated with detection distance than image detections with much larger PCC values. Fig.~\ref{fig:distance_vs_uncertainty} supports this observation by plotting the distribution of regression uncertainties for all predictions (normalized to the same scale). Such uncertainty difference is due to the fact that the density of LiDAR point clouds are largely affected by distance, whereas the camera images do not directly provide the depth information (cf. Fig.~\ref{fig:lidar_camera_visualization}). Conversely, uncertainties in image detections (esp. in bounding box regression) are much more correlated with occlusion than those in LiDAR detections, because objects in images are often occluded and difficult to localize, while this is not the case in LiDAR point clouds (cf. Fig.~\ref{fig:lidar_camera_visualization}). Note that in this experiment we only show the difference between front-view image detection and BEV LiDAR detection. Note that there exist methods which project RGB images onto Bird's eye view~\cite{roddick2019orthographic}, or LiDAR points onto front-view for 3D detection~\cite{meyer2019lasernet}. We leave a comparative study for uncertainty modelling in these methods as an interesting future work. 
\section{\textbf{Challenges and Future Work}}\label{sec:challenges}
In this section, we discuss several challenges and directions for future work in probabilistic object detection.

\subsection{Efficient Epistemic Uncertainty Estimation}
Based on the papers reviewed, explicitly modelling epistemic uncertainty in probabilistic object detection usually requires a sampling mechanism such as MC-Dropout or Deep Ensembles, which are highly time-consuming and memory-intensive. For example, running MC-Dropout with $10$ inferences has a speed of $2-3$ FPS in Tab.~\ref{table:bdd_all_classes}, which is challenging for online autonomous driving. Recently, several works attempt to estimate epistemic uncertainty without sampling. For example, Postel \etal~\cite{postels2019sampling} propose to approximate epistemic uncertainty by an error propagation method. Furthermore, Sensoy \etal~\cite{sensoy2018evidential} and Amini \etal~\cite{amini2019deep} have shown that it is possible to directly predict the high-order conjugate priors of network output distributions (a Dirichlet prior for the multinomial distribution in classification~\cite{sensoy2018evidential}, and a Inverse-Gamma prior for the Gaussian distribution in regression), which are used to estimate epistemic uncertainty from single-shot inference based on evidential deep learning perspective. Incorporating Error Propagation in probabilistic object detectors would be an interesting direction of future work.

\subsection{Aleatoric Uncertainty Decomposition}
Aleatoric uncertainty, defined as the observation noise inherent in sensor measurements, can be decomposed into several sources. For example, aleatoric uncertainty arises from varied weather conditions (e.g. consider night-time, fog, or rain scenarios in camera perception), sensor precision and quality, as well as data annotation errors or ambiguity (e.g. labeling a car with a mountain bike on its roof only as "Car" class, instead of as "Car" and "Bike" objects). Czarnecki \etal~\cite{czarnecki2018towards} defined seven sources of uncertainty a perception module can model during the model development and operation (cf. Appx.~\ref{appendix:epistemic_aleatoric_uncertainty}. However, current probabilistic object detectors only model aleatoric uncertainty as a whole. It is an interesting future research direction to decompose aleatoric uncertainty and model each source of uncertainty separately. In this way, the uncertainty interpretability and detection performance may be potentially improved.

\subsection{Uncertainty Propagation to Downstream Modules}
The majority of works in probabilistic object detection focus on how to leverage uncertainty to improve the detection accuracy, especially in terms of Average Precision (AP), as we have discussed in Sec.~\ref{subsec:evaluation_metrics_pods}. However, object detection is only a building block in the long signal processing chain of the self-driving vehicle autonomy stack. Propagating the frame-by-frame detection uncertainties to the sequential downstream modules is expected to improve both the safety and robustness of the system as a whole. For example, multi-target tracking, an essential autonomous driving task, could benefit from estimated localization uncertainty of detected objects, which can be used to bootstrap object states in multi-target trackers based on Bayesian filtering ~\cite{fruhwirth2020towards}. Another task that could benefit from uncertainty estimates from object detection is the object-based simultaneous localization and mapping~\cite{nicholson2018quadricslam}, where the predicted mean and covariance matrices from bounding boxes can be used as measurement updates. Finally, using uncertainty estimates from probabilistic object detection for behavior prediction of multiple agents in road scenes~\cite{hu2018probabilistic} or for driver assistance system~\cite{liu2019itsc} could be another interesting directions. We encourage further research on methods to incorporate predicted uncertainty from probabilistic object detection into the various tasks required by autonomous vehicles.

\subsection{Better Evaluation of Probabilistic Object Detection}
We have shown in Sec.~\ref{sec:comparative_study} that mean Average Precision (mAP) does not consider the quality of bounding box uncertainty in comparing probabilistic object detectors. Two objects with the same mean but different covariance matrices for their output bounding boxes will yield the same mAP. Following the work in~\cite{harakeh2021estimating}, we recommend researchers to evaluate probabilistic object detection using proper scoring rules for a theoretically-sound evaluation as well as calibration errors for better insight on calibration quality of predicted probability distributions. In addition, PDQ can be useful for analyzing probabilistic object detectors, as it provides a summary of performance as a single scalar. One should note however that PDQ is \textbf{not a proper scoring rule}~\cite{harakeh2021estimating} and should not be used by itself to rank probabilistic object detectors.

Beyond proper scoring rules, we argue that probabilistic object detection should be evaluated in context of the full autonomous vehicle stack (i.e. from perception and prediction to planning and action). It is a difficult task that requires access to an autonomous vehicle platform and a complete software stacks~\cite{kato2018autoware,huang2019apolloscape} that support the integration of uncertainty estimates from probabilistic detectors in subsequent tasks. In this case, building autonomous vehicle simulators with high fidelity could help (e.g. \cite{manivasagam2020lidarsim,yang2020surfelgan}). It is an important and challenging future research direction to build such a simulator that could take the predictive uncertainty from probabilistic object detectors as inputs, and evaluate probabilistic detection in a full software stack with novel metrics.

\subsection{Ground Truth Uncertainty for Datasets}
Current open datasets in object detection (e.g. KITTI~\cite{Geiger2012CVPR} and Waymo~\cite{sun2019scalability}) only provide ground truth boxes and object categorizes, which can be used to train deterministic object detectors. However, those datasets do not provide information to reflect the ambiguity of human annotations (which can be categorized into aleatoric uncertainty), and thus no ground truth information is available that can directly supervise the training of predictive uncertainty in probabilistic object detection. Ground truth uncertainty can be generated during the manual annotation process of object detection datasets, by quantifying the disagreement between different human annotators on the category and the bounding box location of the same object in the scene. Unfortunately, generating accurate ground truth uncertainty will require multiple iterations of labeling, a process that is expensive and time consuming. Recently, \cite{wang2020inferring,feng2020labels2} make the first step to infer the label uncertainty with bounding boxes from LiDAR point clouds, with the help of a generative model. It would be an interesting direction to explore more alternative methods to generate ground truth uncertainty from a dataset, and benchmark probabilistic object detection. 
\section{\textbf{Conclusion}}\label{sec:conclusion}
In this work, we present a survey and comparative study on probabilistic object detection using deep learning approaches in the field of autonomous driving. First, we provide a background knowledge of practical uncertainty estimation methods in deep learning. Next, we systematically summarize existing methods in probabilistic object detection, and evaluate uncertainty estimation methods on an one-stage 2D image detector on three standard autonomous driving datasets. Finally, we discuss challenges and open questions in this research field.
Open source code for benchmarking probabilistic object detection is made available at: \url{https://github.com/asharakeh/pod_compare.git}. We hope that our survey and the benchmarking code will become a useful tool for researchers and practitioners in developing robust perception in autonomous driving.
\bibliographystyle{IEEEtran}{}
\bibliography{IEEEabrv, bibliography}

\begin{thebibliography}{100}
\providecommand{\url}[1]{#1}
\csname url@samestyle\endcsname
\providecommand{\newblock}{\relax}
\providecommand{\bibinfo}[2]{#2}
\providecommand{\BIBentrySTDinterwordspacing}{\spaceskip=0pt\relax}
\providecommand{\BIBentryALTinterwordstretchfactor}{4}
\providecommand{\BIBentryALTinterwordspacing}{\spaceskip=\fontdimen2\font plus
\BIBentryALTinterwordstretchfactor\fontdimen3\font minus
  \fontdimen4\font\relax}
\providecommand{\BIBforeignlanguage}[2]{{%
\expandafter\ifx\csname l@#1\endcsname\relax
\typeout{** WARNING: IEEEtran.bst: No hyphenation pattern has been}%
\typeout{** loaded for the language `#1'. Using the pattern for}%
\typeout{** the default language instead.}%
\else
\language=\csname l@#1\endcsname
\fi
#2}}
\providecommand{\BIBdecl}{\relax}
\BIBdecl

\bibitem{cosmides1996humans}
L.~Cosmides and J.~Tooby, ``Are humans good intuitive statisticians after all?
  rethinking some conclusions from the literature on judgment under
  uncertainty,'' \emph{Cognition}, vol.~58, no.~1, pp. 1--73, 1996.

\bibitem{janai2017computer}
J.~Janai, F.~G{\"u}ney, A.~Behl, A.~Geiger \emph{et~al.}, ``Computer vision for
  autonomous vehicles: Problems, datasets and state of the art,''
  \emph{Foundations and Trends in Computer Graphics and Vision}, vol.~12, no.
  1--3, pp. 1--308, 2020.

\bibitem{mcallister2017concrete}
R.~McAllister, Y.~Gal, A.~Kendall, M.~Van Der~Wilk, A.~Shah, R.~Cipolla, and
  A.~V. Weller, ``Concrete problems for autonomous vehicle safety: Advantages
  of bayesian deep learning,'' in \emph{Int. Joint Conf. on Artificial
  Intelligence}, 2017.

\bibitem{willers2020safety}
O.~Willers, S.~Sudholt, S.~Raafatnia, and S.~Abrecht, ``Safety concerns and
  mitigation approaches regarding the use of deep learning in safety-critical
  perception tasks,'' in \emph{International Conference on Computer Safety,
  Reliability, and Security}, 2020.

\bibitem{hall2020probability}
D.~Hall, F.~Dayoub, J.~Skinner, H.~Zhang, D.~Miller, P.~Corke, G.~Carneiro,
  A.~Angelova, and N.~S{\"u}nderhauf, ``Probabilistic object detection:
  Definition and evaluation,'' in \emph{The IEEE Winter Conference on
  Applications of Computer Vision}, 2020, pp. 1031--1040.

\bibitem{miller2018dropout}
D.~Miller, L.~Nicholson, F.~Dayoub, and N.~S{\"u}nderhauf, ``Dropout sampling
  for robust object detection in open-set conditions,'' in \emph{{IEEE} Int.
  Conf. Robotics and Automation}, 2018.

\bibitem{miller2019evaluating}
D.~Miller, F.~Dayoub, M.~Milford, and N.~S{\"u}nderhauf, ``Evaluating merging
  strategies for sampling-based uncertainty techniques in object detection,''
  in \emph{{IEEE} Int. Conf. Robotics and Automation}, 2019.

\bibitem{feng2018towards}
D.~Feng, L.~Rosenbaum, and K.~Dietmayer, ``Towards safe autonomous driving:
  Capture uncertainty in the deep neural network for lidar 3d vehicle
  detection,'' in \emph{{IEEE} Int. Conf. Intelligent Transp. Syst.}, 2018, pp.
  3266--3273.

\bibitem{feng2018leveraging}
D.~Feng, L.~Rosenbaum, F.~Timm, and K.~Dietmayer, ``Leveraging heteroscedastic
  aleatoric uncertainties for robust real-time lidar 3d object detection,'' in
  \emph{{IEEE} Intelligent Vehicles Symp.}, 2019.

\bibitem{feng2019deep}
D.~Feng, X.~Wei, L.~Rosenbaum, A.~Maki, and K.~Dietmayer, ``Deep active
  learning for efficient training of a lidar 3d object detector,'' in
  \emph{{IEEE} Intelligent Vehicles Symp.}, 2019.

\bibitem{feng2019can}
D.~Feng, L.~Rosenbaum, C.~Gl{\"a}ser, F.~Timm, and K.~Dietmayer, ``Can we trust
  you? on calibration of a probabilistic object detector for autonomous
  driving,'' \emph{{IEEE/RSJ} Int. Conf. Intelligent Robots and Systems}, 2020.

\bibitem{feng2020leveraging}
D.~Feng, Y.~Cao, L.~Rosenbaum, F.~Timm, and K.~Dietmayer, ``Leveraging
  uncertainties for deep multi-modal object detection in autonomous driving,''
  \emph{{IEEE} Intelligent Vehicles Symp.}, 2020.

\bibitem{harakeh2019bayesod}
A.~Harakeh, M.~Smart, and S.~L. Waslander, ``Bayesod: A bayesian approach for
  uncertainty estimation in deep object detectors,'' in \emph{{IEEE} Int. Conf.
  Robotics and Automation}, 2020.

\bibitem{he2019bounding}
Y.~He, C.~Zhu, J.~Wang, M.~Savvides, and X.~Zhang, ``Bounding box regression
  with uncertainty for accurate object detection,'' in \emph{Proc. {IEEE} Conf.
  Computer Vision and Pattern Recognition}, 2019.

\bibitem{meyer2019lasernet}
G.~P. Meyer, A.~Laddha, E.~Kee, C.~Vallespi-Gonzalez, and C.~K. Wellington,
  ``{LaserNet}: An efficient probabilistic 3d object detector for autonomous
  driving,'' in \emph{Proc. {IEEE} Conf. Computer Vision and Pattern
  Recognition}, 2019.

\bibitem{meyer2019learning}
G.~P. Meyer and N.~Thakurdesai, ``Learning an uncertainty-aware object detector
  for autonomous driving,'' \emph{arXiv preprint arXiv:1910.11375}, 2019.

\bibitem{wirges2019capturing}
S.~Wirges, M.~Reith-Braun, M.~Lauer, and C.~Stiller, ``Capturing object
  detection uncertainty in multi-layer grid maps,'' in \emph{{IEEE} Intelligent
  Vehicles Symp.}, 2019.

\bibitem{yu2020bdd100k}
F.~Yu, H.~Chen, X.~Wang, W.~Xian, Y.~Chen, F.~Liu, V.~Madhavan, and T.~Darrell,
  ``Bdd100k: A diverse driving dataset for heterogeneous multitask learning,''
  in \emph{Proc. {IEEE} Conf. Computer Vision and Pattern Recognition}, 2020.

\bibitem{lin2018focal}
T.-Y. Lin, P.~Goyal, R.~Girshick, K.~He, and P.~Doll{\'a}r, ``Focal loss for
  dense object detection,'' \emph{{IEEE} Trans. Pattern Anal. Mach. Intell.},
  2018.

\bibitem{Geiger2012CVPR}
A.~Geiger, P.~Lenz, and R.~Urtasun, ``Are we ready for autonomous driving? the
  {KITTI} vision benchmark suite,'' in \emph{Proc. {IEEE} Conf. Computer Vision
  and Pattern Recognition}, 2012.

\bibitem{kesten2019lyft}
R.~Kesten, M.~Usman, J.~Houston, T.~Pandya, K.~Nadhamuni, A.~Ferreira, M.~Yuan,
  B.~Low, A.~Jain, P.~Ondruska \emph{et~al.}, ``{Lyft Level 5 AV Dataset
  2019},'' \emph{https://level5.lyft.com/dataset}, 2019.

\bibitem{mackay1992practical}
D.~J.~C. MacKay, ``A practical {B}ayesian framework for backpropagation
  networks,'' \emph{Neural Computation}, vol.~4, no.~3, pp. 448--472, 1992.

\bibitem{neal1995bayesian}
R.~M. Neal, ``Bayesian learning for neural networks,'' Ph.D. dissertation,
  University of Toronto, 1995.

\bibitem{sensoy2018evidential}
M.~Sensoy, L.~Kaplan, and M.~Kandemir, ``Evidential deep learning to quantify
  classification uncertainty,'' in \emph{Advances in Neural Information
  Processing Systems}, 2018.

\bibitem{blundell2015weight}
C.~Blundell, J.~Cornebise, K.~Kavukcuoglu, and D.~Wierstra, ``Weight
  uncertainty in neural networks,'' in \emph{Int. Conf. Machine Learning},
  2015.

\bibitem{graves2011practical}
A.~Graves, ``Practical variational inference for neural networks,'' in
  \emph{Advances in Neural Information Processing Systems}, 2011.

\bibitem{depeweg2018decomposition}
S.~Depeweg, J.~Hernandez-Lobato, F.~Doshi-Velez, and S.~Udluft, ``Decomposition
  of uncertainty in bayesian deep learning for efficient and risk-sensitive
  learning,'' in \emph{Int. Conf. Machine Learning}, 2018.

\bibitem{neal2012bayesian}
R.~M. Neal, \emph{Bayesian learning for neural networks}.\hskip 1em plus 0.5em
  minus 0.4em\relax Springer Science \& Business Media, 2012, vol. 118.

\bibitem{welling2011bayesian}
M.~Welling and Y.~W. Teh, ``Bayesian learning via stochastic gradient langevin
  dynamics,'' in \emph{Int. Conf. Machine Learning}, 2011.

\bibitem{chen2014stochastic}
T.~Chen, E.~Fox, and C.~Guestrin, ``Stochastic gradient hamiltonian monte
  carlo,'' in \emph{Int. Conf. Machine Learning}, 2014, pp. 1683--1691.

\bibitem{chen2020statistical}
X.~Chen, J.~D. Lee, X.~T. Tong, Y.~Zhang \emph{et~al.}, ``Statistical inference
  for model parameters in stochastic gradient descent,'' \emph{The Annals of
  Statistics}, vol.~48, no.~1, pp. 251--273, 2020.

\bibitem{mandt2017stochastic}
S.~Mandt, M.~D. Hoffman, and D.~M. Blei, ``Stochastic gradient descent as
  approximate bayesian inference,'' \emph{The Journal of Machine Learning
  Research}, vol.~18, no.~1, pp. 4873--4907, 2017.

\bibitem{maddox2019simple}
W.~J. Maddox, P.~Izmailov, T.~Garipov, D.~P. Vetrov, and A.~G. Wilson, ``A
  simple baseline for bayesian uncertainty in deep learning,'' in
  \emph{Advances in Neural Information Processing Systems}, 2019.

\bibitem{ritter2018scalable}
H.~Ritter, A.~Botev, and D.~Barber, ``A scalable laplace approximation for
  neural networks,'' in \emph{6th International Conference on Learning
  Representations (ICLR)}, 2018.

\bibitem{kendall2017uncertainties}
A.~Kendall and Y.~Gal, ``What uncertainties do we need in {Bayesian} deep
  learning for computer vision?'' in \emph{Advances in Neural Information
  Processing Systems}, 2017.

\bibitem{kendall2015bayesian}
A.~Kendall, V.~Badrinarayanan, and R.~Cipolla, ``Bayesian {SegNet}: Model
  uncertainty in deep convolutional encoder-decoder architectures for scene
  understanding,'' in \emph{Proc. British Machine Vision Conf.}, 2017.

\bibitem{postels2019sampling}
J.~Postels, F.~Ferroni, H.~Coskun, N.~Navab, and F.~Tombari, ``Sampling-free
  epistemic uncertainty estimation using approximated variance propagation,''
  in \emph{Proc. {IEEE} Conf. Computer Vision}, 2019.

\bibitem{cortinhal2020salsanext}
T.~Cortinhal, G.~Tzelepis, and E.~E. Aksoy, ``Salsanext: Fast semantic
  segmentation of lidar point clouds for autonomous driving,'' \emph{arXiv
  preprint arXiv:2003.03653}, 2020.

\bibitem{ilg2018uncertainty}
E.~Ilg, O.~Ci{\c{c}}ek, S.~Galesso, A.~Klein, O.~Makansi, F.~Hutter, and
  T.~Brox, ``Uncertainty estimates and multi-hypotheses networks for optical
  flow,'' in \emph{Proc. Eur. Conf. Computer Vision}, 2018.

\bibitem{gast2018lightweight}
J.~Gast and S.~Roth, ``Lightweight probabilistic deep networks,'' in
  \emph{Proc. {IEEE} Conf. Computer Vision and Pattern Recognition}, 2018.

\bibitem{walz2020uncertainty}
S.~Walz, T.~Gruber, W.~Ritter, and K.~Dietmayer, ``Uncertainty depth estimation
  with gated images for 3d reconstruction,'' \emph{arXiv preprint
  arXiv:2003.05122}, 2020.

\bibitem{gustafsson2020evaluating}
F.~K. Gustafsson, M.~Danelljan, and T.~B. Schon, ``Evaluating scalable bayesian
  deep learning methods for robust computer vision,'' in \emph{CVPR Workshops},
  2020.

\bibitem{eldesokey2020uncertainty}
A.~Eldesokey, M.~Felsberg, K.~Holmquist, and M.~Persson, ``Uncertainty-aware
  cnns for depth completion: Uncertainty from beginning to end,'' in
  \emph{Proc. {IEEE} Conf. Computer Vision and Pattern Recognition}, 2020.

\bibitem{kendall2016modelling}
A.~Kendall and R.~Cipolla, ``Modelling uncertainty in deep learning for camera
  relocalization,'' in \emph{{IEEE} Int. Conf. Robotics and Automation}, 2016.

\bibitem{yang2020d3vo}
N.~Yang, L.~v. Stumberg, R.~Wang, and D.~Cremers, ``D3vo: Deep depth, deep pose
  and deep uncertainty for monocular visual odometry,'' in \emph{Proc. {IEEE}
  Conf. Computer Vision and Pattern Recognition}, 2020.

\bibitem{costante2020uncertainty}
G.~Costante and M.~Mancini, ``Uncertainty estimation for data-driven visual
  odometry,'' \emph{IEEE Transactions on Robotics}, 2020.

\bibitem{Beluch_2018_CVPR}
W.~H. Beluch, T.~Genewein, A.~N\"urnberger, and J.~M. K\"ohler, ``The power of
  ensembles for active learning in image classification,'' in \emph{Proc.
  {IEEE} Conf. Computer Vision and Pattern Recognition}, Jun. 2018.

\bibitem{mackowiak2018cereals}
R.~Mackowiak, P.~Lenz, O.~Ghori, F.~Diego, O.~Lange, and C.~Rother,
  ``Cereals-cost-effective region-based active learning for semantic
  segmentation,'' in \emph{Proc. British Machine Vision Conf.}, 2018.

\bibitem{haussmann2020scalable}
E.~Haussmann, M.~Fenzi, K.~Chitta, J.~Ivanecky, H.~Xu, D.~Roy, A.~Mittel,
  N.~Koumchatzky, C.~Farabet, and J.~M. Alvarez, ``Scalable active learning for
  object detection,'' in \emph{{IEEE} Intelligent Vehicles Symp.}, 2020.

\bibitem{chitta2018large}
K.~Chitta, J.~M. Alvarez, and A.~Lesnikowski, ``Large-scale visual active
  learning with deep probabilistic ensembles,'' \emph{arXiv preprint
  arXiv:1811.03575}, 2018.

\bibitem{danelljan2020probabilistic}
M.~Danelljan, L.~V. Gool, and R.~Timofte, ``Probabilistic regression for visual
  tracking,'' in \emph{Proc. {IEEE} Conf. Computer Vision and Pattern
  Recognition}, 2020.

\bibitem{hu2020probabilistic}
A.~Hu, F.~Cotter, N.~Mohan, C.~Gurau, and A.~Kendall, ``Probabilistic future
  prediction for video scene understanding,'' \emph{Proc. Eur. Conf. Computer
  Vision}, 2020.

\bibitem{meyer2020laserflow}
G.~P. Meyer, J.~Charland, S.~Pandey, A.~Laddha, C.~Vallespi-Gonzalez, and C.~K.
  Wellington, ``Laserflow: Efficient and probabilistic object detection and
  motion forecasting,'' \emph{arXiv preprint arXiv:2003.05982}, 2020.

\bibitem{hudnell2019robust}
M.~Hudnell, T.~Price, and J.-M. Frahm, ``Robust aleatoric modeling for future
  vehicle localization,'' in \emph{CVPR Workshops}, 2019.

\bibitem{makansi2019overcoming}
O.~Makansi, E.~Ilg, O.~Cicek, and T.~Brox, ``Overcoming limitations of mixture
  density networks: A sampling and fitting framework for multimodal future
  prediction,'' in \emph{Proc. {IEEE} Conf. Computer Vision and Pattern
  Recognition}, 2019.

\bibitem{segu2019general}
M.~Seg{\`u}, A.~Loquercio, and D.~Scaramuzza, ``A general framework for
  uncertainty estimation in deep learning,'' \emph{IEEE Robotics and Automation
  Letters}, vol.~5, no.~2, pp. 3153--3160, 2020.

\bibitem{michelmore2019uncertainty}
R.~Michelmore, M.~Wicker, L.~Laurenti, L.~Cardelli, Y.~Gal, and M.~Kwiatkowska,
  ``Uncertainty quantification with statistical guarantees in end-to-end
  autonomous driving control,'' in \emph{{IEEE} Int. Conf. Robotics and
  Automation}, 2020.

\bibitem{tai2019visual}
L.~Tai, P.~Yun, Y.~Chen, C.~Liu, H.~Ye, and M.~Liu, ``Visual-based autonomous
  driving deployment from a stochastic and uncertainty-aware perspective,''
  \emph{{IEEE/RSJ} Int. Conf. Intelligent Robots and Systems}, 2019.

\bibitem{czarnecki2018towards}
K.~Czarnecki and R.~Salay, ``Towards a framework to manage perceptual
  uncertainty for safe automated driving,'' in \emph{International Conference
  on Computer Safety, Reliability, and Security}.\hskip 1em plus 0.5em minus
  0.4em\relax Springer, 2018, pp. 439--445.

\bibitem{Gal2016Uncertainty}
Y.~Gal, ``Uncertainty in deep learning,'' Ph.D. dissertation, University of
  Cambridge, 2016.

\bibitem{gal2017concrete}
Y.~Gal, J.~Hron, and A.~Kendall, ``Concrete dropout,'' in \emph{Advances in
  Neural Information Processing Systems}, 2017.

\bibitem{blum2019fishyscapes}
H.~Blum, P.-E. Sarlin, J.~Nieto, R.~Siegwart, and C.~Cadena, ``Fishyscapes: A
  benchmark for safe semantic segmentation in autonomous driving,'' in
  \emph{CVPR Workshops}, 2019.

\bibitem{lakshminarayanan2017simple}
B.~Lakshminarayanan, A.~Pritzel, and C.~Blundell, ``Simple and scalable
  predictive uncertainty estimation using deep ensembles,'' in \emph{Advances
  in Neural Information Processing Systems}, 2017.

\bibitem{snoek2019can}
J.~Snoek, Y.~Ovadia, E.~Fertig, B.~Lakshminarayanan, S.~Nowozin, D.~Sculley,
  J.~Dillon, J.~Ren, and Z.~Nado, ``Can you trust your model's uncertainty?
  evaluating predictive uncertainty under dataset shift,'' in \emph{Advances in
  Neural Information Processing Systems}, 2019.

\bibitem{choi2018uncertainty}
S.~Choi, K.~Lee, S.~Lim, and S.~Oh, ``Uncertainty-aware learning from
  demonstration using mixture density networks with sampling-free variance
  modeling,'' in \emph{{IEEE} Int. Conf. Robotics and Automation}, 2018.

\bibitem{malinin2018predictive}
A.~Malinin and M.~Gales, ``Predictive uncertainty estimation via prior
  networks,'' in \emph{Advances in Neural Information Processing Systems},
  2018.

\bibitem{amini2019deep}
A.~Amini, W.~Schwarting, A.~Soleimany, and D.~Rus, ``Deep evidential
  regression,'' \emph{arXiv preprint arXiv:1910.02600}, 2019.

\bibitem{gustafsson2020energy}
F.~K. Gustafsson, M.~Danelljan, G.~Bhat, and T.~B. Sch{\"o}n, ``Energy-based
  models for deep probabilistic regression,'' in \emph{Proc. Eur. Conf.
  Computer Vision}, August 2020.

\bibitem{guo2017calibration}
C.~Guo, G.~Pleiss, Y.~Sun, and K.~Q. Weinberger, ``On calibration of modern
  neural networks,'' in \emph{Int. Conf. Machine Learning}, 2017.

\bibitem{kuleshov2018accurate}
V.~Kuleshov, N.~Fenner, and S.~Ermon, ``Accurate uncertainties for deep
  learning using calibrated regression,'' \emph{Thirty-fifth Interntional
  Conference on Machine Learning (ICML)}, 2018.

\bibitem{brier_1950}
G.~W. Brier, ``Verification of forecasts expressed in terms of probability,''
  \emph{Monthly weather review}, vol.~78, no.~1, pp. 1--3, 1950.

\bibitem{bernardo_1979}
J.~M. Bernardo, ``Expected information as expected utility,'' \emph{the Annals
  of Statistics}, pp. 686--690, 1979.

\bibitem{quinonero2005evaluating}
J.~Quinonero-Candela, C.~E. Rasmussen, F.~Sinz, O.~Bousquet, and
  B.~Sch{\"o}lkopf, ``Evaluating predictive uncertainty challenge,'' in
  \emph{Machine Learning Challenges Workshop}.\hskip 1em plus 0.5em minus
  0.4em\relax Springer, 2005, pp. 1--27.

\bibitem{kohonen2005lessons}
J.~Kohonen and J.~Suomela, ``Lessons learned in the challenge: making
  predictions and scoring them,'' in \emph{Machine Learning Challenges
  Workshop}.\hskip 1em plus 0.5em minus 0.4em\relax Springer, 2005, pp.
  95--116.

\bibitem{gneiting2007strictly}
T.~Gneiting and A.~E. Raftery, ``Strictly proper scoring rules, prediction, and
  estimation,'' \emph{Journal of the American statistical Association}, vol.
  102, no. 477, pp. 359--378, 2007.

\bibitem{kumar2019verified}
A.~Kumar, P.~S. Liang, and T.~Ma, ``Verified uncertainty calibration,'' in
  \emph{Advances in Neural Information Processing Systems}, 2019.

\bibitem{brocker2009reliability}
J.~Br{\"o}cker, ``Reliability, sufficiency, and the decomposition of proper
  scores,'' \emph{Quarterly Journal of the Royal Meteorological Society: A
  journal of the atmospheric sciences, applied meteorology and physical
  oceanography}, vol. 135, no. 643, pp. 1512--1519, 2009.

\bibitem{gneiting2007probabilistic}
T.~Gneiting, F.~Balabdaoui, and A.~E. Raftery, ``Probabilistic forecasts,
  calibration and sharpness,'' \emph{Journal of the Royal Statistical Society:
  Series B (Statistical Methodology)}, vol.~69, no.~2, pp. 243--268, 2007.

\bibitem{brocker2007properscoring}
J.~Br{\"o}cker and L.~A. Smith, ``Scoring probabilistic forecasts: The
  importance of being proper,'' \emph{Weather and Forecasting}, vol.~22, no.~2,
  pp. 382--388, 2007.

\bibitem{feng2019modal}
D.~Feng, C.~Haase-Schuetz, L.~Rosenbaum, H.~Hertlein, F.~Timm, C.~Glaeser,
  W.~Wiesbeck, and K.~Dietmayer, ``Deep multi-modal object detection and
  semantic segmentation for autonomous driving: Datasets, methods, and
  challenges,'' \emph{{IEEE} Trans. Intell. Transp. Syst.}, pp. 1--20, 2020.

\bibitem{kraus2019uncertainty}
F.~Kraus and K.~Dietmayer, ``Uncertainty estimation in one-stage object
  detection,'' in \emph{{IEEE} Int. Conf. Intelligent Transp. Syst.}, 2019.

\bibitem{miller2019benchmarking}
D.~Miller, N.~S{\"u}nderhauf, H.~Zhang, D.~Hall, and F.~Dayoub, ``Benchmarking
  sampling-based probabilistic object detectors.'' in \emph{CVPR Workshops},
  2019.

\bibitem{bertoni2019monoloco}
L.~Bertoni, S.~Kreiss, and A.~Alahi, ``Monoloco: Monocular 3d pedestrian
  localization and uncertainty estimation,'' in \emph{Proc. {IEEE} Conf.
  Computer Vision}, 2019.

\bibitem{simonyan2014very}
K.~Simonyan and A.~Zisserman, ``Very deep convolutional networks for
  large-scale image recognition,'' \emph{Third International Conference on
  Learning Representations (ICLR)}, 2015.

\bibitem{he2016deep}
K.~He, X.~Zhang, S.~Ren, and J.~Sun, ``Deep residual learning for image
  recognition,'' in \emph{Proc. {IEEE} Conf. Computer Vision and Pattern
  Recognition}, 2016.

\bibitem{szegedy2015going}
C.~Szegedy, W.~Liu, Y.~Jia, P.~Sermanet, S.~Reed, D.~Anguelov, D.~Erhan,
  V.~Vanhoucke, and A.~Rabinovich, ``Going deeper with convolutions,'' in
  \emph{Proc. {IEEE} Conf. Computer Vision and Pattern Recognition}, 2015.

\bibitem{le2018uncertainty}
M.~T. Le, F.~Diehl, T.~Brunner, and A.~Knol, ``Uncertainty estimation for deep
  neural object detectors in safety-critical applications,'' in \emph{{IEEE}
  Int. Conf. Intelligent Transp. Syst.}, 2018.

\bibitem{choi2019gaussian}
J.~Choi, D.~Chun, H.~Kim, and H.-J. Lee, ``Gaussian yolov3: An accurate and
  fast object detector using localization uncertainty for autonomous driving,''
  in \emph{Proc. {IEEE} Conf. Computer Vision}, 2019.

\bibitem{pan2020towards}
H.~Pan, Z.~Wang, W.~Zhan, and M.~Tomizuka, ``Towards better performance and
  more explainable uncertainty for 3d object detection of autonomous
  vehicles,'' in \emph{{IEEE} Int. Conf. Intelligent Transp. Syst.}, 2020.

\bibitem{wang2020inferring}
Z.~Wang, D.~Feng, Y.~Zhou, W.~Zhan, L.~Rosenbaum, F.~Timm, K.~Dietmayer, and
  M.~Tomizuka, ``Inferring spatial uncertainty in object detection,'' in
  \emph{{IEEE/RSJ} Int. Conf. Intelligent Robots and Systems}, 2020.

\bibitem{dong2020probabilistic}
X.~Dong, P.~Wang, P.~Zhang, and L.~Liu, ``Probabilistic oriented object
  detection in automotive radar,'' in \emph{CVPR Workshops}, 2020.

\bibitem{liu2016ssd}
W.~Liu, D.~Anguelov, D.~Erhan, C.~Szegedy, S.~Reed, C.-Y. Fu, and A.~C. Berg,
  ``{SSD}: Single shot multibox detector,'' in \emph{Proc. Eur. Conf. Computer
  Vision}, 2016.

\bibitem{ren2015faster}
S.~Ren, K.~He, R.~Girshick, and J.~Sun, ``{Faster R-CNN}: Towards real-time
  object detection with region proposal networks,'' in \emph{Advances in Neural
  Information Processing Systems}, 2015.

\bibitem{lee2020localization}
Y.~Lee, J.-w. Hwang, H.-I. Kim, K.~Yun, and J.~Park, ``Localization uncertainty
  estimation for anchor-free object detection,'' \emph{arXiv preprint
  arXiv:2006.15607}, 2020.

\bibitem{he2020deep}
Y.~He, Z.~Jianren~Wang, Wang, and K.~Jia, ``Deep mixture density network for
  probabilistic object detection,'' 2020.

\bibitem{feng2020labels}
D.~Feng, L.~Rosenbaum, F.~Timm, and K.~Dietmayer, ``Labels are not perfect:
  Improving probabilistic object detection via label uncertainty,'' in
  \emph{Eur. Conf. Computer Vision Workshops}, 2020.

\bibitem{wirges2018object}
S.~Wirges, T.~Fischer, C.~Stiller, and J.~B. Frias, ``Object detection and
  classification in occupancy grid maps using deep convolutional networks,'' in
  \emph{{IEEE} Int. Conf. Intelligent Transp. Syst.}, 2018.

\bibitem{lin2014microsoft}
T.-Y. Lin, M.~Maire, S.~Belongie, J.~Hays, P.~Perona, D.~Ramanan,
  P.~Doll{\'a}r, and C.~L. Zitnick, ``Microsoft {COCO}: Common objects in
  context,'' in \emph{Proc. Eur. Conf. Computer Vision}, 2014.

\bibitem{mccormac2017scenenet}
J.~McCormac, A.~Handa, S.~Leutenegger, and A.~J. Davison, ``Scenenet rgb-d: Can
  5m synthetic images beat generic imagenet pre-training on indoor
  segmentation?'' in \emph{Proc. {IEEE} Conf. Computer Vision}, 2017.

\bibitem{sunderhauf2016place}
N.~S{\"u}nderhauf, F.~Dayoub, S.~McMahon, B.~Talbot, R.~Schulz, P.~Corke,
  G.~Wyeth, B.~Upcroft, and M.~Milford, ``Place categorization and semantic
  mapping on a mobile robot,'' in \emph{{IEEE} Int. Conf. Robotics and
  Automation}.\hskip 1em plus 0.5em minus 0.4em\relax IEEE, 2016.

\bibitem{nuscenes2019}
H.~Caesar, V.~Bankiti, A.~H. Lang, S.~Vora, V.~E. Liong, Q.~Xu, A.~Krishnan,
  Y.~Pan, G.~Baldan, and O.~Beijbom, ``nu{S}cenes: A multimodal dataset for
  autonomous driving,'' \emph{Proc. {IEEE} Conf. Computer Vision and Pattern
  Recognition}, 2020.

\bibitem{everingham2010pascal}
M.~Everingham, L.~Van~Gool, C.~K. Williams, J.~Winn, and A.~Zisserman, ``The
  {PASCAL} visual object classes ({VOC}) challenge,'' \emph{Int. J. Computer
  Vision}, vol.~88, no.~2, pp. 303--338, 2010.

\bibitem{braun2019eurocity}
M.~Braun, S.~Krebs, F.~Flohr, and D.~M. Gavrila, ``Eurocity persons: A novel
  benchmark for person detection in traffic scenes,'' \emph{IEEE transactions
  on pattern analysis and machine intelligence}, vol.~41, no.~8, pp.
  1844--1861, 2019.

\bibitem{yang2018pixor}
B.~Yang, W.~Luo, and R.~Urtasun, ``{PIXOR}: Real-time 3d object detection from
  point clouds,'' in \emph{Proc. {IEEE} Conf. Computer Vision and Pattern
  Recognition}, 2018.

\bibitem{chen2020monopair}
Y.~Chen, L.~Tai, K.~Sun, and M.~Li, ``Monopair: Monocular 3d object detection
  using pairwise spatial relationships,'' \emph{Proc. {IEEE} Conf. Computer
  Vision and Pattern Recognition}, 2020.

\bibitem{bodla2017soft}
N.~Bodla, B.~Singh, R.~Chellappa, and L.~S. Davis, ``{Soft-NMS}-improving
  object detection with one line of code,'' in \emph{Proc. {IEEE} Conf.
  Computer Vision}, 2017.

\bibitem{shao2018crowdhuman}
S.~Shao, Z.~Zhao, B.~Li, T.~Xiao, G.~Yu, X.~Zhang, and J.~Sun, ``Crowdhuman: A
  benchmark for detecting human in a crowd,'' \emph{arXiv preprint
  arXiv:1805.00123}, 2018.

\bibitem{wang2017detecting}
J.~Wang, C.~Xie, Z.~Zhang, J.~Zhu, L.~Xie, and A.~Yuille, ``Detecting semantic
  parts on partially occluded objects,'' \emph{Proc. British Machine Vision
  Conf.}, 2017.

\bibitem{sun2019scalability}
P.~Sun, H.~Kretzschmar, X.~Dotiwalla, A.~Chouard, V.~Patnaik, P.~Tsui, J.~Guo,
  Y.~Zhou, Y.~Chai, B.~Caine, V.~Vasudevan, W.~Han, J.~Ngiam, H.~Zhao,
  A.~Timofeev, S.~Ettinger, M.~Krivokon, A.~Gao, A.~Joshi, Y.~Zhang, J.~Shlens,
  Z.~Chen, and D.~Anguelov, ``Scalability in perception for autonomous driving:
  Waymo open dataset,'' in \emph{Proc. {IEEE} Conf. Computer Vision and Pattern
  Recognition}, 2020, pp. 2446--2454.

\bibitem{feng2020labels2}
D.~Feng, Z.~Wang, Y.~Zhou, W.~Zhan, L.~Rosenbaum, F.~Timm, K.~Dietmayer, and
  M.~Tomizuka, ``Labels are not perfect: Inferring spatial uncertainty in
  object detection,'' \emph{arXiv preprint arXiv:1910.02600}, 2020.

\bibitem{harakeh2021estimating}
A.~Harakeh and S.~L. Waslander, ``Estimating and evaluating regression
  predictive uncertainty in deep object detectors,'' in \emph{International
  Conference on Learning Representations}, 2021.

\bibitem{Detectron2018}
R.~Girshick, I.~Radosavovic, G.~Gkioxari, P.~Doll\'{a}r, and K.~He,
  ``Detectron,'' \url{https://github.com/facebookresearch/detectron}, 2018.

\bibitem{huber2004robust}
P.~J. Huber, \emph{Robust statistics}.\hskip 1em plus 0.5em minus 0.4em\relax
  John Wiley \& Sons, 2004, vol. 523.

\bibitem{ashukha2020pitfalls}
A.~Ashukha, A.~Lyzhov, D.~Molchanov, and D.~Vetrov, ``Pitfalls of in-domain
  uncertainty estimation and ensembling in deep learning,'' \emph{Eighth
  International Conference on Learning Representations (ICLR)}, 2020.

\bibitem{roddick2019orthographic}
T.~Roddick, A.~Kendall, and R.~Cipolla, ``Orthographic feature transform for
  monocular 3d object detection,'' in \emph{Proc. British Machine Vision
  Conf.}, 2019.

\bibitem{fruhwirth2020towards}
C.~Fruhwirth-Reisinger, G.~Krispel, H.~Possegger, and H.~Bischof, ``Towards
  data-driven multi-target tracking for autonomous driving,'' in \emph{Computer
  Vision Winter Workshop (CVWW)}, 2020.

\bibitem{nicholson2018quadricslam}
L.~Nicholson, M.~Milford, and N.~S{\"u}nderhauf, ``Quadricslam: Dual quadrics
  from object detections as landmarks in object-oriented slam,'' \emph{IEEE
  Robotics and Automation Letters}, vol.~4, no.~1, pp. 1--8, 2018.

\bibitem{hu2018probabilistic}
Y.~Hu, W.~Zhan, and M.~Tomizuka, ``Probabilistic prediction of vehicle semantic
  intention and motion,'' in \emph{{IEEE} Intelligent Vehicles Symp.}\hskip 1em
  plus 0.5em minus 0.4em\relax IEEE, 2018.

\bibitem{liu2019itsc}
S.~{Liu}, K.~{Koch}, B.~{Gahr}, and F.~{Wortmann}, ``Brake maneuver prediction
  – an inference leveraging rnn focus on sensor confidence,'' in \emph{{IEEE}
  Int. Conf. Intelligent Transp. Syst.}, 2019, pp. 3249--3255.

\bibitem{kato2018autoware}
S.~Kato, S.~Tokunaga, Y.~Maruyama, S.~Maeda, M.~Hirabayashi, Y.~Kitsukawa,
  A.~Monrroy, T.~Ando, Y.~Fujii, and T.~Azumi, ``Autoware on board: Enabling
  autonomous vehicles with embedded systems,'' in \emph{2018 ACM/IEEE 9th
  International Conference on Cyber-Physical Systems (ICCPS)}.\hskip 1em plus
  0.5em minus 0.4em\relax IEEE, 2018, pp. 287--296.

\bibitem{huang2019apolloscape}
X.~Huang, P.~Wang, X.~Cheng, D.~Zhou, Q.~Geng, and R.~Yang, ``The apolloscape
  open dataset for autonomous driving and its application,'' \emph{IEEE
  transactions on pattern analysis and machine intelligence}, vol.~42, no.~10,
  pp. 2702--2719, 2019.

\bibitem{manivasagam2020lidarsim}
S.~Manivasagam, S.~Wang, K.~Wong, W.~Zeng, M.~Sazanovich, S.~Tan, B.~Yang,
  W.-C. Ma, and R.~Urtasun, ``Lidarsim: Realistic lidar simulation by
  leveraging the real world,'' in \emph{Proc. {IEEE} Conf. Computer Vision and
  Pattern Recognition}, 2020.

\bibitem{yang2020surfelgan}
Z.~Yang, Y.~Chai, D.~Anguelov, Y.~Zhou, P.~Sun, D.~Erhan, S.~Rafferty, and
  H.~Kretzschmar, ``Surfelgan: Synthesizing realistic sensor data for
  autonomous driving,'' in \emph{Proc. {IEEE} Conf. Computer Vision and Pattern
  Recognition}, 2020.

\bibitem{kendall2015posenet}
A.~Kendall, M.~Grimes, and R.~Cipolla, ``Posenet: A convolutional network for
  real-time 6-dof camera relocalization,'' in \emph{Proceedings of the IEEE
  international conference on computer vision}, 2015.

\bibitem{brostow2009semantic}
G.~J. Brostow, J.~Fauqueur, and R.~Cipolla, ``Semantic object classes in video:
  A high-definition ground truth database,'' \emph{Pattern Recognition
  Letters}, vol.~30, no.~2, pp. 88--97, 2009.

\bibitem{song2015sun}
S.~Song, S.~P. Lichtenberg, and J.~Xiao, ``{SUN RGB-D}: A {RGB-D} scene
  understanding benchmark suite,'' in \emph{Proceedings of the IEEE conference
  on computer vision and pattern recognition}, 2015.

\bibitem{cordts2016cityscapes}
M.~Cordts, M.~Omran, S.~Ramos, T.~Rehfeld, M.~Enzweiler, R.~Benenson,
  U.~Franke, S.~Roth, and B.~Schiele, ``The {Cityscapes} dataset for semantic
  urban scene understanding,'' in \emph{Proc. {IEEE} Conf. Computer Vision and
  Pattern Recognition}, 2016.

\bibitem{butler2012naturalistic}
D.~J. Butler, J.~Wulff, G.~B. Stanley, and M.~J. Black, ``A naturalistic open
  source movie for optical flow evaluation,'' in \emph{Proc. Eur. Conf.
  Computer Vision}, 2012.

\bibitem{colyar2007us}
J.~Colyar and J.~Halkias, ``Us highway 101 dataset,'' \emph{Federal Highway
  Administration (FHWA), Tech. Rep. FHWA-HRT-07-030}, 2007.

\bibitem{dosovitskiy2017carla}
A.~Dosovitskiy, G.~Ros, F.~Codevilla, A.~Lopez, and V.~Koltun, ``Carla: An open
  urban driving simulator,'' in \emph{Conference on Robot Learning}, 2017.

\bibitem{gruber2019gated2depth}
T.~Gruber, F.~Julca-Aguilar, M.~Bijelic, and F.~Heide, ``Gated2depth: Real-time
  dense lidar from gated images,'' in \emph{Proc. {IEEE} Conf. Computer
  Vision}, 2019.

\bibitem{wu2013online}
Y.~Wu, J.~Lim, and M.-H. Yang, ``Online object tracking: A benchmark,'' in
  \emph{Proc. {IEEE} Conf. Computer Vision and Pattern Recognition}, 2013, pp.
  2411--2418.

\bibitem{mueller2016benchmark}
M.~Mueller, N.~Smith, and B.~Ghanem, ``A benchmark and simulator for uav
  tracking,'' in \emph{Proc. Eur. Conf. Computer Vision}, 2016.

\bibitem{kiani2017need}
H.~Kiani~Galoogahi, A.~Fagg, C.~Huang, D.~Ramanan, and S.~Lucey, ``Need for
  speed: A benchmark for higher frame rate object tracking,'' in \emph{Proc.
  {IEEE} Conf. Computer Vision}, 2017.

\bibitem{deng2009imagenet}
J.~Deng, W.~Dong, R.~Socher, L.-J. Li, K.~Li, and L.~Fei-Fei, ``{ImageNet}: A
  large-scale hierarchical image database,'' in \emph{Proc. {IEEE} Conf.
  Computer Vision and Pattern Recognition}, 2009.

\bibitem{krizhevsky2009learning}
A.~Krizhevsky, ``Learning multiple layers of features from tiny images,''
  University of Toronto, Tech. Rep., 2009.

\bibitem{filos2019systematic}
A.~Filos, S.~Farquhar, A.~N. Gomez, T.~G. Rudner, Z.~Kenton, L.~Smith,
  M.~Alizadeh, A.~de~Kroon, and Y.~Gal, ``A systematic comparison of bayesian
  deep learning robustness in diabetic retinopathy tasks,'' in \emph{Workshop
  on Bayesian Deep Learning (NeurIPS 2019)}, 2019.

\bibitem{redmon2016you}
J.~Redmon, S.~Divvala, R.~Girshick, and A.~Farhadi, ``You only look once:
  Unified, real-time object detection,'' in \emph{Proc. {IEEE} Conf. Computer
  Vision and Pattern Recognition}, 2016.

\bibitem{asvadi2017depthcn}
A.~Asvadi, L.~Garrote, C.~Premebida, P.~Peixoto, and U.~J. Nunes, ``{DepthCN}:
  Vehicle detection using 3d-lidar and {ConvNet},'' in \emph{{IEEE} Int. Conf.
  Intelligent Transp. Syst.}, 2017.

\bibitem{engelcke2017vote3deep}
M.~Engelcke, D.~Rao, D.~Z. Wang, C.~H. Tong, and I.~Posner, ``{Vote3Deep}: Fast
  object detection in 3d point clouds using efficient convolutional neural
  networks,'' in \emph{{IEEE} Int. Conf. Robotics and Automation}, 2017.

\bibitem{tian2019fcos}
Z.~Tian, C.~Shen, H.~Chen, and T.~He, ``Fcos: Fully convolutional one-stage
  object detection,'' in \emph{Proc. {IEEE} Conf. Computer Vision}, 2019.

\bibitem{carion2020end}
N.~Carion, F.~Massa, G.~Synnaeve, N.~Usunier, A.~Kirillov, and S.~Zagoruyko,
  ``End-to-end object detection with transformers,'' \emph{Proc. Eur. Conf.
  Computer Vision}, 2020.

\bibitem{chen2017multi}
X.~Chen, H.~Ma, J.~Wan, B.~Li, and T.~Xia, ``Multi-view {3D} object detection
  network for autonomous driving,'' in \emph{Proc. {IEEE} Conf. Computer Vision
  and Pattern Recognition}, 2017.

\bibitem{ku2017joint}
J.~Ku, M.~Mozifian, J.~Lee, A.~Harakeh, and S.~Waslander, ``Joint 3d proposal
  generation and object detection from view aggregation,'' in \emph{{IEEE/RSJ}
  Int. Conf. Intelligent Robots and Systems}, 2018.

\bibitem{simon2018complex}
M.~Simon, S.~Milz, K.~Amende, and H.-M. Gross, ``Complex-yolo: An
  euler-region-proposal for real-time 3d object detection on point clouds,'' in
  \emph{Proc. Eur. Conf. Computer Vision}, 2018.

\bibitem{zhou2017voxelnet}
Y.~Zhou and O.~Tuzel, ``{VoxelNet}: End-to-end learning for point cloud based
  3d object detection,'' in \emph{Proc. {IEEE} Conf. Computer Vision and
  Pattern Recognition}, 2018.

\bibitem{shi2020pv}
S.~Shi, C.~Guo, L.~Jiang, Z.~Wang, J.~Shi, X.~Wang, and H.~Li, ``{PV-RCNN}:
  Point-voxel feature set abstraction for 3d object detection,'' in \emph{Proc.
  {IEEE} Conf. Computer Vision and Pattern Recognition}, 2020.

\bibitem{yan2018second}
Y.~Yan, Y.~Mao, and B.~Li, ``Second: Sparsely embedded convolutional
  detection,'' \emph{Sensors}, vol.~18, no.~10, p. 3337, 2018.

\bibitem{lang2018pointpillars}
A.~H. Lang, S.~Vora, H.~Caesar, L.~Zhou, J.~Yang, and O.~Beijbom,
  ``{PointPillars}: Fast encoders for object detection from point clouds,'' in
  \emph{Proc. {IEEE} Conf. Computer Vision and Pattern Recognition}, 2018.

\bibitem{fei2020semanticvoxels}
J.~Fei, W.~Chen, P.~Heidenreich, S.~Wirges, and C.~Stiller, ``Semanticvoxels:
  Sequential fusion for 3d pedestrian detection using lidar point cloud and
  semantic segmentation,'' in \emph{IEEE Int. Conf. Multisensor Fusion and
  Integration}, 2019.

\bibitem{qi2017pointnet}
C.~R. Qi, H.~Su, K.~Mo, and L.~J. Guibas, ``{PointNet}: Deep learning on point
  sets for 3d classification and segmentation,'' in \emph{Proc. {IEEE} Conf.
  Computer Vision and Pattern Recognition}, 2017.

\bibitem{xu2017pointfusion}
D.~Xu, D.~Anguelov, and A.~Jain, ``{PointFusion}: Deep sensor fusion for {3D}
  bounding box estimation,'' in \emph{Proc. {IEEE} Conf. Computer Vision and
  Pattern Recognition}, 2018.

\bibitem{shi2019pointrcnn}
S.~Shi, X.~Wang, and H.~Li, ``{Point-RCNN}: 3d object proposal generation and
  detection from point cloud,'' in \emph{Proc. {IEEE} Conf. Computer Vision and
  Pattern Recognition}, 2019.

\bibitem{liu2020deep}
L.~Liu, W.~Ouyang, X.~Wang, P.~Fieguth, J.~Chen, X.~Liu, and
  M.~Pietik{\"a}inen, ``Deep learning for generic object detection: A survey,''
  \emph{International journal of computer vision}, vol. 128, no.~2, pp.
  261--318, 2020.

\end{thebibliography}

\begin{IEEEbiography}[{\includegraphics[width=1in,height=1.25in,clip,keepaspectratio]{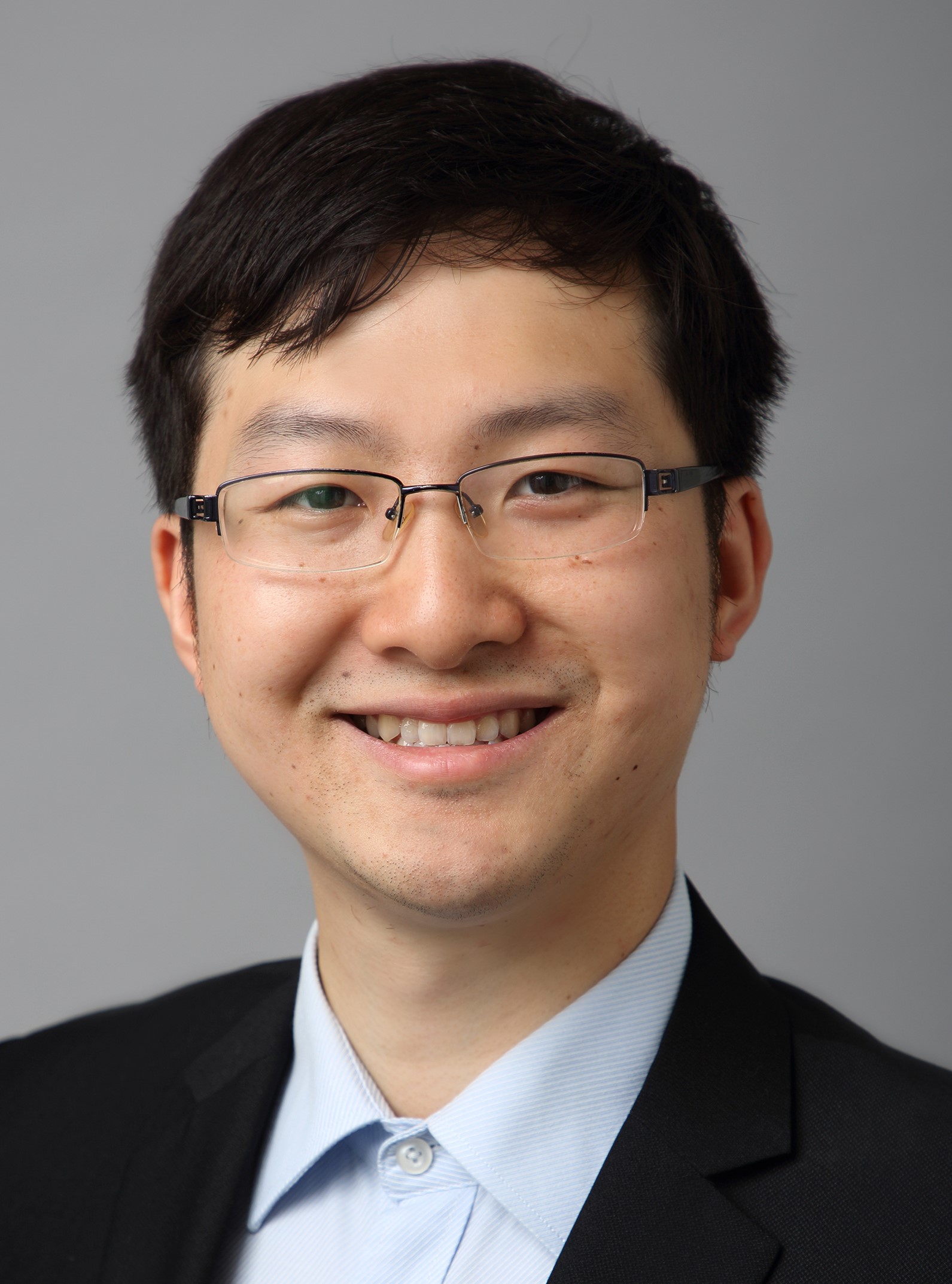}}]{Di Feng} (Member, IEEE) received the bachelor’s degree in mechatronics from the Tongji University, Shanghai, in 2014, the master’s degree in electrical and computer engineering from the Technical University of Munich, Germany, in 2017, and the Dr.-Ing. degree (equivalent to Ph.D.) from the Ulm University, Germany, in 2021, in cooperation with the Corporate Research of Robert Bosch GmbH. He was also a visiting researcher at the University of California, Berkeley, in 2020/2021. Currently, he is a research engineer at the Corporate Research of Robert Bosch GmbH, where he developed perception algorithms for autonomous driving functions. Research interests include environmental perception, multi-modal object detection, uncertainty estimation in deep learning, and machine learning in robotics and autonomous driving. 
\end{IEEEbiography}

\begin{IEEEbiography}[{\includegraphics[width=1in,height=1.25in,clip,keepaspectratio]{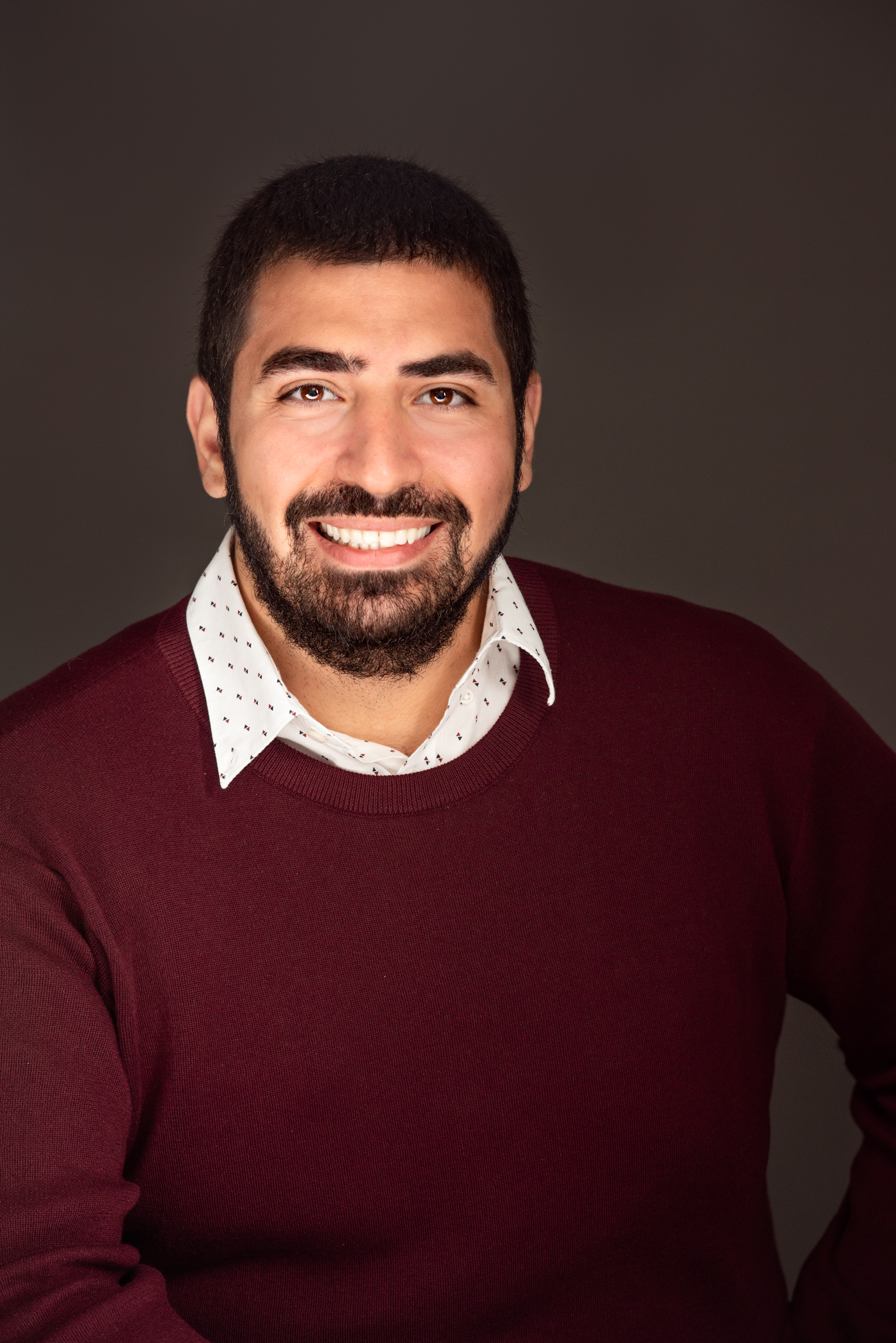}}]{Ali Harakeh}
(Member, IEEE) is currently Postdoctral Research Fellow at Mila - Quebec AI Institute. He earned his Ph.D. in 2021 at The University of Toronto Institute for Aerospace Studies (UTIAS), Toronto, Canada.  During his studies, he developed real-time perception algorithms that were part of autonomous driving live demonstrations in the Consumer Electronics Show (CES) and The Vehicular Technology Conference (VTC) in 2017. He also worked on developing cutting edge perception algorithms in cooperation with companies such as Huawei Technologies, Renesas Electronics, and LG Electronics. He received his master's degree in mechanical engineering from The American University of Beirut, Beirut, Lebanon in 2016. He currently specializes in perception algorithms for autonomous driving, working on fundamental research within 2D and 3D object detection, semantic segmentation, uncertainty estimation for deep neural networks, and synthetic data generation.
\end{IEEEbiography}

\begin{IEEEbiography}[{\includegraphics[width=1in,height=1.25in,clip,keepaspectratio]{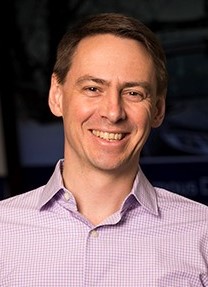}}]{Steven L. Waslander} (Senior Member, IEEE) is a leading authority on autonomous aerial and ground vehicles, including multirotor drones and autonomous driving vehicles, Simultaneous Localization and Mapping (SLAM) and multi-vehicle systems.  He received his B.Sc.E.in 1998 from Queen’s University, his M.S. in 2002 and his Ph.D. in 2007, both from Stanford University in Aeronautics and Astronautics. He joined the University of Waterloo in 2008, where he founded and directed the Waterloo Autonomous Vehicle Laboratory (WAVELab). In 2018, he joined the University of Toronto Institute for Aerospace Studies (UTIAS), and founded the Toronto Robotics and Artificial Intelligence Laboratory (TRAILab). Prof. Waslander’s innovations were recognized by the Ontario Centres of Excellence Mind to Market award for the best Industry/Academia collaboration (2012, with Aeryon Labs), as well as best paper and best poster awards at the Computer and Robot Vision Conference (2018).  His work on autonomous vehicles has resulted in the Autonomoose, the first autonomous vehicle created at a Canadian University to drive on public roads. 
\end{IEEEbiography}

\begin{IEEEbiography}
[{\includegraphics[width=1in,height=1.25in,clip,keepaspectratio]{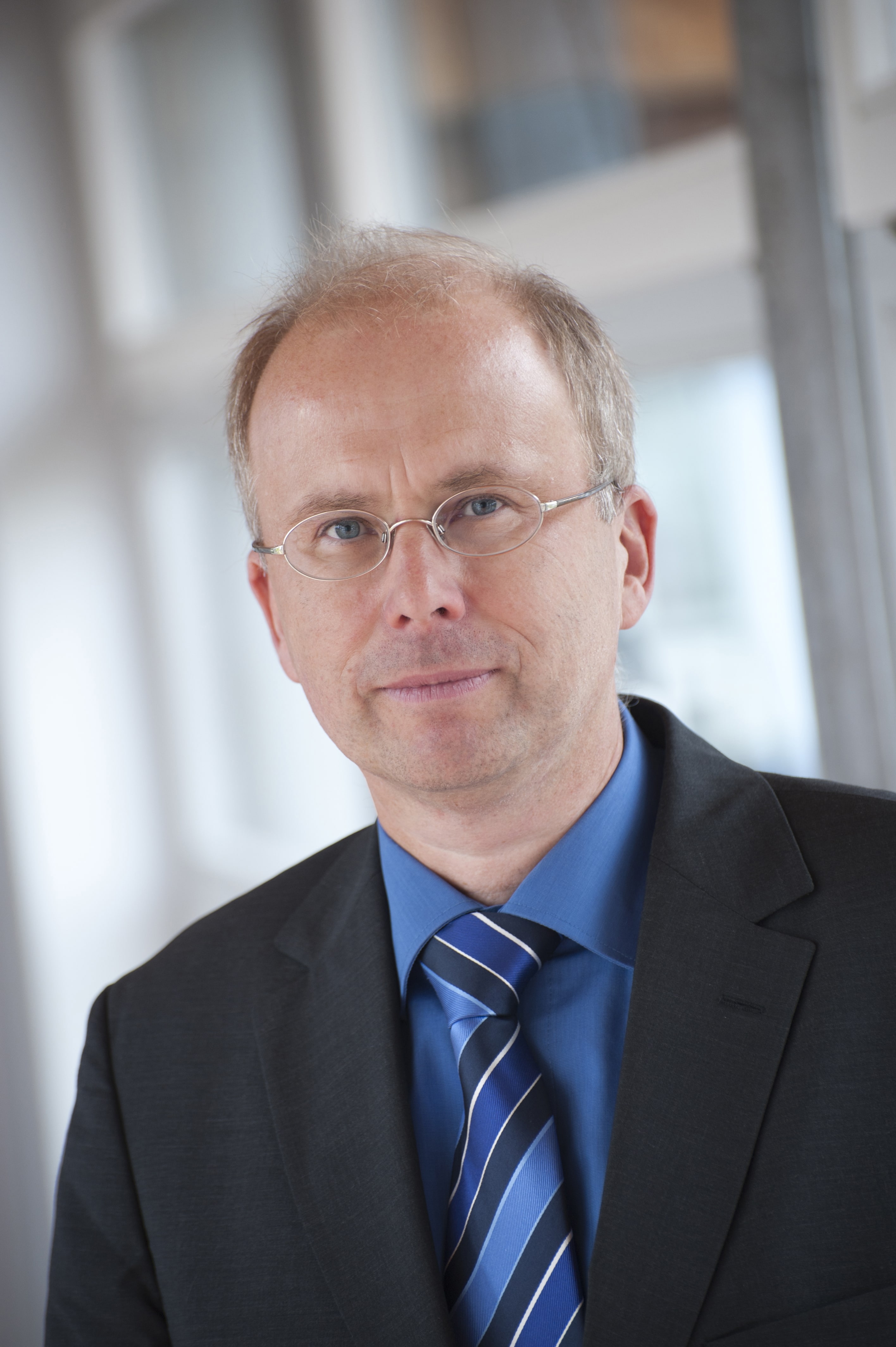}}]{Klaus Dietmayer} (Member, IEEE) was born in Celle, Germany in 1962. He received his Diploma degree in 1989 in Electrical Engineering from the Technical University of Braunschweig (Germany), and the Dr.-Ing. degree (equivalent to PhD) in 1994 from the University of Armed Forces in Hamburg (Germany). In 1994 he joined the Philips Semiconductors Systems Laboratory in Hamburg, Germany as a research engineer. Since 1996 he became a manager in the field of networks and sensors for automotive applications. In 2000 he was appointed to a professorship at the University of Ulm in the field of measurement and control. Currently he is Full Professor and Director of the Institute of Measurement, Control and Microtechnology in the school of Engineering and Computer Science at the University of Ulm. Research interests include information fusion, multi-object tracking, environment perception, situation understanding and trajectory planning for autonomous driving. Klaus Dietmayer is member of the IEEE and the German society of engineers VDI / VDE.
\end{IEEEbiography}
\clearpage
\appendix
This supplementary material to the main paper is structured as follows: Appx.~\ref{appendix:epistemic_aleatoric_uncertainty} introduces epistemic and aleatoric uncertainty in details, as well as another uncertainty categorization based on the development and operation of a perception system. Appx.~\ref{appendix:softmax_function_with_gaussian} explains how to estimate classification uncertainty with a distribution in softmax logits. Appx.~\ref{appendix:non-proper_scoring_rules} lists several common non-proper scoring metrics to quantify uncertainty estimates, and Appx.~\ref{appendix:uncertainty_benchmarking} discusses several uncertainty estimation benchmarks. After introducing generic object detection in Appx.~\ref{appendix:deterministic_object_detection}, the datasets used in our experiments are compared in Appx.~\ref{appendix:datasets}.

\renewcommand{\arraystretch}{1.6}
\begin{table*}[t] 
	\normalsize
	\caption{The applications of uncertainty estimation in autonomous driving and example references.}\label{tab:summary_uncertainty_estimation_application}
	\resizebox{1\linewidth}{!}{\begin{tabular}{>{\raggedright\arraybackslash} p{0.23\linewidth}
	>{\raggedright\arraybackslash}p{0.23\linewidth}
	>{\raggedright\arraybackslash}p{0.07\linewidth}
	>{\raggedright\arraybackslash}p{0.1\linewidth}
	>{\raggedright\arraybackslash}p{0.38\linewidth}} 
			\rowcolor{lightgray!35}  
			\textbf{Problem} &  \textbf{Example Reference} & \textbf{Year} & \textbf{Unc. Type} & \textbf{Dataset}  \\
			Camera pose estimate & Kendall \etal~\cite{kendall2016modelling} & 2016 & E & Cambridge Landmarks~\cite{kendall2015posenet} \\
			Semantic segmentation & Kendell \etal~\cite{kendall2015bayesian} & 2017 & E & CamVid~\cite{brostow2009semantic}, SUN RGB-D~\cite{song2015sun}, Pascal-VOC~\cite{everingham2010pascal} \\
			Image annotation & Mackowiak \etal~\cite{mackowiak2018cereals} & 2018 & E & Cityscapes~\cite{cordts2016cityscapes} \\
			Optical flow & Ilg \etal~\cite{ilg2018uncertainty} & 2018 & E+A & KITTI~\cite{Geiger2012CVPR}, MPI-Sintel Flow~\cite{butler2012naturalistic}  \\ 
			Learning by demonstration of driving control & Choi \etal~\cite{choi2018uncertainty} & 2018 & A & US Highway 101~\cite{colyar2007us} \\
			End-to-end driving & Michelmore \etal~\cite{michelmore2019uncertainty} & 2019 & E & CARLA Simulation~\cite{dosovitskiy2017carla} \\
			Trajectory prediction & Hu \etal~\cite{hu2020probabilistic} & 2020 & A & Self-recorded data \\
			Depth completion & Walz \etal~\cite{walz2020uncertainty} & 2020 & A & Gated2Depth~\cite{gruber2019gated2depth}\\
			Visual tracking & Danelljan \etal~\cite{danelljan2020probabilistic} & 2020 & A & OTB100~\cite{wu2013online}, UAV123~\cite{mueller2016benchmark}, NFS~\cite{kiani2017need} and four others\\
			\bottomrule
	\end{tabular}}
\end{table*}

\subsection{Epistemic and Aleatoric Uncertainty}\label{appendix:epistemic_aleatoric_uncertainty}
\begin{figure}[tpb]
	\centering
	\begin{minipage}{1\linewidth}
    	\centering
    	\subfigure[Image]{\label{fig:uncertainty_1}\includegraphics[width=0.3\linewidth]{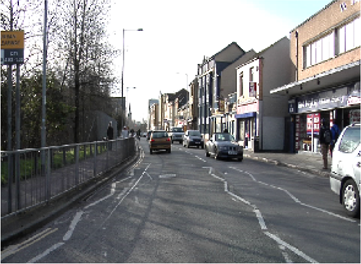}}
    	\hfill
    	\subfigure[Label]{\label{fig:uncertainty_2}\includegraphics[width=0.3\linewidth]{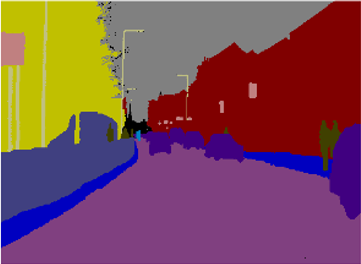}}
    	\hfill
    	\subfigure[Prediction]{\label{fig:uncertainty_3}\includegraphics[width=0.3\linewidth]{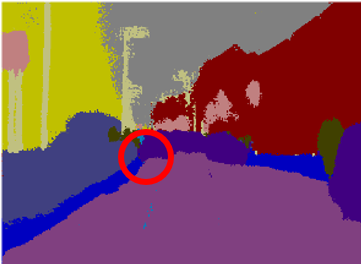}}
    	\vfill
    	\hspace{1.3cm}
        \subfigure[Aleatoric uncertainty]{\label{fig:uncertainty_4}\includegraphics[width=0.3\linewidth]{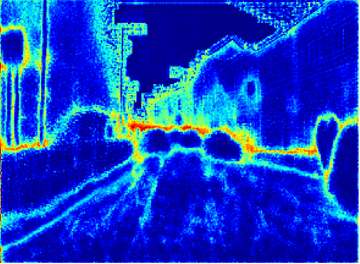}}
        \hfill
        \subfigure[Epistemic uncertainty]{\label{fig:uncertainty_5}\includegraphics[width=0.3\linewidth]{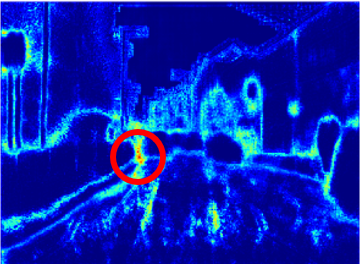}}
        \hspace{1.3cm}
    \end{minipage}
	\caption{An illustration of epistemic and aleatoric uncertainty in semantic segmentation on the CamVid dataset~\cite{brostow2009semantic}. The epistemic uncertainty is measured by the Mutual Information, and the aleatoric uncertainty by the Shannon Entropy. Uncertainty estimation networks are adapted from Postel \etal~\cite{postels2019sampling}. The different behaviours of epistemic vs aleatoric uncertainty can be observed in (d) and (e). For example, the aleatoric uncertainty is more related to the object boundary, whereas the epistemic uncertainty is affected by the prediction inaccuracy (the front car marked with circle  in the image is badly-segmented, and depicts high epistemic uncertainty in (e)).} \label{fig:uncertainty}
\end{figure}

In Sec.~\ref{uncertainty_estimation:what_uncertainty_can_we_model}, we have introduced that predictive uncertainty can be decomposed into \textit{epistemic} uncertainty and \textit{aleatoric} uncertainty following the Bayesian Neural Network framework. Fig.~\ref{fig:uncertainty} shows an example of epistemic and aleatoric uncertainties in visual semantic segmentation, and Tab.~\ref{tab:summary_uncertainty_estimation_application} lists several uncertainty estimation applications in autonomous driving. Note that although epistemic and aleatoric uncertainty are defined in the context of Bayesian Neural Networks, the two types of uncertainty can be captured by non-Bayesian methods such as Deep Ensembles and Direct Modelling as well (cf. Sec.~\ref{uncertainty_estimation:methods}). Furthermore, both types of uncertainty are not mutually exclusive~\cite{kendall2017uncertainties}; explicitly modeling only one may result in a predictive probability that reflects the properties of the other type of uncertainty as well.

\subsubsection{Another Uncertainty Categorization Method} 
\begin{figure}[tpb]
	\centering
	\includegraphics[width=1\linewidth]{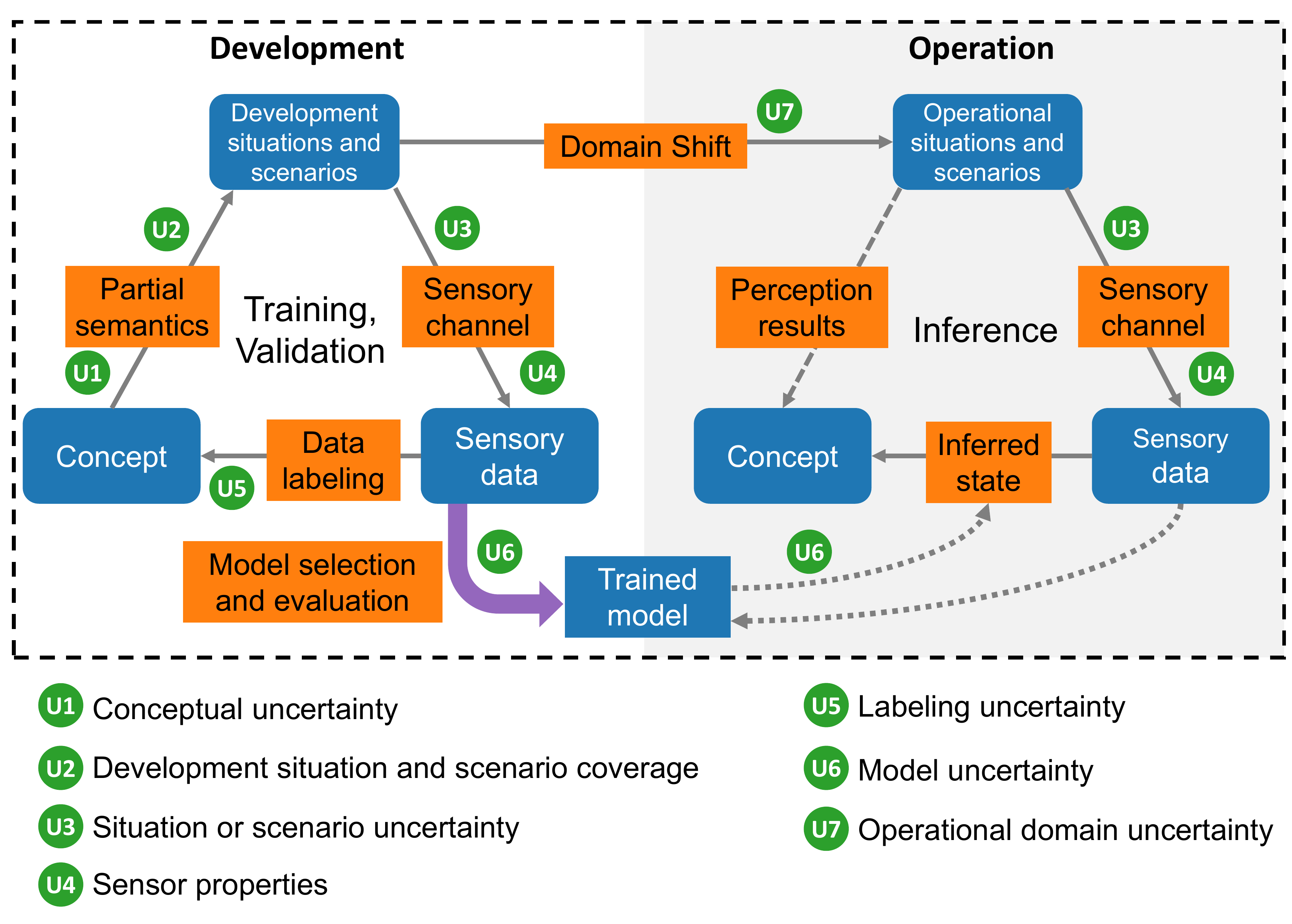}
	\caption{Uncertainty categorization according to the procedure of developing and deploying a perception system. Seven distinct sources of uncertainty are defined. Adapted from~\cite{czarnecki2018towards}.}\label{fig:uncertainty_categorization}
\end{figure}

Beyond epistemic and aleatoric uncertainty, Czarnecki \etal~\cite{czarnecki2018towards} define seven sources of perceptual uncertainty to be considered for the development and operation of a perception system (such as object detection, semantic segmentation etc.), as shown by Fig.~\ref{fig:uncertainty_categorization}. During development, there exists uncertainty when defining the perception problems (U1), collecting and annotating data with selected scenarios (U2-U5), as well as training the model (U6). During operation, an additional type of uncertainty arises due to distribution shift between the training and testing data splits (U7).

Taking object detection as an example. Conceptual uncertainty (U1) would occur when defining class labels (e.g. shall we distinguish between ``Cars'' and ``Vans'', or regard them as ``Vehicles''?). Due to the limitation of weather conditions or locations when recording data, the training dataset could not cover all driving scenarios, which introduces U2 and U3 uncertainties. Labeling the bounding box in a RGB camera image captured during night drive is difficult due to low illumination (U4 uncertainty), and depending on the expertise of annotators, the quality of such bounding box label is different (U5 uncertainty). The network architecture has an impact on how an object detector fits the training dataset, resulting in U7 uncertainty. Finally, employing the object detector trained with highway datasets to urban scenarios would cause the domain shift problem, and result in U7 uncertainty.


\subsection{Estimating Classification Uncertainty with Softmax Logits}\label{appendix:softmax_function_with_gaussian}
Instead of the common way of using the standard softmax function with the cross-entropy loss, Kendall \etal~\cite{kendall2017uncertainties} combine the softmax function with a Gaussian distribution to estimate classification uncertainty. Classification uncertainty is learnt by assuming that each element in the softmax logit vector is independently Gaussian distributed, with its mean and variance directly predicted by the network output layers. A softmax logit is the unnormalized log probability output the model before the softmax function. Let a softmax logit regression variable for the class $c$ be $l_c$, its distribution is given in the form: $p(l_c|\mathbf{x},\mathbf{W}) = \mathcal{N}\big(l_c|\hat{\mu}_c(\mathbf{x},\mathbf{W}), \hat{\sigma}_c^2(\mathbf{x},\mathbf{W})\big)$, where $\hat{\mu}_c(\mathbf{x},\mathbf{W}), \hat{\sigma}_c^2(\mathbf{x},\mathbf{W})$ refer to the predicted mean and variance of the softmax logit. Training this softmax logit is different from the normal regression variable shown in Eq.~\ref{eq:attenuated_loss}, as there are only ground truth for object categories. In this regard, Kendall \etal~\cite{kendall2017uncertainties} propose to sample the softmax logit based on the predicted Gaussian distribution via the re-parametrization trick, and then transform the sampled softmax logit into the softmax score to calculate a standard classification loss, such as the cross-entropy loss or the focal loss~\cite{lin2018focal}.

\subsection{Additional Scoring Metrics For Evaluating Uncertainty In Literature}
\label{appendix:non-proper_scoring_rules}
The following section introduces several common evaluation methods for uncertainty estimation. Several benchmarks based on those evaluation methods are discussed in Appx.~\ref{appendix:uncertainty_benchmarking}.
\subsubsection{Shannon Entropy }\label{uncertainty_estimation:metrics:se}
Shannon Entropy (SE) is a common evaluation metric to estimate the quality of predictive uncertainty for classification tasks \cite{Gal2016Uncertainty}. For a class label $y$ of $C$ classes, the SE, $\mathcal{H}$, is measured by: 
\begin{flalign}\label{eq:shannon_entropy}
\begin{split}
\mathcal{H}(y|\mathbf{x}, \mathcal{D}) = -\sum_{c=1}^C p(y=c|\mathbf{x},\mathcal{D}) \log \big(p(y=c|\mathbf{x},\mathcal{D})\big).
\end{split}
\end{flalign}
SE reaches a minimum value when the network is certain in its prediction, i.e. $p(y=c|\mathbf{x},\mathcal{D})=0$ or $1$, and maximum when the prediction follows an uniform distribution, i.e. $p(y=c|\mathbf{x},\mathcal{D})=\frac{1}{C}$. 

\subsubsection{Mutual Information} \label{uncertainty_estimation:metrics:mi}
Another common evaluation metric for measuring the quality of uncertainty in classification tasks is Mutual Information (MI). A practical MI formulation was introduced in~\cite{Gal2016Uncertainty}, which measures the information gain of the predictive probability, when introducing the posterior distribution of model parameters $\mathbf{W}$:
\begin{equation}\label{eq:mutual_information}
\mathcal{I}(y,\mathbf{W}|\mathbf{x}, \mathcal{D}) = \mathcal{H}(y|\mathbf{x},\mathcal{D})- \mathbb{E}_{p(\mathbf{W}|\mathcal{D})} \big[\mathcal{H}(y|\mathbf{x},\mathbf{W})\big], 
\end{equation}
where $\mathbb{E}[\cdot]$ is the expectation operation, and $\mathcal{H}(y|\mathbf{x},\mathbf{W})$ is the conditional Shannon Entropy with respect to $\mathbf{W}$, given by:
\begin{equation}
\mathcal{H}(y|\mathbf{x},\mathbf{W})= -\sum_{c=1}^C p(y=c|\mathbf{x},\mathbf{W}) \log \big(p(y=c|\mathbf{x},\mathbf{W})\big).
\end{equation}
The first term on the right-hand side of Eq.~\ref{eq:mutual_information} is calculated using Eq.~\ref{eq:shannon_entropy}. The expectation in the second term is approximated by averaging the network's predictions with the network's weight samples (e.g. sampled from the MC-Dropout~\cite{Gal2016Uncertainty} or the Deep Ensembles~\cite{lakshminarayanan2017simple} approaches). Different from SE which directly measures the predictive uncertainty, MI captures the variations within a model brought by the network's weights $\mathbf{W}$. Therefore, it reflects the model uncertainty. An MI score ranges between $[0,1]$, with a larger value indicating higher uncertainty. 

\subsubsection{Error Curve}
This method proposed by Ilg \etal~\cite{ilg2018uncertainty} measures how predictive uncertainty matches true prediction errors. It assumes that a well-estimated predictive uncertainty should correlate with the true error, and by gradually removing the predictions with the highest uncertainty, the average errors in the rest of the predictions will decrease. To draw an error curve, a model makes predictions with the uncertainty estimates through a test dataset. Those predictions are ranked according to their uncertainties, such as the Shannon Entropy or Mutual Information scores in classification, and predicted variances in regression. The error curve reports the percentage of removed testing samples according to their uncertainty ranking on the horizontal axis, and the averaged errors in the rest of the predictions on the vertical axis (such as the cross entropy errors in classification, and mean squared errors in regression).

\subsubsection{Total Variance (TV)}
The metric proposed by Feng \etal~\cite{feng2018towards} is designed to measure the dispersion of a probability distribution in regression tasks, by calculating the trace of its covariance matrix, i.e. summing up all variances in the diagonal line of the covariance matrix. A TV score ranges within $[0, +\infty)$, with a larger value indicating higher uncertainty. Note that TV only quantifies the variance in each regression variable, and ignores the correlations among regression variables.

\subsection{Benchmarking Uncertainty Estimation Methods}\label{appendix:uncertainty_benchmarking}
Previous works have evaluated and compared existing uncertainty estimation methods mainly in open-world scenarios. In this setting, the test data distribution is different from the training dataset (also known as \textit{domain shift}), or there exist objects which a perception module has never seen before (also known as \textit{out-of-distribution}). This setting is of great interest from the practical perspective of applications such as autonomous driving. Snoeck \etal~\cite{snoek2019can} conduct a large-scale study to benchmark classification uncertainty estimation, including image classification using MNIST, ImageNet~\cite{deng2009imagenet} and CIFAR~\cite{krizhevsky2009learning} datasets, as well the text and ad-click classification. They compare softmax output, MC-dropout~\cite{Gal2016Uncertainty}, Deep Ensembles~\cite{lakshminarayanan2017simple}, temperature scaling (as the post-hoc calibration tool)~\cite{guo2017calibration}, as well as several Bayesian inference approaches under the domain shift setting. Filos \etal~\cite{filos2019systematic} benchmark uncertainty estimation methods in the medical image application (diabetic retinopathy diagnosis). Blum~\etal~\cite{blum2019fishyscapes} build the Fishyscapes dataset to benchmark uncertainty estimation methods in semantic segmentation for anomaly detection. The dataset is based on Cityscapes~\cite{cordts2016cityscapes}, which is a popular semantic segmentation dataset for urban autonomous driving. 

\subsection{Detailed Introduction of Generic Object Detection}
\label{appendix:deterministic_object_detection}
Object detection is a multi-task problem that requires to jointly recognize objects with classification scores, and localize objects with tightly fitting bounding boxes. State-of-the-art object detectors using deep learning can follow one-stage, two-stage, or the recently-proposed sequence-to-sequence-mapping detection pipelines. In the one-stage object detection pipeline, a deep learning model is employed to directly map the input data to bounding boxes and classification scores in a single shot (e.g. YOLO~\cite{redmon2016you}, SSD~\cite{liu2016ssd}, and RetinaNet \cite{lin2018focal}). In contrast, the two-stage object detection pipeline extracts several class-agnostic object candidates called regions of interests (ROIs) or region proposals (RPs) in the first stage, and refines their class and bounding box attributes in the second stage. Those ROI can be extracted by non-deep learning approaches such as clustering~\cite{asvadi2017depthcn} and voting~\cite{engelcke2017vote3deep}, or by a neural network such as a Region Proposal Network (RPN) in the Faster-RCNN architecture~\cite{ren2015faster}. While common object detectors require prior anchors and output post-processing stage such as Non-Maximum Suppression (NMS) to suppress duplicate detections. Usually, both one-stage and two-stage pipelines require a prior anchor grid and an output post-processing stage such as Non-Maximum Suppression (NMS) to suppress duplicate detection results. Recently, Tian~\etal~\cite{tian2019fcos} propose an anchor-free detector by estimating objects at every pixel, and designing a center-ness loss to allow efficient training of such formulation. Carion~\etal~\cite{carion2020end} take a more drastic approach to avoid prior anchor grid, by reformulating the object detection task as a sequence-to-sequence mapping problem, leading to Detection Transformers (DETR). DETR directly maps a sequence of query points into objects in the scene without the prior anchor grid. The method also avoids duplicate detections through training with the optimal positive-negative sample assignment, obviating the need of a post-processing step such as NMS.

While the standard 2D convolution is the dominant method to process RGB camera images, there are several methods to process point clouds from LiDAR and Radar sensors, as well as pseudo point cloud from Stereo cameras. Those point clouds can be projected onto 2D feature maps, and processed by standard 2D CNNs~\cite{chen2017multi,ku2017joint,simon2018complex,yang2018pixor}. They can also be discretized in voxel~\cite{zhou2017voxelnet,shi2020pv,yan2018second,lang2018pointpillars,fei2020semanticvoxels} and processed through 3D CNN with sparse convolution operations~\cite{yan2018second}, or processed in the continuous vector space without voxelization using the Point-Net based backbone~\cite{qi2017pointnet,xu2017pointfusion,shi2019pointrcnn}. For a more detailed summary we refer interested readers to~\cite{feng2019modal} for object detection in 3D point clouds, and~\cite{liu2020deep} for object detection in 2D images.

\subsection{Experimental Setup: Datasets and Dataset Shift}
\label{appendix:datasets}
Recent work on evaluating uncertainty estimates of deep learning models for classification tasks~\cite{snoek2019can} suggests the necessity of uncertainty evaluation under dataset shift. In this regard, we chose three common object detection datasets with RGB camera images for autonomous driving, in order to evaluate and compare our probabilistic object detectors:
\begin{itemize}
    \item \textbf{Berkeley Deep Drive 100K (BDD) Dataset~\cite{yu2020bdd100k}} has $80,000$ image frames at a $1280\times720$ resolution, with $70,000/10,000$ training/validation data split. The BDD dataset was recorded in North America with diverse scene types such as city streets, residential areas, and highways, and in diverse weather conditions. The dataset contains an equal number of frames recorded during daytime and nighttime.
    \item \textbf{KITTI autonomous driving dataset~\cite{Geiger2012CVPR}} contains $7,481$ image frames at a resolution of $1242\times375$. The dataset was recorded in Europe, only in clear weather and in daytime hours.
    \item \textbf{Lyft autonomous driving dataset~\cite{kesten2019lyft}} has $158,757$ image frames at a $1224\times1024$ resolution. The Lyft Dataset was recorded in a single city (Palo Alto) in the US, and only contains 3D bounding box annotations. We generate 2D bounding box annotations for objects in the scene by projecting 3D bounding boxes using the Lyft Dataset API. Similar to KITTI, the Lyft dataset was captured during daytime and in clear weather.
\end{itemize}

\begin{figure}[!tpb]
	\centering
	\includegraphics[width=\columnwidth,trim=1 1 1 1,clip]{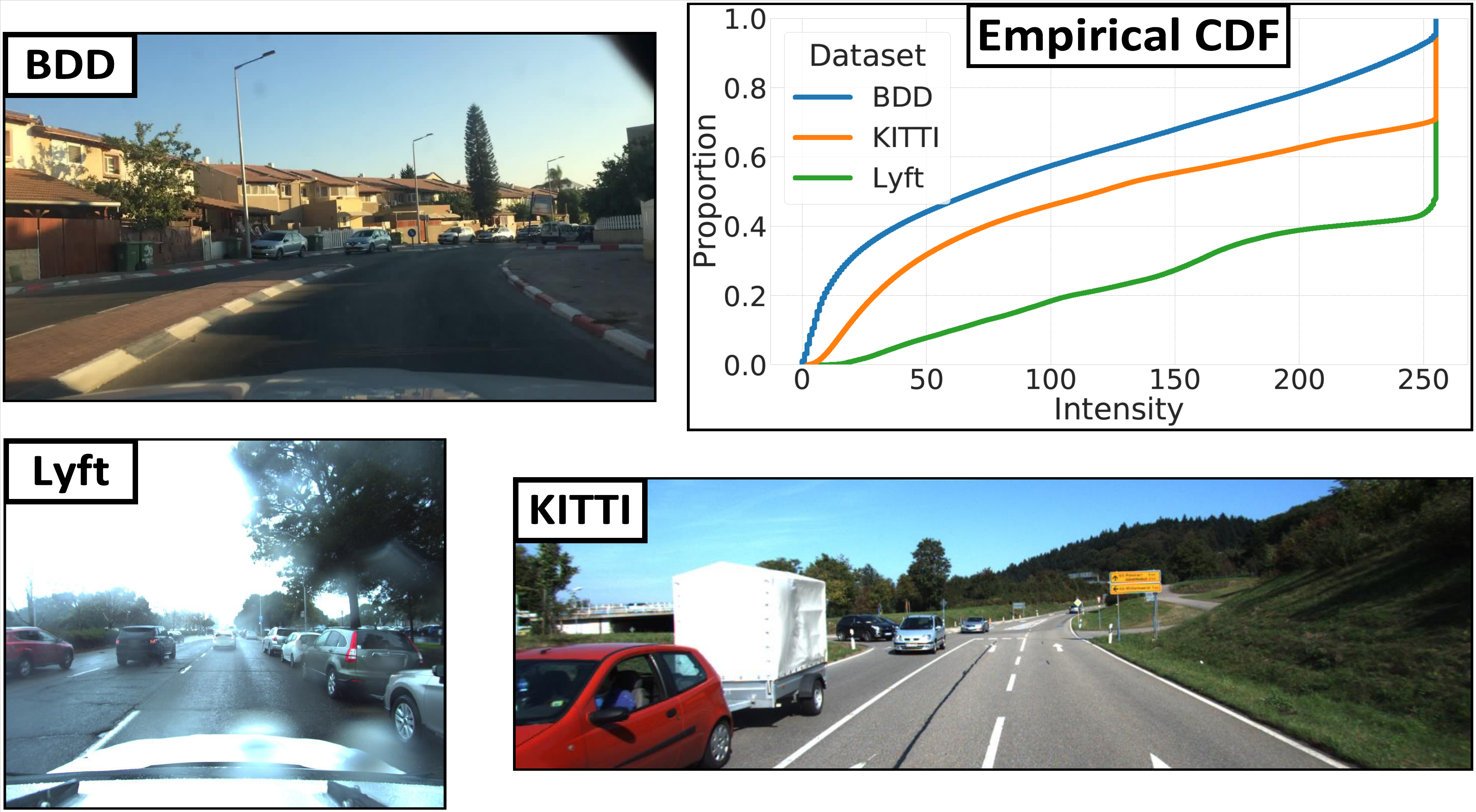}
	\caption{Sample images from BDD, Lyft and KITTI datasets showing differences in aspect ratios, camera view points, and scene appearance between the three datasets. \textbf{Top Right:} Plot of the empirical cumulative distribution function (CDF) of the grayscale intensity values of all images in BDD, KITTI, and Lyft datasets. The CDF of KITTI is much closer to that of BDD than the CDF of Lyft.}\label{fig:dataset_images}
\end{figure}

Both KITTI and Lyft have a different camera view point, image resolution, and image quality compared to BDD, which is visible when looking at example images from the three datasets in Fig~\ref{fig:dataset_images}. Furthermore, Fig~\ref{fig:dataset_images} also shows that the empirical cumulative distribution function (CDF) of the greyscale intensity values of images from the KITTI dataset is much closer to the CDF of BDD than the CDF of Lyft, suggesting a smaller domain shift from BDD to KITTI than that from BDD to Lyft. The CDF plots in Fig~\ref{fig:dataset_images} provide insight on the lack of exposure compensation in the Lyft dataset, where $\sim 60\%$ of pixels have an intensity greater than 250, and are therefore very bright (see Fig~\ref{fig:dataset_images}). The dataset shift from BDD to Lyft is further amplified as 2D bounding boxes are directly labeled on images for the BDD dataset, whereas 2D bounding boxes are generated by projecting 3D labels in the Lyft dataset. In summary, both KITTI and Lyft data is naturally shifted from the data in the BDD dataset, with KITTI having a lower magnitude of shift when compared to Lyft.

\end{document}